\documentclass[runningheads]{llncs}

\RequirePackage{fix-cm}
\RequirePackage{rotating}
\usepackage{graphicx}
\usepackage{cite}
\usepackage{amsmath}
\usepackage{xcolor}
\usepackage{float}
\usepackage{booktabs}
\usepackage{rotating}
\usepackage{amssymb}
\usepackage{pifont}
\newcommand{\cmark}{\ding{51}}%
\definecolor{green}{RGB}{0, 150, 0}
\begin{document}

\title{An overview of mixing augmentation methods and augmentation strategies}

\author{Dominik Lewy \and
Jacek Ma{\'n}dziuk}
\institute{
    Faculty of Mathematics and Information Science, Warsaw University of Technology, Koszykowa 75, 00-662 Warsaw, Poland\\
    \email{dominik.lewy@gmail.com, mandziuk@mini.pw.edu.pl}
    }
       
\maketitle

\begin{abstract}
Deep Convolutional Neural Networks have made an incredible progress in many Computer Vision tasks. This progress, however, often relies on the availability of large amounts of the training data, required to prevent over-fitting, which in many domains entails significant cost of manual data labeling. An alternative approach is application of data augmentation (DA) techniques that aim at model regularization by creating additional observations from the available ones. This survey focuses on two DA research streams: image mixing and automated selection of augmentation strategies. First, the presented methods are briefly described, and then qualitatively compared with respect to their key characteristics. Various quantitative comparisons are also included based on the results reported in recent DA literature. This review mainly covers the methods published in the materials of top-tier conferences and in leading journals in the years 2017-2021.

\end{abstract}

\section{Introduction}
\label{sec:start}

In recent years Deep Learning (DL) research has made an incredible progress in solving complex image-related problems. This progress and the accompanying raising interest in this area were greatly motivated by the first widely-known successful application of Convolutional Neural Networks (CNNs) to the problem of image classification in $2012$, when AlexNet~\cite{AlexNet} clearly outperformed shallow methods. There were three main reasons for this success: advancement in deep network architectures, the availability of huge computation power, and access to large amounts of training data. An acclaimed example of a large data set that fostered further development in this area is ImageNet~\cite{ImageNet} around which a competition called ImageNet Large Scale Visual Recognition Challenge (ILSVRC) was organized. This competition enabled tracking the performance of various CNN models over time and spurred the development of several renowned architectures like VGG~\cite{VGG}, GoogleNet~\cite{GoogleNet} or ResNet~\cite{ResNet}. Throughout the years one could observe how the classification top-5 error plummets from more than $25\%$ in 2010-2011, when shallow methods were used, to less than $5\%$ in 2015 with the use of DL.

CNNs are successfully used in various computer vision tasks like image classification, object localization, object detection, image segmentation, action recognition in videos
or image captioning (in the last task usually in combination with Recurrent Neural Networks). The main focus of this survey is on image classification problem, since the vast majority of mixing augmentation methods were designed specifically with this task in mind.

Generally speaking, when creating any DL algorithm for image categorization the goal is to train the model to correctly predict classes for new (previously unseen images) based on available training data. This ability is called generalization as the model is able to generalize the knowledge extracted from training images to correctly classify images not encounter during training. A failure to generalize is referred to as model over-fitting and is an unwanted property. The goal of DL practitioner is to find a model that is complex enough to learn based on all training samples (training examples) but simple enough so as not to memorize these samples. In many practical cases a viable approach is to create a model that is overcomplex for a given training set but, at the same time, introducing mechanisms that prevent the algorithm from over-fitting.
There are various ways to increase generalization ability of DL models (and avoid over-fitting), for example by means of regularization mechanisms~\cite{Regularization_survey} such as dropout, weight decay or batch normalization, as well as an application of specific training strategies like transfer learning or zero-shot learning.

Although many tasks can be solved using the data sets, model architectures and computational power available right now, in quite many domains the lack of sufficient amounts of training data remains one of the key obstacles in DL application. This problem is especially severe in the areas where gathering of labeled training samples is costly or requires special skills (e.g. medical images), or the scale of the business case is not large enough to justify an investment into labeled training data gathering.

One of possible remedies, leading to partial alleviation of this problem, is data augmentation (DA), which is one of the regularization methods~\cite{Regularization_survey} designed to create additional observations based on available ones and thus increase the size of the training set. The focus of this survey is on DA methods that apply certain transformations to the source (original) images in order to create new training examples. Both, data warping (i.e. applying any label preserving transformation to the image) and synthetic over-sampling (i.e. creation of artificial samples) approaches are considered. Additionally, we discuss strategies for selection of the augmentation technique(s) best suited for a given task.

This review paper complements recent DA survey~\cite{Augmentation_for_DL_survey} that covered a full spectrum of various data augmentation methods, ranging from traditional augmentation methods to Generative Adversarial Network (GAN) based augmentations.
The paper significantly extends the above-cited work in the area of \emph{mixing augmentation methods} and \emph{augmentation strategies}, which are only briefly covered in~\cite{Augmentation_for_DL_survey}.

The main source of papers for this review are recent top-tier conferences related to Artificial Intelligence, Computer Vision, and Machine Learning, from the period 2017-2021.
%
Certain exceptions include the cases of relevant and highly influential papers published in other venues.

\subsection{Taxonomy of presented data augmentation methods}
\label{sec:taxonomy}

Presented methods are divided into two high level categories: \emph{data augmentation methods} and \emph{strategies for data augmentation method(s) selection}. The first group is further divided into approaches that consist in \emph{erasing part of the image} and those relying on \emph{image mixing}. Technically, the former methods can also be thought of as mixing ones, in which a given image is mixed with an empty image. However, due to their somewhat different specificity, and following the taxonomy generally agreed within the DA community, these two groups of methods are considered separately.

Strategies for data augmentation selection are mostly applied in the literature to traditional DA techniques, i.e. label preserving geometric transformations or operations in the color space, and only occasionally to more advanced augmentation methods.

While these two areas (mixing DA methods and DA selection strategies) are compared in the literature they are rather not combined within one DA system. In this review both groups of methods are intentionally considered together, since we hypothesize that future development of DA research will largely include application of DA selection strategies to advanced mixing or GAN-based augmentation methods.

\subsection{Notation and general remarks}
\label{sec:notation}

\begin{figure}[!ht]
\includegraphics[width=1\linewidth]{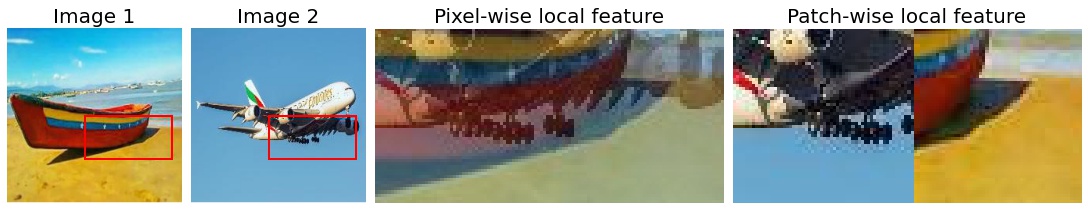}
\caption{
From left to right: two sample images and examples of pixel-wise and patch-wise mixing, respectively. The pixel-wise and patch-wise images present the zoomed region indicated by a red rectangle to show the detailed characteristics of the mixed images.
}
\label{fig:lin_vs_non_lin}
\end{figure}

Let us start with an introduction of the base notation.
First of all, mixing DA techniques are divided into two main classes: those that mix images using pixel-wise weighted average (referred to as \textbf{pixel-wise mixing}) and those that mix images spatially by means of extracting patches from different images and joining them together (referred to as \textbf{patch-wise mixing} methods). Examples from both classes are presented in Figure~\ref{fig:lin_vs_non_lin}.

Furthermore, whenever a result of applying any given DA technique is presented, the \textbf{baseline result} or simply \textbf{baseline} would refer to the respective result obtained using the same architecture, trained on the same data and using the same training hyper-parameters, but with no application of the analyzed DA method.
For the sake of consistency all DA methods will be written in \emph{italic font}, e.g. \emph{Mixup}, \emph{CutMix} or \emph{Attentive CutMix}.

The remainder of the survey is organized as follows: section~\ref{sec:erasing_image} presents DA methods that rely on \emph{erasing part of the image} and section~\ref{sec:mixing_images} introduces the \emph{image mixing} methods.
The mixing methods are further \emph{qualitatively compared and analyzed in the context of their specific aspects} (e.g. in which part of the learning model the DA is applied, how many images are involved in a single DA sample preparation, whether DA method is pixel-wise or patch-wise, whether or not the DA technique mixes labels, and other) in section~\ref{sec:comparison}. A \emph{thorough quantitative experimental comparison} of the methods is presented in section~\ref{sec:erasing_mixing_evaluation}. The vast majority of results refer to CIFAR-10~\cite{cifar_10} and CIFAR-100~\cite{cifar_100} data sets which are most frequently used in DA methods assessment. Section~\ref{sec:augmentation_strategy} presents and compares strategies for optimal DA method selection. The last section concludes the paper by summarizing its main findings and pointing some possible future prospects in DA domain - notably application of the DA selection strategies to the class of mixing DA methods.

\section{Data augmentation by \emph{erasing part of the image}}
\label{sec:erasing_image}

Methods in this group rely on removing part of the image
either by masking it out~\cite{cutout,random_erasing} or replacing with certain noise~\cite{patch_gaussian}.


A foundational method in this area is \emph{Cutout}~\cite{cutout} that erases / masks a square region of the input image by zeroing the respective pixels. \emph{Cutout} was an inspiration for several subsequent methods, both erasing (e.g. \emph{Patch Gaussian}~\cite{patch_gaussian}) as well as mixing (e.g. \emph{CutMix}~\cite{cutmix}, \emph{SmoothMix}~\cite{smoothmix},
\emph{Attentive CutMix}~\cite{attentive_cutmix}, \emph{Puzzle Mix}~\cite{puzzle_mix}, \emph{Saliency Mix}~\cite{saliency_mix} or \emph{SnapMix}~\cite{snap_mix}) which are presented in further sections.

The main motivation behind \emph{Cutout} is
better utilization of the entire context of the image,
which is especially important when dealing with partial occlusion. Due to masking a particular region of the image in the input layer, the network needs to recreate the missing information based on the context. Conceptually, the technique is similar to dropout~\cite{dropout} and can be thought of as an extension of dropout to the input layer. However, there are two key differences: firstly, \emph{Cutout} is applied exclusively to the input layer (not to all layers as dropout is) and secondly, it drops \emph{regions} of the input image, rather than individual pixels. While there are variants of dropout that work with continuous regions, e.g. DropBlock~\cite{dropblock}, they are applied to randomly selected regions, different in every layer, so, unlike in \emph{Cutout}, the key regularization mechanism in this case relies on randomness.

The method uses just one hyperparameter, the size of the filter. The authors of~\cite{cutout} claim to have considered one more hyperparameter - the shape of the filter, but as the size turned out to be much more important, all experiments are focused on size optimization of the cutout, with a square shape assumed by default. Based on the experiments, the relation between model accuracy and filter size has a parabolic shape - the accuracy increases with an increase of the filter size, up to an optimal point after which masking out too much information deteriorates the resulting accuracy.


An important point to make is that \emph{Cutout}'s occlusion ratio (a proportion of pixels masked to all pixels) can vary even for filters of the same size due to various possible positions of the center point of the cutout region, which is selected randomly. This diversity of the effective filter size is crucial for achieving high performance. An alternative approach is to always select a center point of the filter far enough from the image edges, so as the entire mask would fall within the image, but apply the method with $50\%$ probability~\cite{cutout}.

Another erasing augmentation approach, \emph{Random Erasing}~\cite{random_erasing} is a deep dive into \emph{Cutout}
%
and its high level description is very similar to that of \emph{Cutout}. However, \emph{Random Erasing} is a more flexible approach and transforms the image in a slightly different manner. The method admits rectangular shape of the masking patch and its effective application depends on $3$ (instead of $1$ as in \emph{Cutout}) hyperparameters:
the \textit{erasing probability} $p$, the range of \textit{area ratio} $S_e/S\in[s_l, s_h]$ (\ref{eq:randomerasing:se}) between the masked region and the whole image, which controls the patch size, and the range of \textit{aspect ratio} $r_e\in[r_1, r_2]$ (\ref{eq:randomerasing:re}) between height and width of the masked region (\ref{eq:randomerasing:hewe}), which controls the patch shape.

Additionally, the method fills in the erased region differently to \emph{Cutout}. Each pixel's value within the masked region is replaced with a randomly selected value between $0$ and $255$ (other less successful approaches are discussed in~\cite{random_erasing}).

\begin{figure} [!ht]
\includegraphics[width=1\linewidth]{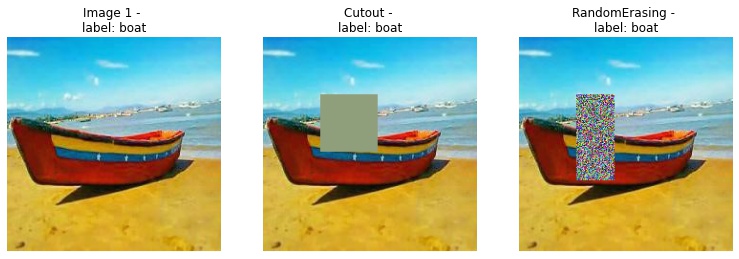}
\caption{
The effects of \emph{Cutout} and \emph{Random Erasing}.
}
\label{fig:cutout_random_erasing}
\end{figure}

Furthermore, the entire patch area must not extend beyond the original image range, so the following patch selection process (\ref{eq:randomerasing:se})-(\ref{eq:randomerasing:xeye}) is repeated until an appropriate patch is found.

\begin{equation}
\label{eq:randomerasing:se}
S_e \leftarrow \textnormal{Rand} (s_l, s_h) \times S
\end{equation}
\begin{equation}
\label{eq:randomerasing:re}
r_e \leftarrow \textnormal{Rand} (r_1, r_2)
\end{equation}
\begin{equation}
\label{eq:randomerasing:hewe}
H_e \leftarrow \sqrt{S_e \times r_e}, W_e \leftarrow \sqrt{\dfrac{S_e}{r_e}}
\end{equation}
\begin{equation}
\label{eq:randomerasing:xeye}
x_e \leftarrow \textnormal{Rand} (0, W), y_e \leftarrow \textnormal{Rand} (0, H)
\end{equation}

where $S$ is the image size, $S_e, r_e, H_e, W_e$ are the area, the aspect ratio, the height and the width of the filtering rectangle, respectively, and $x_e, y_e$ indicate the upper left corner of this rectangle. The effects of \emph{Cutout} and \emph{Random Erasing} are compared in Figure~\ref{fig:cutout_random_erasing}.

Generally, \emph{Random Erasing} yields better results than the baseline for all reasonable parameter settings, however, the final parameter values need to be optimized for a given data set. For instance, for CIFAR-10, it is recommended in~\cite{random_erasing} that the values of $p$, $s_h$ and $r_1$ be equal to $0.5$, $0.4$ and $0.3$, respectively.
The method is generally not the best choice for a standalone DA technique. It is inferior, for instance, to random flipping~\cite{Augmentation_for_DL_survey} or random cropping~\cite{Augmentation_for_DL_survey}, although when added on top of any of them, as a complementary augmentation method it improves classification accuracy.

\begin{figure}[!ht]
\includegraphics[width=1\linewidth]{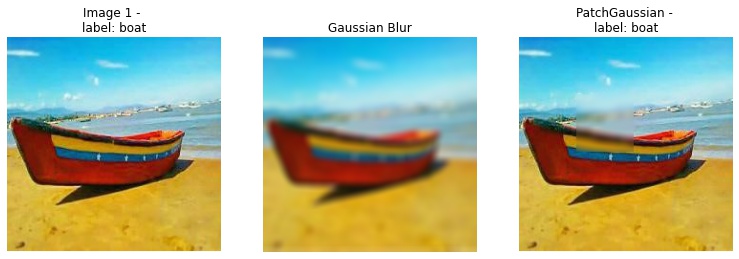}
\caption{
Original image, its version after application of Gaussian blur, and the effects of \emph{Patch Gaussian} augmentation.
}
\label{fig:patch_gaussian}
\end{figure}

The next erasing method, \emph{Patch Gaussian}~\cite{patch_gaussian}, is a hybrid of \emph{Cutout} and a Gaussian noise augmentation procedure. Adding Gaussian noise to the image region selected by \emph{Cutout} improves performance of the trained model
on corrupted images, in addition to its efficient dealing with partial occlusion.
The motivation behind \emph{Patch Gaussian} is twofold: (a) \emph{Cutout} is generally able to increase the model's accuracy but without ensuring its robustness, while (b) addition of Gaussian noise increases robustness but hurts accuracy of the model. The aim of \emph{Patch Gaussian} is to get the best of both worlds and combine high accuracy on a held-out set of images with the robustness to corrupted images. The method is parameterized by a patch size (same as in \emph{Cutout}) and the maximum standard deviation of Gaussian noise. Its effects are visualized in Figure~\ref{fig:patch_gaussian} and compared with application of Gaussian blur.

In order to confirm that high accuracy and robustness to noise are indeed caused by application of \emph{Patch Gaussian} augmentation and are not just a result of stacking data augmentation techniques, in ablation study presented in~\cite{patch_gaussian} \emph{Patch Gaussian} was compared with an application of \emph{Cutout} followed by the Gaussian noise addition, as well as with either of these two techniques applied alone to half of the batches. Among these approaches \emph{Patch Gaussian} achieved the smallest error on corrupted data while maintaining high accuracy on non-corrupted test samples.

An interesting approach is a combination of \emph{Patch Gaussian} with regularization strategy, or another DA technique. Using \emph{Patch Gaussian} together with any of the following strategies: weight decay~\cite{DL_book}, label smoothing~\cite{label_smoothing}, DropBlock~\cite{dropblock}, or \emph{AutoAugment}~\cite{AutoAugment} leads to decreasing of the error on corrupted data whilst making the error on non-corrupted data just marginally higher than the baseline.

The genre of erasing methods is relatively small since researchers quickly shifted their attention towards methods described in section~\ref{sec:mixing_images} that mix images in a patch-wise manner. Patch-wise mixing has similar effects to erasing part of the original image, but without reducing the training signal provided by each image. In case of mixing methods a masked part of the image is replaced by a part of another image, thus providing a stronger training stimulus.

\section{Data augmentation by \emph{image mixing}}
\label{sec:mixing_images}

\begin{figure}[!ht]
\includegraphics[width=1\linewidth]{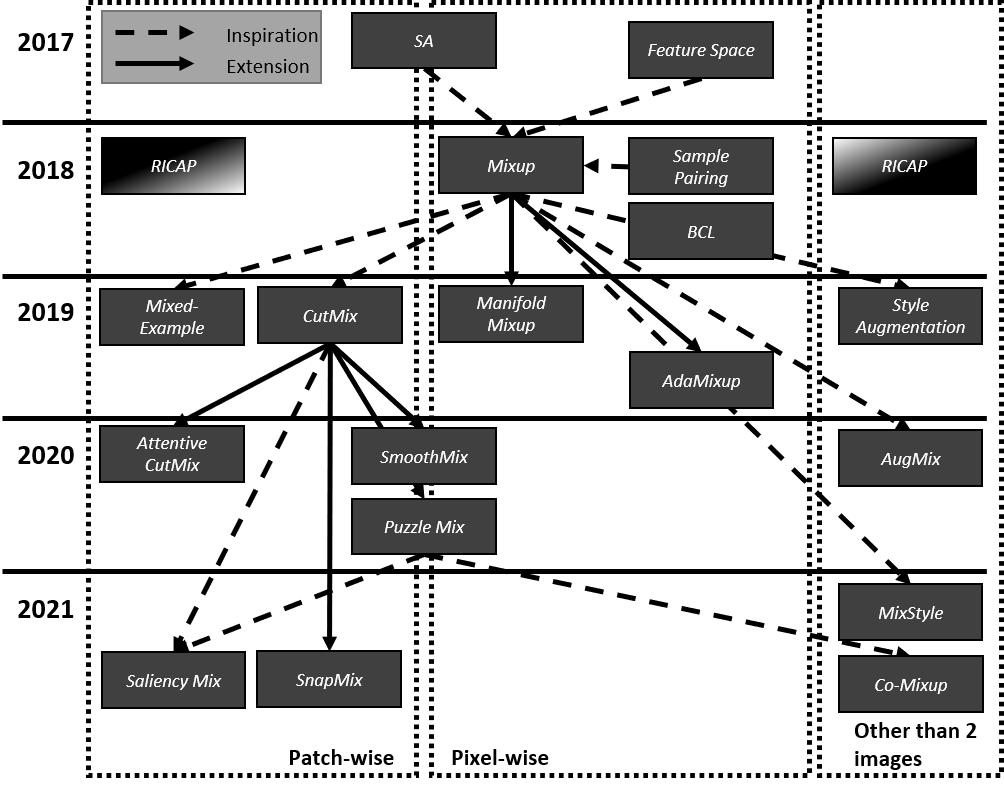}
\caption{
Image mixing DA methods presented on a time scale, with key characteristics and dependencies indicated. Dotted regions separate methods in which mixing takes pixel-wise form (Pixel-wise) from those with spatial mixing (Patch-wise), and those with mixing applied not to a pair of images, but either just one image and its transformed version or more than 2 images (Other than 2 images). Directed lines indicate inspirations (dotted lines) or direct extensions (solid lines) of the methods. Each method appears once except \emph{RICAP} which in a patch-wise manner mixes 4 images, hence appears in both Patch-wise and Other than 2 images groups.
}
\label{fig:mixing_map}
\end{figure}

Image mixing DA methods rely on blending two input images and their corresponding labels according to the following equations: %

\begin{equation}
\label{eq:mixing_images_x}
\tilde{x} = B \odot x_1 + (I-B) \odot x_2
\end{equation}
\begin{equation}
\label{eq:mixing_images_y}
\tilde{y} = \lambda y_1 + (1-\lambda) y_2
\end{equation}
where $x_1, x_2$ are original input images, $y_1, y_1$ are one-hot label encodings, $\lambda$ is a mixing ratio, $B$ is a mixing mask matrix suitable for both pixel-wise and patch-wise mixing and $I$ is an identity matrix of the same dimensionality as $B$. $\odot$ denotes element-wise matrix multiplication operation. The vast majority of approaches described in this section are built around equations (\ref{eq:mixing_images_x})-(\ref{eq:mixing_images_y}) and mainly differ by the method of $\lambda$ selection and construction of matrix $B$.

Figure~\ref{fig:mixing_map} presents a map of the mixing methods, indicating for each of them the publication date, certain key characteristics and relations to other methods. In the remainder of this section
the methods presented in Figure~\ref{fig:mixing_map} are described and discussed in more detail.

A founding mixing method is \emph{Mixup}~\cite{mixup} introduced in 2018, which
laid ground for many subsequent papers in this area~\cite{adamixup, manifold_mixup, mixed_example, cutmix, smoothmix, attentive_cutmix, puzzle_mix, saliency_mix, co_mixup, snap_mix, aug_mix, style_augmentation, mix_style}.

\emph{Mixup} constructs new training samples according to equations~(\ref{eq:mixing_images_x})-(\ref{eq:mixing_images_y}) by means of the same weighted mean of the images and their labels, i.e. the entire matrix $B$ is populated with $\lambda$. The underlying assumption of \emph{Mixup} is that linear interpolation of feature vectors should lead to an adequate linear combination of the associated labels. This linear combination of images / classes is controlled by $\lambda$, e.g., $\lambda=0.5$ leads to averaging the images and their corresponding labels, while $\lambda\in\{0,1\}$ preserves one of the original images and its label.

The authors of \emph{Mixup} additionally checked whether it is more beneficial to sample image pairs
from the entire data set vs. sampling them solely from observations belonging to the same class, and whether is it preferred to apply weighted average (\ref{eq:mixing_images_x})-(\ref{eq:mixing_images_y}) in the input layer vs. its application in subsequent layers.
The results presented in Table~\ref{tab:mixup_ablation} prove that it is more beneficial to mix images selected at random rather than within the same class and that mixing images in the input is more advantageous than mixing their latent representations.
\begin{table}[!ht]
\begin{tabular}{lll}
\toprule
                                          & Specification & Test error \\
                                                &               &            \\
\midrule
ERM - baseline                    &                         & 5.53       \\
\midrule
\emph{Mixup}                                    & RP       & 4.28       \\
                                                & KNN      & 4.98       \\
\midrule
mix labels and latent 			 & Layer 1       & 4.44       \\
representations (AC + RP)		& Layer 2       & 4.56       \\
                                                & Layer 3       & 5.39       \\
                                                & Layer 4       & 5.95       \\
                                                & Layer 5       & 5.39      \\
\midrule
\end{tabular}
\caption{
Part of the ablation study presented in~\cite{mixup}. Experiments were performed on CIFAR-10 data set and measured the median test error over the last 10 epochs (the rightmost column). The ERM denotes Empirical Risk Minimization which is the baseline approach in this study.
The first experiment checks whether images in a pair should be selected at random (RP) or should rather represent similar observations (KNN). The next one verifies whether it is more beneficial to apply data augmentation procedure to the latent representation (Layers 1-5) or to the input images (RP result in the first experiment).}
\label{tab:mixup_ablation}
\end{table}

Although~\cite{mixup} is considered to be a founding paper in the image mixing augmentation research line, historically, there were three papers published prior to \emph{Mixup} that tackled the problem of mixing images~\cite{smart_augmentation,feature_space,sample_pairing} but approached it from different angles.

\emph{Smart Augmentation} (\emph{SA})~\cite{smart_augmentation}, the first of these methods, attempts to learn the way of mixing that minimizes the loss. This learning-based approach is quite specific since the vast majority of the current augmentation papers focus on
applying effective, yet arbitrary chosen, mixing techniques. \emph{SA} employs two separate networks: the first one (called Augmentor) is responsible for mixing two or more images, while the other one (Target) is responsible for image classification. Selection of images for mixing is performed randomly and is limited to observations from the same class. Parameters of both networks are updated based on the loss of the Target network. The loss of the Augmentor network is furthered extended with an MSE-based comparison of the mixed image with randomly selected image (from the same class) other than the ones used to create the mixed sample. Conceptually, the approach is flexible and architectures of both networks can vary as long as the Augmentor network returns the output of the same size as the input. Once the training is completed the Augmentor is discarded.

Mixing images in a learned \emph{Feature Space} is proposed in the second of pre-\emph{Mixup} papers~\cite{feature_space}. The authors did not propose any particular name for their method and referred to it as \textit{``Data set augmentation in Feature Space''}. For the sake of brevity, we coined the name \textit{Feature Space} to this method. In \textit{Feature Space} augmentation, latent features are generated using a sequence-to-sequence model which enables cross domain application of the method, e.g. to both text and image classification. The set of operations performed on those latent features encompasses adding a random noise, as well as interpolation and extrapolation of features from different observations.

Technically, the method uses a sequence autoencoder (a stacked LSTM \cite{LSTM} with encoder and decoder layers) to generate the training features. Whilst method execution an image is propagated through the encoder network and subsequently its hidden representation is either augmented with Gaussian noise or mixed with a similar observation.
The mixing process consists in finding $K$ nearest neighbors in the feature space with the same class label as the selected image. For each pair of hidden representations (of a given image and one of its $K$ nearest neighbours) the mixing is performed according to one of the following equations:


\begin{equation}
\label{eqn:interpolation}
\ c'=(c_k-c_j)\lambda + c_j
\end{equation}
\begin{equation}
\label{eqn:extrapolation}
\ c'=(c_j-c_k)\lambda + c_j
\end{equation}

where $c_j$ is the vector representing latent features of a given image, $c_k$ is its neighboring vector in the latent feature space, and $\lambda\in [0,1]$ controls the degree of interpolation (\ref{eqn:interpolation}) and extrapolation (\ref{eqn:extrapolation}), respectively.
It is suggested in~\cite{feature_space} that extrapolation should be preferred over interpolation as it is able to create samples that display higher variability.

The third DA approach that appeared prior to \emph{Mixup}, \emph{SamplePairing}~\cite{sample_pairing}, is a relatively straightforward technique that constructs a new data sample by a pixel-wise averaging of two randomly selected images. Unlike \emph{Mixup} it does not consider the label of the second image and simply assigns the label of the first image to the newly created mixed training sample. 

\begin{figure}[!ht]
\includegraphics[width=1\linewidth]{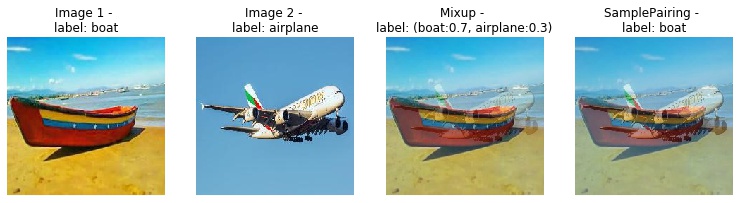}
\caption{
Comparison of \emph{Mixup} and \emph{SamplePairing} augmentations.
}
\label{fig:mixup_samplepairing}
\end{figure}

In the image classification task \emph{SamplePairing} is used as a pre-training mechanism which is turned on and off at regular intervals before being completely disabled at the end of the pre-training phase.
The effects of \emph{Mixup} and \emph{SamplePairing} are compared in Figure~\ref{fig:mixup_samplepairing}.
Both methods mix images in a pixel-wise manner: \emph{Mixup} with a user defined mixing ratio, in this case $0.7$ and \emph{SamplePairing} with a fixed ratio of $0.5$.

Another approach related to \emph{Mixup} is \emph{Between-Class learning} (\emph{BCL})~\cite{between_class}, published in the same year.
\emph{BCL} was initially proposed in the context of mixing sound waveforms~\cite{bcl_source} and then adjusted to image domain~\cite{between_class}. A major distinction between \emph{Mixup} and \emph{BCL} is that the latter mixes images from different classes only, whereas \emph{Mixup} selects two images at random, which makes it possible to mix two images from the same class.

There are two \emph{BCL} implementations proposed in~\cite{between_class}. The first one follows eq. (\ref{eq:mixing_images_x}) and the other one (named \emph{BC+}), instead of (\ref{eq:mixing_images_x}) applies eq. (\ref{eq:between_class}), which takes into account characteristics of the image, i.e. pixel-wise image mean and standard deviation:
%
%
\begin{equation}
\label{eq:between_class}
\tilde{x}=\frac{p(x_1-\mu_1)+(1-p)(x_2-\mu_2)}{\sqrt{p^2+(1-p)^2}} \textnormal{,}\ \ \ \ \ \ \ p = \frac{1}{1+ \frac{\sigma_1}{\sigma_2}*\frac{1-\lambda}{\lambda}}
\end{equation}
where $\lambda$ is the mixing ratio, $\sigma_1, \sigma_2$ are pixel-wise standard deviations, and $\mu_1, \mu_2$ are pixel-wise means, respectively. Both methods apply eq. (\ref{eq:mixing_images_y}) to create mixed label encoding.

During CNN training the goal of the network is to predict the newly created mixed label encoding which is equivalent to predicting the mixing ratio used to create the mixed sample.
For this task standard categorical cross entropy loss function is replaced with Kullback-Leibler divergence, which proved to be more effective in the ablation study presented in~\cite{between_class}.
Based on the ablation experiments it can also be concluded that arbitrary choosing one of the two possible labels for the mixed sample, instead of their linear combination, worsens the performance. Furthermore, while mixing examples from different classes renders better results, the performance also improves (compared to the baseline) if mixed images belong to the same class. Similarly to \emph{Mixup} study~\cite{mixup}, \cite{between_class} found that application of the mixing process in the input layer is more beneficial than its usage in subsequent layers. In the latter case, for deeper CNN layers mixing may even deteriorate performance.

Going back to the \emph{Mixup} method, subsequent related research was mainly concentrated along two axes: direct improvements of the method~\cite{adamixup,manifold_mixup} (discussed in section~\ref{sec:Mixup-direct}), and questioning the efficacy of pixel-wise mixing~\cite{mixed_example,cutmix,smoothmix,attentive_cutmix,saliency_mix} (discussed in section~\ref{sec:Mixup-non-linear}).

\subsection{Direct extensions of \emph{Mixup}}
\label{sec:Mixup-direct}

\emph{AdaMixup}~\cite{adamixup} was the first direct extension of \emph{Mixup}. The method attempts to learn a better than \emph{Mixup} mixing distribution to avoid the so-called \emph{manifold intrusion} problem, i.e. a situation in which a synthetically created image coincides with  a class which is different from any of the two classes assigned to images being mixed.

\emph{AdaMixup} extends \emph{Mixup} by adding 2 neural networks called Intrusion Discriminator (ID) and Policy Region Generator (PRG). The first one is responsible for predicting whether or not the resultant mixed sample would lead to manifold intrusion and the role of the latter one is to propose the mixing policy.
All three networks are trained jointly using the following loss function:
\begin{equation}
\label{eqn:ada_mixup}
\ L_{total} = L_{D}(H) + L_{D^{'}}(H,\{\pi_k\})+L_{intr}(\{\pi_k\}, \varphi)
\end{equation}
where $L_{D}$, $L_{D^{'}}$ and $L_{intr}$ are standard loss, data space regularization loss and intrusion regularization, respectively. $H$, $\varphi$ and $\{\pi_k\}$ denote the main classifier, ID and PRG, respectively.

Another direct extension of \emph{Mixup} is \emph{Manifold Mixup}~\cite{manifold_mixup}, which is motivated by an observation that \emph{Mixup} produces sharp decision boundaries in latent image representations.
\emph{Manifold Mixup} addresses this issue by performing image mixing operations in hidden layers with latent feature representations. More precisely, the method proposes to select randomly one layer in a network, either hidden or input, and then process two random mini-batches of data until reaching the selected layer, at which point the two mini-batches are mixed according to (\ref{eq:mixing_images_x}). Afterwards, the processing is continued until the output layer is reached where the loss is calculated and all network parameters are updated according to the gradients. If randomly selected layer happens to be the input one, the method is equivalent to \emph{Mixup}.

\emph{Manifold Mixup} achieves comparative or higher performance than \emph{Mixup} and \emph{AdaMixup}. Note that the results are somewhat contradictory to the conclusion presented in~\cite{mixup} that mixing should be applied in the input layer.
%
%
\subsection{Patch-wise extensions of \emph{Mixup}}
\label{sec:Mixup-non-linear}

The other stream of \emph{Mixup} follow-up papers, which question the efficacy of mixing pixels linearly, share many properties with the augmentation methods focused on occluding parts of the image, described in section~\ref{sec:erasing_image}. A primary approach here is \emph{Mixed-Example} method~\cite{mixed_example} which tests various patch-wise mixing approaches to conclude that they work comparably well to pixel-wise methods and even excel them in certain settings.

\begin{figure}[!ht]
\includegraphics[width=1\linewidth]{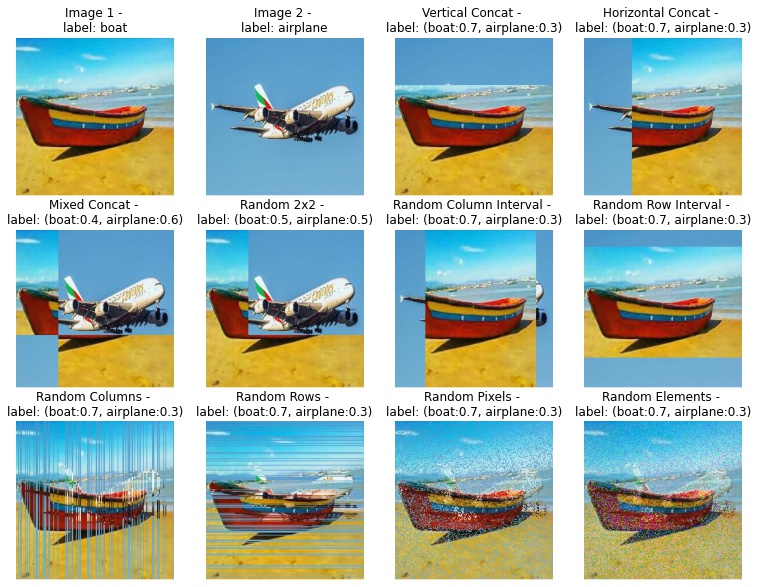}
\caption{
Effects of various patch-wise \emph{Mixed-Example} augmentation methods. \emph{Random Square} variant, which is visually equivalent to \emph{CutMix}, is omitted in the figure and presented in Figure~\ref{fig:cutmix_and_extensions}.}
\label{fig:mixed_example}
\end{figure}

An underlying assumption of \emph{Mixed-Example}~\cite{mixed_example} is that linear interpolations represent just a small subset of mixing operations that can potentially be used for data augmentation.
Consequently, the following patch-wise mixing schemes are proposed in~\cite{mixed_example}:
\begin{itemize}
\item \emph{Vertical Concat} - a method where bottom and upper parts of the new image come from two different images. This effect can be implemented with eq. (\ref{eq:mixing_images_x}) where $B$ has $1$s in the bottom part and $0$s in the upper part or vice versa;
\item \emph{Horizontal Concat} - same as \emph{Vertical Concat}, but in reference to the left and right parts of the new image;
\item \emph{Mixed Concat} - a combination of both the above methods in which a new image is constructed from 4 patches and patches on each diagonal come from the same source image;
\item \emph{Random 2x2} - a randomized version of \emph{Mixed Concat} where the source of each of the 4 patches is independently selected at random (either the first or the second image). \emph{Random 2x2} implements an idea similar to \emph{RICAP} method described below in this section, which combines 4 images and each patch comes from a different image;
\item \emph{Random Square} - a method in which a randomly chosen square from one image is placed on the other image. \emph{Random Square} is similar to \emph{CutMix} method described below in this section;
\item \emph{Random Column Interval} - a generalization of \emph{Horizontal Concat} where a randomly selected vertical stripe is picked in one image and pasted onto the other image;
\item \emph{Random Row Interval} - an analogous generalization of \emph{Vertical Concat};
\item \emph{Random Columns/Random Rows} - a further generalization of \emph{Random Interval} approaches in which a subset of rows/columns, respectively is chosen in one image and pasted onto another image;
\item \emph{Random Pixels} - a new image is created based on pixels sampled uniformly from both images;
\item \emph{Random Elements} - a new image is created based on color channels values of the respective pixels sampled uniformly from both images.
\end{itemize}
The effects of patch-wise \emph{Mixed-Example} augmentations are presented in Figure~\ref{fig:mixed_example}.

\emph{Mixed-Example} paper~\cite{mixed_example} additionally proposes several strategies which combine pixel-wise and patch-wise mixing, e.g. a blend of \emph{Vertical Concat} and \emph{Mixup}, where \emph{Mixup} is applied to one of the input images and followed by \emph{Vertical Concat} procedure.

\begin{figure}[!ht]
\includegraphics[width=1\linewidth]{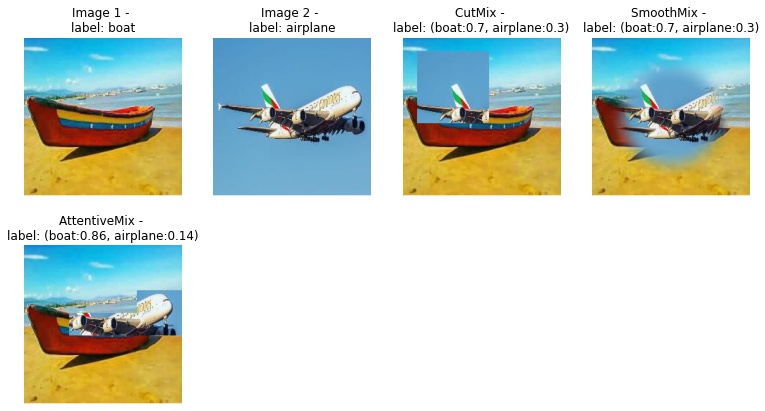}
\caption{
Effects of \emph{CutMix} (which is visually equivalent to \emph{Random Square} variant of \emph{Mixed-Example} method described above), \emph{SmoothMix} and \emph{Attentive CutMix} augmentations.
}
\label{fig:cutmix_and_extensions}
\end{figure}

\begin{figure}[!ht]
\includegraphics[width=1\linewidth]{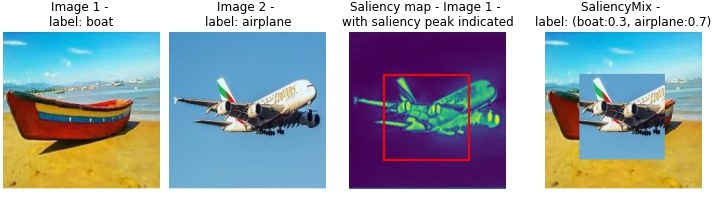}
\caption{
Effects of \emph{SaliencyMix} augmentation method with the corresponding mixed label. From left: two original images, the saliency map with a bounding box around saliency peak region, and the resulting image after \emph{SaliencyMix} application.
}
\label{fig:SaliencyMix}
\end{figure}

Several subsequent papers in this line of research either build on~\cite{smoothmix,attentive_cutmix, saliency_mix} or explore in-depth~\cite{cutmix} one of particular augmentation methods presented in~\cite{mixed_example}. \emph{CutMix}~\cite{cutmix} mixes images by occluding part of one image with a patch extracted from the other image. \emph{SmoothMix}~\cite{smoothmix} and \emph{Attentive CutMix}~\cite{attentive_cutmix} extend \emph{CutMix} by either smoothing sharp edges in the resultant image (\emph{SmoothMix}) or by using a specific way of patch selection (\emph{Attentive CutMix}). \emph{Saliency Mix} \cite{saliency_mix} attempts to achieve the same properties as \emph{Attentive CutMix} but using saliency information. All four above-mentioned methods are discussed in more detail below and the effects of their application are illustrated in Figure~\ref{fig:cutmix_and_extensions} (\emph{CutMix}, \emph{SmoothMix}, \emph{Attentive CutMix}) and Figure~\ref{fig:SaliencyMix} (\emph{Saliency Mix}).

\emph{CutMix}~\cite{cutmix} is motivated by the \emph{Cutout} method (described in section~\ref{sec:erasing_image}), however, unlike \emph{Cutout} which simply clears out part of the image with possible loss of relevant information, \emph{CutMix} introduces new information to the resultant image extracted from the other image it is mixed with. The method is regarded as a strong alternative to~\emph{Mixup}.

A patch size in \emph{CutMix} is proportional to the image size and calculated according to the following equations:
\begin{equation}
\label{eq:cutmix}
\ r_w = W \sqrt{1-\lambda},\ \ \ \ \ \ \ r_h = H \sqrt{1-\lambda}
\end{equation}
where $r_w$ and $r_h$ are the width and the height of the patch, respectively, $W$ and $H$ are the width and the height of original images and $\lambda$ is the mixing ratio.
According to~\cite{cutmix} the method should be applied in the input layer (to raw images) and its application to hidden layer representations degrades performance.

\emph{SmoothMix}~\cite{smoothmix} is a direct extension of \emph{CutMix} that attempts to avoid \emph{sharp edges}, i.e. unnatural changes in pixel values around the implemented patch.
\emph{SmoothMix} creates a new observation according to eqs. (\ref{eq:mixing_images_x})-(\ref{eq:mixing_images_y}) with a specific form of matrix $B$ which has $0$s at the edges of a patch and $1$ in its center. $B$ implements a smooth transition from the edge pixels towards the patch center, i.e. its elements change gradually from $0$ to $1$ along with moving further away from the edges and approaching the central region of the patch. Both, circular and rectangular patches were tested in~\cite{smoothmix} with the final preference for a circular one.

\emph{Attentive CutMix}~\cite{attentive_cutmix} is another extension of \emph{CutMix}. The method divides both images into patches (typically a $7\times 7$ grid) and in one of the images selects some number of patches that activate the most the respective class-related output.
Next, the method pastes those most-representative patches onto the other image in the respective places. The label of the new image is calculated based on the proportion of the pixels corresponding to each of the two source images. \emph{Attentive CutMix} addresses one of the problems with \emph{CutMix}, i.e. susceptibility to copying a non-representative part of the image (e.g. background) and consequently producing an example with a noisy label.  \emph{Attentive CutMix}, on the contrary, puts special attention to selection of the meaningful and class-representative patches to be transferred to the resultant image. One disadvantage of \emph{Attentive CutMix}, compared to other discussed methods, is the requirement of having a pretrained feature extractor that is used for selecting the most relevant patches.

The idea of selecting the most representative patches is also implemented in \emph{SaliencyMix} method~\cite{saliency_mix} which uses saliency information to find the most representative pixels and then, similarly to \emph{CutMix}, selects a patch of a size determined by $\lambda$. In \emph{SaliencyMix} the patch is either centered on the most salient pixel (if the entire patch fits within the image) or keeps this pixel within the patched area (otherwise). The method follows equations (\ref{eq:mixing_images_x})-(\ref{eq:mixing_images_y}) and is implemented by setting the values of matrix $B$ to $1$ within the patch area (selected based on the saliency peak region) and to $0$ elsewhere. Figure~\ref{fig:SaliencyMix} presents an example of saliency map and the effect of \emph{SaliencyMix} application.

In~\cite{saliency_mix} four saliency calculation methods were considered, relaying either on statistical approaches~\cite{statistical_saliency,statistical_saliency_2,statistical_saliency_3} or a learning based approach~\cite{learning_based_saliency}, and eventually decided to use the statistical saliency method from~\cite{statistical_saliency} as it offered slightly better classification accuracy (with \emph{SaliencyMix}) and, additionally, is invariant to image size, as opposed to learning based methods which are limited by the input size of the architecture used.
Various placements of the salient patch of the image were also tested, including placing the patch on salient / non-salient / corresponding part of the other image, with a conclusion that the choice of a corresponding place is the best option in terms of regularization.
%

The following two sections consider methods that deviate from certain baseline assumptions that were commonly considered in the methods discussed so far. Section \ref{sec:Mixup-puzzle} presents methods that does not follow equation (\ref{eq:mixing_images_x})~\cite{puzzle_mix, snap_mix} and section \ref{sec:other-than-two} discusses the methods that use the number of images other than two to produce a mixed sample~\cite{RICAP, aug_mix, style_augmentation, mix_style, co_mixup}.

\subsection{Beyond equation (\ref{eq:mixing_images_x})}
\label{sec:Mixup-puzzle}

The next method cannot be definitively classified as either pixel-wise or patch-wise.
\emph{Puzzle Mix}~\cite{puzzle_mix} is motivated by the same principles, as the previously-described \emph{Attentive CutMix} and \emph{Saliency Mix} methods,
namely ensuring that the resultant image contains patches that are relevant for both classes. Similarly to \emph{Saliency Mix} this goal is achieved through utilization of the saliency information~\cite{saliency} to determine the most important patches in both images, though \emph{Puzzle Mix} additionally performs an optimal transportation of these most salient parts that would maximize their exposure in the resultant augmented image. This results in placing a relevant patch in the target image not in the place that corresponds to its location in the source image but in a destination that contains low saliency features compared to the rest of the target image. The overall goal is achieved through jointly seeking an optimal mixing mask and an optimal placements for the patches coming from both source images in the augmented image.
Consequently, instead of (\ref{eq:mixing_images_x}) the following equation is used for creating a mixed sample in \emph{Puzzle Mix}:
\begin{equation}
\tilde{x} = (1-B) \odot \prod_{1} x_1 + B \odot \prod_{2} x_2
\end{equation}
where $B$ is a mask with $b_{ij}\in [0,1]$ and $\prod_{1}, \prod_{2}$ are $n \times n$ transportation matrices. $\prod_{i}, i=1,2$ encodes the placement of patches extracted from the source images $x_i, i=1,2$ in the augmented sample. The mixing ratio $\lambda$ is defined as $\lambda = \frac{1}{n} \sum_{ij} b_{ij}$.

Since there are two goals in \emph{Puzzle Mix} the problem is solved via iterative alternating of the following two phases: (1) minimization of $B$ and (2) simultaneous optimization of $\prod_{1}$ and $\prod_{2}$. Although certain improvements of both CPU and GPU implementations, as well as down-sampling strategies for mask optimization are proposed in~\cite{puzzle_mix}, the method still requires substantially more computational resources than previous DA techniques, e.g. \emph{Mixup} or \emph{CutMix}.
On the other hand, whenever \emph{Puzzle Mix} is compared with other methods in the same experiment setup, it achieves higher accuracy. Numerical results are presented in section \ref{sec:erasing_mixing_evaluation}.

\begin{figure}[!ht]
\includegraphics[width=1\linewidth]{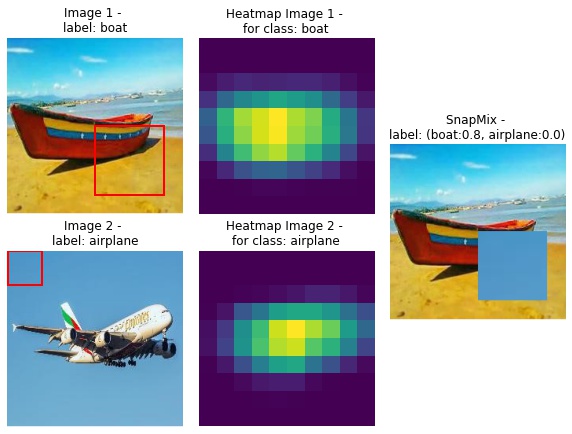}
\caption{
Effects of \emph{SnapMix} augmentation with the corresponding mixed label. Left: original images with randomly selected patches of different sizes. Middle: heatmaps presenting the output of CAM for the respective class. Right: a resulting image after \emph{SnapMix} application. Observe that elements of the label vector (right figure) do not have to sum up to $1$.
}
\label{fig:SnapMix}
\end{figure}

The other method that departs from eq. (\ref{eq:mixing_images_x}) is \emph{SnapMix}~\cite{snap_mix}. The method is dedicated to fine-grained image classification problem, in which classes differentiate by details only like, for instance, in CUB \cite{birds}, Stanford Cars \cite{cars} or FGVC Aircrafts \cite{aircrafts} datasets. Consequently, \emph{SnapMix} assumes that
visual features indicative for a given class may take a small part of the image, in which case traditional methods like \emph{Mixup} \cite{mixup} or \emph{CutMix}~\cite{cutmix} would not be effective. An example of \emph{SnapMix} application is depicted in Figure~\ref{fig:SnapMix}.

\emph{SnapMix} modifies \emph{CutMix} in the two following aspects: the way of label vector calculation for the mixed sample, and independent selection of the size and placement of the patch in each of the two input images (the patch is not automatically pasted in the corresponding location of the target image).

\emph{SnapMix} method can be expressed as follows:

\begin{equation}
\label{eq:snapmix_x}
\tilde{x} = (I-B_{\lambda^{1}}) \odot x_1 + T_{\theta}(B_{\lambda^{2}} \odot x_2)
\end{equation}

where $B_{\lambda^{1}}$ and $B_{\lambda^{2}}$ are two binary masks containing random box regions with the area ratio $\lambda^{1}$ and $\lambda^{2}$, respectively, and $T_{\theta}$ is a function that maps the patch from $x_2$ onto the patch in $x_1$.

In order to calculate the label for the augmented image (\ref{eq:snapmix_x}), the class activation map (CAM)~\cite{CAM} is used. CAM is a transformation applied on top of the last convolutional layer of the network so as to point locations of the class-discriminative features. The output of CAM is normalized to get Semantic Percent Map (SPM) using eq. (\ref{eq:snapmix_spm}):

\begin{equation}
\label{eq:snapmix_spm}
\ S(x_{i}) = \frac{CAM(x_{i})}{sum(CAM(x_{i}))}
\end{equation}

The label of the image created according to (\ref{eq:snapmix_x}) is finally calculated as follows:

\begin{equation}
\label{eq:snapmix_y}
\ p_{1} = 1 - sum(B_{\lambda^{1}} \odot S(x_{1})), \ \ \ \ \ p_{2} = sum(B_{\lambda^{2}} \odot S(x_{2}))
\end{equation}

where $p_{1}, p_{2} \in [0,1]$ are partial labels assigned to classes corresponding to the class of images 1 and 2, respectively.

An interesting aspect of the method is the lack of a constraint that partial labels should sum up to 1. This way it is possible to indicate in the label that as a result of the operation one image has become relatively more / less relevant than previously (for instance, when the patch masked a discriminative feature in that image).

Another relevant \emph{SnapMix} feature is the ability to offer more effective augmentations as the training process progresses. This is due to the fact that CAM works on top of the classifier, and therefore, becomes more accurate along with the classification improvement, making the resulting augmented samples more effective.

\subsection{Mixing other than 2 images}
\label{sec:other-than-two}

Until now all described methods used 2 images to create a mixed sample. The next group of approaches~\cite{RICAP,aug_mix,style_augmentation,mix_style,co_mixup} veered off in the other direction. \emph{RICAP}~\cite{RICAP} combines 4 images in a pixel-wise fashion while \emph{AugMix}~\cite{aug_mix} and \emph{Style Augmentation}~\cite{style_augmentation} use one image in the augmentation process. \emph{MixStyle}~\cite{mix_style} on the other hand, relies only on the per channel statistics, like mean and standard deviation calculated from several hidden layers of the second image, disregarding the input pixels. A potential breakthrough is proposed in \emph{Co-mixup}~\cite{co_mixup}, one of the youngest methods, that performs mixing of the entire mini-batch rather than each pair separately.
All five methods are briefly discussed below.

\emph{RICAP} augmentation process consists of the three following steps. Firstly, four images are randomly selected - one for each of the four parts of the new image: upper left, upper right, lower left and lower right. Secondly, the images are randomly cropped with respect to a boundary position point, i.e. a point that determines the area a given image will occupy in the newly created augmented image. The boundary point $BP=(w,h)$ is calculated using  the following equations:
\begin{equation}
\begin{aligned}
&w = round(w^{\prime}I_x),\ \ \ &h = round(h^{\prime}I_y) \\
&w^{\prime} \sim Beta(\beta, \beta),\ \ \ &h^{\prime} \sim Beta(\beta, \beta)
\end{aligned}
\end{equation}
where $I_x$ and $I_y$ are width and height of the original image, respectively.

\begin{figure}[!ht]
\includegraphics[width=1\linewidth]{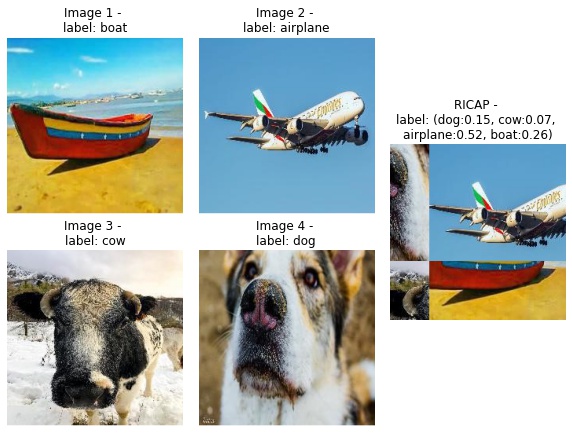}
\caption{
Effects of \emph{RICAP} augmentation with the corresponding mixed label. Left: original images: a boat, an airplane, a cow and a dog. Right: a resulting augmented image.
}
\label{fig:RICAP}
\end{figure}

For selecting the BP coordinates the ratios $w^{\prime}$ and $h^{\prime}$ ale sampled from $Beta$ distribution parameterized by $\beta \in (0,\infty)$.
Based on $BP$ the cropping sizes $[w_i, h_i], i=1,\ldots,4$ are calculated, i.e. $w_1=w_3=w$, $w_2=w_4=I_x-w$, $h_1=h_3=h$ and $h_2=h_4=I_y-h$.
The new class label is calculated as a combination of four original one-hot encoded labels weighted by the relative size of the area assigned to each of the four images in the newly created image. An example of \emph{RICAP} application is presented in Figure~\ref{fig:RICAP}.

The next two methods apply mixing procedure to the original image and its transformed version(s). \emph{AugMix}~\cite{aug_mix} starts off with application of traditional augmentation operations (e.g. translation, shear, rotation, etc.) to the input image.
More precisely a set of $k$ ($k=3$ by default) augmentation chains, each composed of 3 operations, are selected and applied independently to $k$ copies of the original image. The resulting $k$ images are mixed together linearly with random weights. Next this augmented image is mixed linearly with the original non-augmented image. \emph{AugMix} creates 2 images in the above-described manner.

In order to enforce a consistent embedding across diverse augmentations of the same input image a consistency loss in the form of Jensen-Shannon (JS) divergence~\cite{Jensen_Shannon}
is included in the model loss function (\ref{eqn:aug_mix_js}):
\begin{equation}
\label{eqn:aug_mix_js}
\ L(p_{orig},y) + \lambda JS(p_{orig};p_{aug1};p_{aug2})
\end{equation}
\begin{equation}
\ JS(p_{orig};p_{aug1};p_{aug2}) = \frac{1}{3}(KL[p_{orig}||M] + KL[p_{aug1}||M] + KL[p_{aug2}||M])
\end{equation}
where $M = (p_{orig} + p_{aug1} + p_{aug2})/3$ and $p_{orig}, p_{aug1}, p_{aug2}$ are posterior distributions of the original sample and its augmented variants.

\begin{figure}[!ht]
\includegraphics[width=1\linewidth]{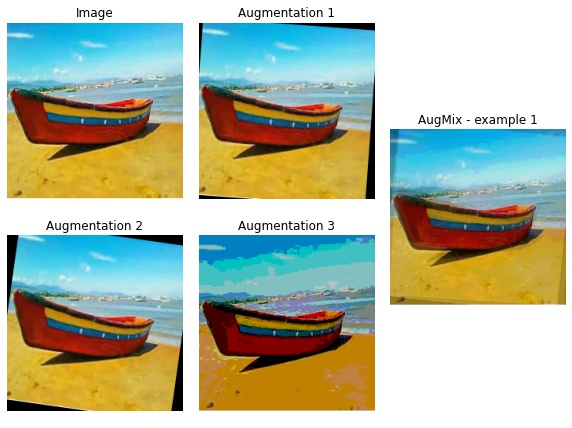}
\caption{
Effects of \emph{AugMix} method with $k=3$. Left and middle: original image and its augmentations (translate, rotate and posterize). Right: the image after \emph{AugMix} application (mixing augmentations 1-3).
}
\label{fig:augmix}
\end{figure}

Additionally, two alternative approaches were tested in~\cite{aug_mix} that use respectively 1 or 3 augmented samples to calculate the consistency loss. The former setup performed the worse whereas the gains from the latter one were marginal and did not justify the related computational complexity increasing.
The results of \emph{AugMix} application are visualized in Figure~\ref{fig:augmix}.

The other single-image method, \emph{Style Augmentation}~\cite{style_augmentation}, randomly changes the texture, contrast and colors of the image while preserving the objects' shapes and the semantic meaning of the image. This is achieved by using two neural networks: $P$ and $T$, described in detail in~\cite{style_transfer}. Network $P$ is trained on image styles and provides style embeddings which are subsequently used by network $T$ that performs style transfer operation.

In practice, for the sake of computational efficiency, \emph{Style Augmentation} does not operationally use network $P$ to provide a style embedding from a randomly sampled image, but instead randomly samples a style embedding vector directly from a distribution with mean and covariance matching those used for training network $P$.
Such a random sampling (from appropriate distribution) simulates the process of choosing an image from the training data set and calculating its style embedding, without computational load related to its actual calculation.

\begin{figure}[!ht]
\includegraphics[width=1\linewidth]{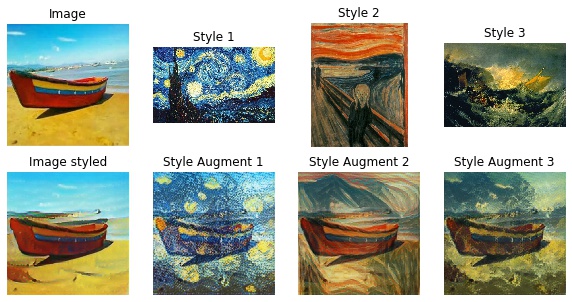}
\caption{
Effects of \emph{Style Augmentation} method with 3 arbitrary style images. Upper row: original image and 3 style images. Lower row: original sample after application of the style extracted from the same (original) sample, and the 3 augmented samples (Style Augment 1,2,3) obtained by applying style vectors in the form of a linear combination of the original image style (Image styled) and the respective image style vector (Style 1,2,3). The results are presented for style mixing parameter $\alpha=0.75$.
}
\label{fig:style_augmentation}
\end{figure}

Additionally, in order to control the strength of the augmentation process, a randomly sampled embedding vector is linearly mixed with a style embedding of the input image. The effective embedding vector $z$ used for style transfer is therefore defined in the following way:
\begin{equation}
\ z = \alpha \mathcal{N}(\mu, \Sigma) + (1 - \alpha) P(c)
\end{equation}
where $P$ is the style predictor network, $c$ is the input image, and $\mu, \Sigma$ are the mean vector and covariance matrix of embeddings from the training set.
An example of \emph{Style Augmentation} application is presented in Figure~\ref{fig:style_augmentation}.

Another method, \emph{MixStyle}~\cite{mix_style}, attempts to achieve a similar goal as \emph{Style Augmentation} but using a different approach to style transfer. The method is dedicated to the problem of Domain Generalization (DG), i.e. construction of classifiers robust to domain shift, able to generalize to unseen domains. To this end \emph{MixStyle} does
not mix pixels but instance-level feature statistics of the two images. These statistics extracted from early layers of a CNN are calculated in the following manner:

\begin{equation}
\label{eq:mixstyle_param1}
\ \gamma_{mix} = \lambda \sigma (x_1) + (1-\lambda)\sigma (x_2)
\end{equation}
\begin{equation}
\label{eq:mixstyle_param2}
\ \beta_{mix} = \lambda \mu (x_1) + (1-\lambda)\mu (x_2)
\end{equation}

where
$\gamma_{mix}, \beta_{mix}$
are the mix feature statistics, $\lambda$ is the instance-wise weight sampled from $Beta(\beta, \beta)$ distribution parameterized with $\beta \in (0,\infty)$ and $\mu(x_i)$, $\sigma(x_i)$, $i=1,2$ are mean and standard deviations computed for the set of all elements in each color channel, in each instance. Once the mixing statistics are calculated, the mixed sample is defined as:

\begin{equation}
\label{eq:mixstyle_augmented_sample}
\ \tilde{x} = \gamma_{mix} \frac{x_1-\mu(x_1)}{\sigma(x_1)} + \beta_{mix}
\end{equation}

Equation (\ref{eq:mixstyle_augmented_sample}) is inspired by the arbitrary style transfer~\cite{arbitrary_style_transfer} and is applied with probability $0.5$. Otherwise, the instance is not augmented.
It is recommended in~\cite{mix_style} to apply \emph{MixStyle} to multiple lower level layers which yields the best performance. The optimal combination of layers depends on the final task and including the last layer always deteriorates the performance. An interesting observation is that if multiple layers are selected the same sample can be processed with different mixed statistics as the shuffling process that selects pairs of samples is independent in each layer. This property constitutes additional regularization mechanism in \emph{MixStyle}.

The final method in this area, \emph{Co-Mixup}~\cite{co_mixup}, lifts the idea of mixing from the level of pairs of images to the level of entire training mini-batches. The method, similarly to \emph{Saliency Mix}, uses saliency information to generate mixed samples but at the same time looks at the whole mini-batch to encourage diversity among the constructed mixed samples.
Since \emph{Co-Mixup} works on the entire mini-batch equations (\ref{eq:mixing_images_x})-(\ref{eq:mixing_images_y}) are no longer valid and the mixing process is described as follows:

\begin{equation}
\ h(x_B) = (g(z_1 \odot x_B),...,g(z_{m'} \odot x_B))
\end{equation}

where $z_j \in \mathcal{L}^{m \times n}$ for $j = 1,...,m^{\prime}$ and $\mathcal{L}^{m\times n} = \{ \frac{l}{L} | l=0,1,...,L \}$ is a discretized mask of dimensions equal to the number of images in the mini-batch $(m)$ and the number of regions into which the image is divided into $(n)$, for optimization purposes (see below). $x_B$ is the mini-batch and $g: \mathbb{R}^{m \times n} \rightarrow\mathbb{R}^{n} $ returns a column-wise sum of matrix $z_j \odot x_B$ for $j = 1,...,m^{\prime}$. As $z_j$ is a 2D matrix, one can interpret the $k^{th}$ column of $z_j$ ($z_{j,k} \in \mathcal{L}^{m}$) as the mixing ratio for $m$ inputs from a mini-batch at the $k^{th}$ area.

Additionally an optimization procedure is applied as the method aims to maximize the exposed saliency, maintain local smoothness within the images (adjacent locations in the mixed sample are similar to one another) and also encourage diversity among the constructed mixed samples~\cite{co_mixup}.

\section{Comparison of DA methods based on particular properties}
\label{sec:comparison}

In this section the aforementioned mixing methods are compared on the basis of certain aspects which refer to their operational properties, efficacy and computational complexity.
The key characteristic features of the methods are summarized in Table~\ref{tab:mixing_all}.


\subsection{\emph{Where} is the augmentation applied?}
\label{sec:where}

All methods apply the DA procedure in the input layer except three: \emph{Feature Space}, \emph{Manifold Mixup} and \emph{MixStyle} (cf. the first column of Table~\ref{tab:mixing_all}). \emph{Feature Space} uses additional encoder-decoder network to create image embeddings which are then mixed. \emph{Manifold Mixup} is a direct extension of \emph{Mixup} with the mixing mechanism applied to a pool of eligible layers that includes both input and hidden layers. The layer in which mixing is performed is selected randomly. \emph{MixStyle} does not actually mix images but the instance-level feature statistics extracted from early layers of a CNN.


\begin{sidewaystable}
\begin{tabular}{llllllllllllllllc}
\toprule
{} &  \multicolumn{2}{c}{Where} & \multicolumn{3}{c}{How} & \multicolumn{3}{c}{Pixel-wise} &  \multicolumn{2}{c}{Mix labels} & \multicolumn{2}{c}{Loss function} & \multicolumn{3}{c}{Nr. images} & Computational    \\
 & I & H & AM & O & R & Yes & No & Both & Yes & No & CCE & Non-CCE & 3+ & 1 & 2 & complexity\\
Method           &       &        &                     &              &      &        &            &      &      &      &                           &              &      &      &      \\
\midrule
\emph{AdaMixup}         &   \cmark &        &                 \cmark &              &      &    \cmark &            &      &  \cmark &      &                           &          \cmark &      &      &  \cmark & B  \\
\emph{Attentive CutMix} &   \cmark &        &                 \cmark &              &      &        &        \cmark &      &  \cmark &      &                       \cmark &              &      &      &  \cmark & {\color{black}B}  \\
\emph{AugMix}           &   \cmark &        &                     &              &  \cmark &    \cmark &            &      &      &  \cmark &                           &          \cmark &      &  \cmark &   & {\color{black}B}     \\
\emph{BCL}    &   \cmark &        &                     &              &  \cmark &    \cmark &            &      &  \cmark &      &                           &          \cmark &      &      &  \cmark & {\color{black}A}  \\
\emph{Co-Mixup}         &   \cmark &        &                     &          \cmark &      &        &            &  \cmark &  \cmark &      &                       \cmark &              &  \cmark &      &  & {\color{black}B}      \\
\emph{CutMix}           &   \cmark &        &                     &              &  \cmark &        &        \cmark &      &  \cmark &      &                       \cmark &              &      &      &  \cmark & {\color{black}A}  \\
\emph{Feature Space}    &       &    \cmark &                     &              &  \cmark &    \cmark &            &      &      &  \cmark &                       \cmark &              &      &      &  \cmark & {\color{black}B}  \\
\emph{Manifold Mixup}   &       &    \cmark &                     &              &  \cmark &    \cmark &            &      &  \cmark &      &                       \cmark &              &      &      &  \cmark & {\color{black}B}  \\
\emph{Mixed-Example}    &   \cmark &        &                     &              &  \cmark &        &        \cmark &      &  \cmark &      &                       \cmark &              &      &      &  \cmark & {\color{black}A}  \\
\emph{MixStyle}         &       &    \cmark &                     &              &  \cmark &    \cmark &            &      &      &  \cmark &                       \cmark &              & &   & \cmark  & {\color{black}B}     \\
\emph{Mixup}            &   \cmark &        &                     &              &  \cmark &    \cmark &            &      &  \cmark &      &                       \cmark &              &      &      &  \cmark & {\color{black}A}  \\
\emph{Puzzle Mix}       &   \cmark &        &                     &          \cmark &      &        &            &  \cmark &  \cmark &      &                       \cmark &              &      &      &  \cmark & {\color{black}B}  \\
\emph{RICAP}            &   \cmark &        &                     &              &  \cmark &        &        \cmark &      &  \cmark &      &                       \cmark &              &  \cmark &      &  & {\color{black}A}    \\
\emph{Saliency Mix}     &   \cmark &        &                 \cmark &              &      &        &        \cmark &      &  \cmark &      &                       \cmark &              &      &      &  \cmark & {\color{black}B}  \\
\emph{Sample Pairing}   &   \cmark &        &                     &              &  \cmark &    \cmark &            &      &      &  \cmark &                       \cmark &              &      &      &  \cmark & {\color{black}B}  \\
\emph{Smart Augmentation}        &   \cmark &        &                 \cmark &              &      &        &            &  \cmark &      &  \cmark &                           &          \cmark &      &      &  \cmark & {\color{black}B}  \\
\emph{SmoothMix}        &   \cmark &        &                     &              &  \cmark &        &            &  \cmark &  \cmark &      &                       \cmark &              &      &      &  \cmark & {\color{black}A}  \\
\emph{SnapMix}          &   \cmark &        &                     &              &  \cmark &        &        \cmark &      &  \cmark &      &                       \cmark &              &      &      &  \cmark & {\color{black}B}  \\
\emph{Style Augmentation}        &   \cmark &        &                 \cmark &              &      &    \cmark &            &      &      &  \cmark &                       \cmark &              &      &  \cmark & & {\color{black}B}       \\
\bottomrule
\end{tabular}
\caption{
Comparison of data augmentation techniques with respect to particular baseline properties: \emph{where}, \emph{how} and \emph{in which form} the augmentation is applied, whether or not it \emph{mixes labels} or utilizes a \emph{specific loss function}, \emph{how many images} take part in a single augmentation and what is the \emph{computational complexity} of the method. I - input layer, H - hidden layer, AM - auxiliary mechanism (either network or other), O - optimization, R - rule, CEE - Categorical Cross Entropy, {\color{black}A - there is no significant computational overhead, B - requires either special training process, multiple evaluations or an auxiliary component that incurs additional computational cost.}}
\label{tab:mixing_all}
\end{sidewaystable}

\subsection{\emph{How} is the augmentation applied?}
\label{sec:How}

Generally, there are three ways in which the DA techniques that we discuss are applied. The first one, which includes \emph{Smart Augmentation}, \emph{AdaMixup}, \emph{Attentive CutMix}, \emph{Saliency Mix} and \emph{Style Augmentation}, is to rely on an auxiliary mechanism to aid the augmentation process. In the case of \emph{Smart Augmentation} a CNN based augmentation model precedes the classification network and is trained together to optimize the effects of augmentation for a given problem. In \emph{AdaMixup} an additional network is used to predict whether or not mixing a given pair of images will result in manifold intrusion. In \emph{Attentive CutMix} a CNN is utilized to detect the regions that are most representative for a given class, whereas \emph{Saliency Mix} achieves the same objective by using the saliency information. \emph{Style Augmentation} applies a style transfer network to change visual attributes of an image without changing its meaning.

In the second group, encompassing all but two of the remaining methods, an augmentation technique follows
a certain rule or procedure which is applied in some randomized way. These rules differ in terms of complexity and the number of hyperparameters, however the main dividing line is between methods whose application results in pixel-wise mixing and those which lead to patch-wise mixing. This aspect is further discussed in the following section.

The third set of methods is represented by \emph{Puzzle Mix} and \emph{Co-Mixup}. The former attempts to find an optimal way of mixing a given pair of images by solving an optimization task with two objectives. The first objective is to find an optimal mask, i.e. decide how much of the image should be concealed with the other image in a given image region. The second goal is to find optimal
locations for the patches extracted from original images in the mixed image, in order to maximize the exposed saliency.

The other method, \emph{Co-Mixup}, works on the whole mini-batch and attempts to optimize the exposed saliency, as well as encourage diversity among created (augmented) samples.

Both methods bear some resemblance to \emph{Saliency Mix} yet the key difference is that \emph{Saliency Mix} uses saliency directly to make a decision while \emph{Puzzle Mix} and \emph{Co-Mixup} utilize this information to steer the optimization process.
A summary of how is augmentation applied in each method is presented in the second column of Table~\ref{tab:mixing_all}.

\subsection{Is the resulting mix \emph{pixel-wise or patch-wise}?}
\label{sec:linear}

Probably the most fundamental division of mixing augmentation methods is whether they output pixel-wise or patch-wise combination of the images (cf. the third column of Table~\ref{tab:mixing_all}).
A distinction between pixels-wise and patch-wise mixing was clarified
in section \ref{sec:notation}. Please recall that the term \emph{pixel-wise} is used for mixing images using pixel-wise weighted average and \emph{patch-wise} for mixing images spatially, by means of taking patches from original images and joining them together.

Pixel-wise methods are specifically well suited for dealing with corrupted images and adversarial attacks whereas patch-wise augmentations are useful in the context of partial occlusion and weakly supervised object localization. A detailed performance comparison of both types of methods is presented in section \ref{sec:erasing_mixing_evaluation}.

\emph{Smart Augmentation}, \emph{SmoothMix}, \emph{Puzzle Mix} and \emph{Co-Mixup} are the exceptions from this division as they combine aspects of both types of mixing. \emph{Smart Augmentation} uses an auxiliary network to learn the optimal way of mixing and the resulting approach can be either pixel-wise or patch-wise.
\emph{SmoothMix} constructs a graded mixing mask in a way that at the center of the mask and at the edges of the image the resulting mixing is patch-wise as it uses just one image. As we transition from the center of the mask to the edge, mixing becomes pixel-wise as the mask gradually changes its values from $1$ to $0$.
\emph{Puzzle Mix} also exhibits properties of both pixel-wise and patch-wise approaches as it is steered by the saliency information which can result in the patch containing just one image for salient patches and a mixture of two images for less salient patches.
\emph{Co-Mixup} is also guided by saliency information and can exhibit both pixel-wise and patch-wise properties in certain patches of the augmented image.

\subsection{Does the augmentation technique \emph{mix labels}?}
\label{sec:mix_labels}

Label smoothing has proven to be a very effective regularization technique in the presence of label noise~\cite{label_smoothing, DL_book}. Maximizing $\log{p(y|x)}$ when $y$ actually represents an incorrect label can be harmful and one way to prevent this is explicit modeling of label noise. Label smoothing regularizes the model by replacing the $0$ and $1$ targets with the targets $\epsilon / (k-1)$ and $1-\epsilon$, where $k$ is the number of output classes.

Mixing techniques that combine images from various classes together with mixing their corresponding labels also benefit from this regularization property.
The majority of DA techniques mix labels. The ones that do not could be divided in 3 distinctive groups: early methods, methods designed to mix samples from the same class only,
and methods that need just one image to work.

\emph{SamplePairing}, the only representative of early methods that we consider, was designed in a way to assign just one label (from one of the images being mixed) to the augmented image. This resulted in a complicated training procedure that requires the DA to be alternately turned on and off some number of times, and this idea was not continued in subsequent research.

The methods designed and tested only in the context of mixing images from the same class are \emph{Feature Space} and \emph{Smart Augmentation}.

The last genre encompasses \emph{AugMix}, \emph{Style Augmentation} and \emph{MixStyle} which all use the content of just one image during the mixing process, hence there is no room for label mixing. \emph{AugMix} mixes an image with its versions processed with traditional (label preserving) techniques and \emph{Smart Augmentation} mixes an image with its own version transformed using the style transfer algorithm.
In \emph{MixStyle} the second image is technically required however the method does not refer to its pixel-based (raw) representation but considers the instance-level feature statistics that correspond to visual features.

A taxonomy of the methods based on the label mixing property is summarized in the fourth column of Table~\ref{tab:mixing_all}.

\subsection{Does the augmentation technique use standard \emph{loss function}?}
\label{sec:loss_function}

Most of DA techniques described in this survey
use standard loss function during training, as presented in the fifth column of Table~\ref{tab:mixing_all}. There are four exceptions:
\emph{AdaMixup} introduces a separate element in the loss function (\ref{eqn:ada_mixup}) responsible for penalizing manifold intrusion (a situation when an augmented sample corresponds to a class different from any of the two image classes).
\emph{AugMix} extends the standard loss function to ensure that augmented samples are as close to original ones as possible (\ref{eqn:aug_mix_js}).
\emph{Between-Class learning} employs KL-divergence without specifying reasons other than better experimental outcomes~\cite{between_class}.
Finally, \emph{Smart Augmentation} uses an auxiliary Augmentor network whose error is added to the standard loss during training.

\subsection{\emph{How many images} are used to create an augmented image?}
\label{sec:nr_images}

Amongst discussed methods all but four use 2 images to obtain an augmented sample (see the second to last column of Table~\ref{tab:mixing_all}). Out of those four methods, two use more than 2 images: \emph{RICAP}, which mixes four images in a patch-wise fashion by cropping them and afterwards patching into one image and \emph{Co-Mixup} which works with the entire training mini-batch at once. \emph{AugMix} and \emph{Style Augmentation} mix an image with a transformed version of itself. 


\subsection{What is the \emph{computational complexity} of the augmentation technique?}
\label{sec:computational_complexity}

The two key factors contributing to the computational complexity of the considered methods are: (1) whether an augmentation uses a simple rule or a relies on more complex approach, as well as, (2) whether there are any additional components or specific training process that is required for the method to work. Based on the above, the methods can be roughly divided into the two following groups {\color{black}(cf. the last column of Table~\ref{tab:mixing_all})}.
\begin{itemize}
    \item Group A - methods that do not incur any significant computational overhead, as they mix images using simple, often randomized, rules, i.e. \emph{RICAP}~\cite{RICAP}, \emph{Mixup}~\cite{mixup}, \emph{Between-Class learning}~\cite{between_class}, \emph{Mixed-Example}~\cite{mixed_example},  \emph{CutMix}~\cite{cutmix}, \emph{SmoothMix}~\cite{smoothmix}, \emph{Cutout}~\cite{cutout}, \emph{Random Erasing}~\cite{random_erasing} and \emph{Patch Gaussian}~\cite{patch_gaussian}.
    \item Group B - methods that require either special training process, or multiple evaluations, or an auxiliary component that incur additional computational cost. 
\end{itemize}
The functional overheads of particular methods related to using the above-mentioned additional components are summarized below.
    \begin{itemize}
        \item \emph{Smart Augmentation}~\cite{smart_augmentation} -  utilizes additional neural network responsible for mixing two images.
        \item \emph{Feature Space} augmentation ~\cite{feature_space} - utilizes additional network for transforming input data into context vectors.
        \item \emph{SamplePairing}~\cite{sample_pairing} - although the augmentation itself does not incur additional computational cost, 
        its training process requires that augmentation be turned on and off alternately which greatly increases the number of training epochs required.
        \item \emph{Manifold Mixup}~\cite{manifold_mixup} - requires the samples to be propagated through the network before mixing them in hidden layers, hence increasing the computational burden in forward step.
        \item \emph{AdaMixup}~\cite{adamixup} - solves two additional tasks (discrimination of intrusions and generation of policy regions) on top of the standard classification; for the latter task 
        an additional neural network is employed
        \item \emph{Style Augmentation}~\cite{style_augmentation} -  requires two additional forward steps through the style transfer network prior to mixing the images.
        \item \emph{Attentive CutMix}~\cite{attentive_cutmix} -  utilizes an auxiliary network to propose important regions that should be included in the mixing process (an additional forward step of that network is required).
        \item \emph{Puzzle Mix}~\cite{puzzle_mix} and \emph{Co-mixup}~\cite{co_mixup} - both require calculation of saliency information and utilize an optimization procedure for optimal mixing that incurs additional computational burden. 
        \item \emph{AugMix}~\cite{aug_mix} - even though applies standard augmentations and simple mixing rule, because of the modified loss function requires more than one forward step to calculate the JS divergence.
        \item \emph{Saliency Mix}~\cite{saliency_mix} requires additional step of saliency calculation
        \item \emph{SnapMix}~\cite{snap_mix} requires two forward steps to calculate CAM which is required in the mixing process.
        \item \emph{MixStyle}~\cite{mix_style} - applies different statistics at each layer which incurs additional computational cost 
        of their calculation.
    \end{itemize}

Generally speaking, it can be roughly estimated that -- depending on the augmentation method -- the use of the above-listed functional components introduces between a few percents to a few hundred percents of computational time overhead compared to the baseline (straightforward augmentation methods).

\section{Experimental evaluation on standard benchmarks}
\label{sec:erasing_mixing_evaluation}
This section summarizes the accuracy results of both erasing and mixing methods on popular benchmark sets. Additionally, the methods are compared based on their relative improvement over baseline results.

We start in section~\ref{sec:results_clean_data} with performance analysis on classification task on clean data using $3$ data sets. Next, the methods are compared on classification of corrupted images in section~\ref{sec:results_corrupted_images}, and in the context of adversarial examples in section~\ref{sec:results_adversarial_images}. The final comparison is made in the problem of weakly supervised object localization and for the partial occlusion
task - section~\ref{sec:results_WSOL_occlusion}.
The next two sections address the possibility of application of analysed augmentation methods to image-related tasks other than classification (section~\ref{sec:different_tasks}) and to modalities other than images (section~\ref{sec:different_modalities}). Section~\ref{sec:robustness} discusses the robustness of augmentation methods to parameter selection.

Please note that quantitative comparisons of the methods presented in this section rely on experiments that compared at least 3 different methods using a common architecture. The only exception from this rule are the tables presenting top results (Tables~\ref{tab:top_10_CIFAR_10},~\ref{tab:top_10_CIFAR_100} and~\ref{tab:top_10_ImageNet}), which are compiled based on all available outcomes.

\subsection{Evaluation on clean data}
\label{sec:results_clean_data}
Quantitative assessment of the methods on clean data is presented on three widely-used benchmarks:
CIFAR-10, CIFAR-100 and ImageNet.

Unless stated otherwise, a reference to baseline result, labeled as ``NO DA'', would mean the outcomes of the same network architecture, the same training conditions and pre-processing of the input data, but with no use of analyzed data augmentation.

\subsubsection{CIFAR-10}
\label{sec:CIFAR-10}

\begin{figure}[!ht]
\includegraphics[width=1\linewidth]{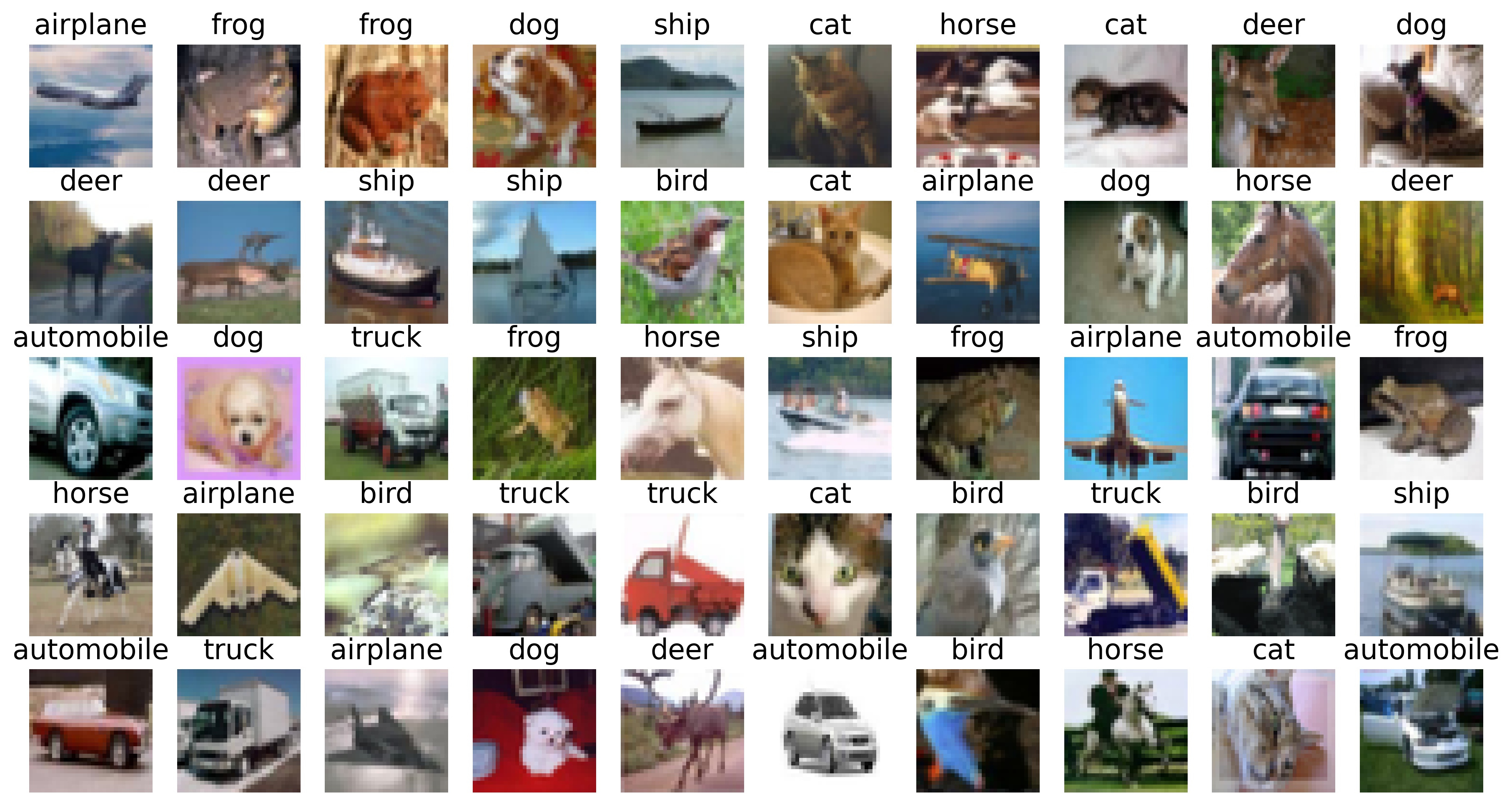}
\caption{
Example images from the CIFAR-10 data set.
}
\label{fig:cifar_10}
\end{figure}

%
%
CIFAR-10 data set~\cite{cifar_10} consists of $60\,000$ color images of size $32x32$ grouped into $10$ classes (airplane, automobile, bird, cat, deer, dog, frog, horse, ship and truck), with $6\,000$ images per class. The set is composed of $50\,000$ training samples and $10\,000$ test ones. There are $1\,000$ images per class in the test subset. Examples of CIFAR-10 images are depicted in Figure~\ref{fig:cifar_10}.

\begin{figure}[!ht]
\includegraphics[width=1\linewidth]{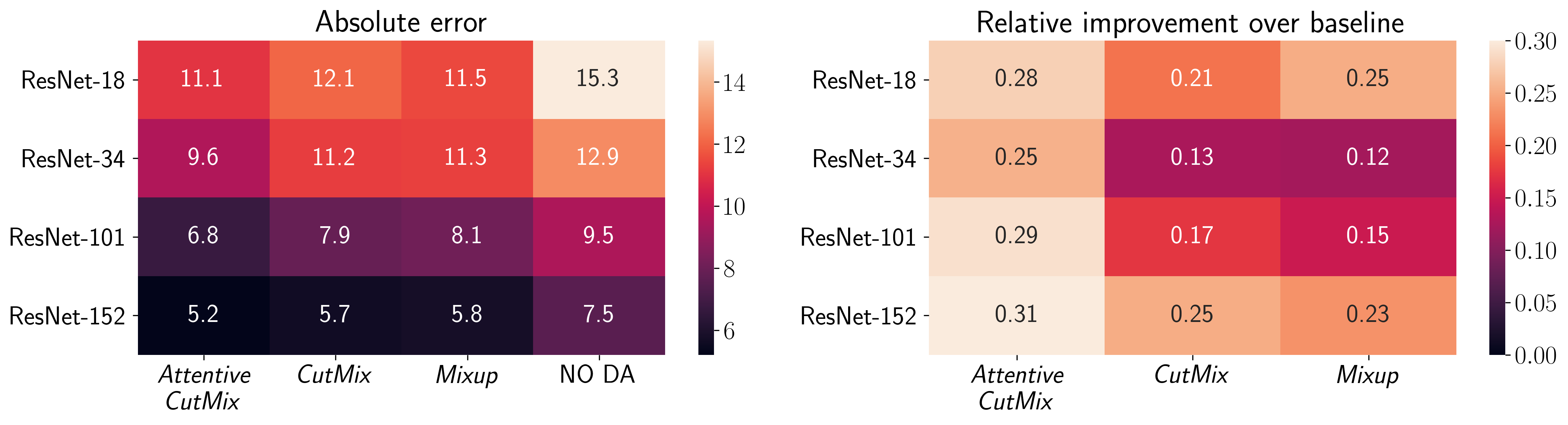}
\includegraphics[width=1\linewidth]{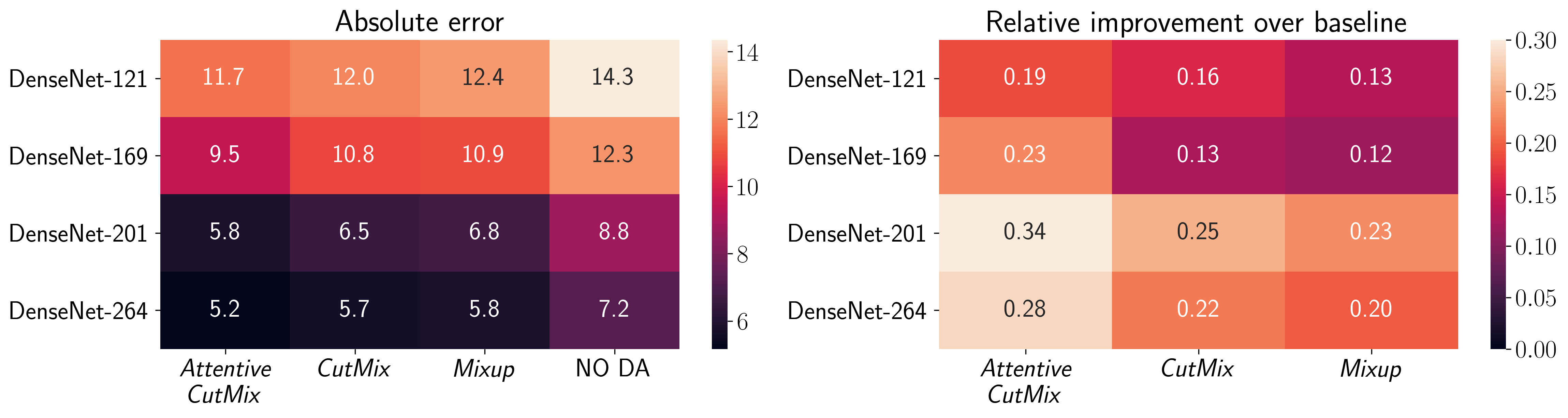}
\includegraphics[width=1\linewidth]{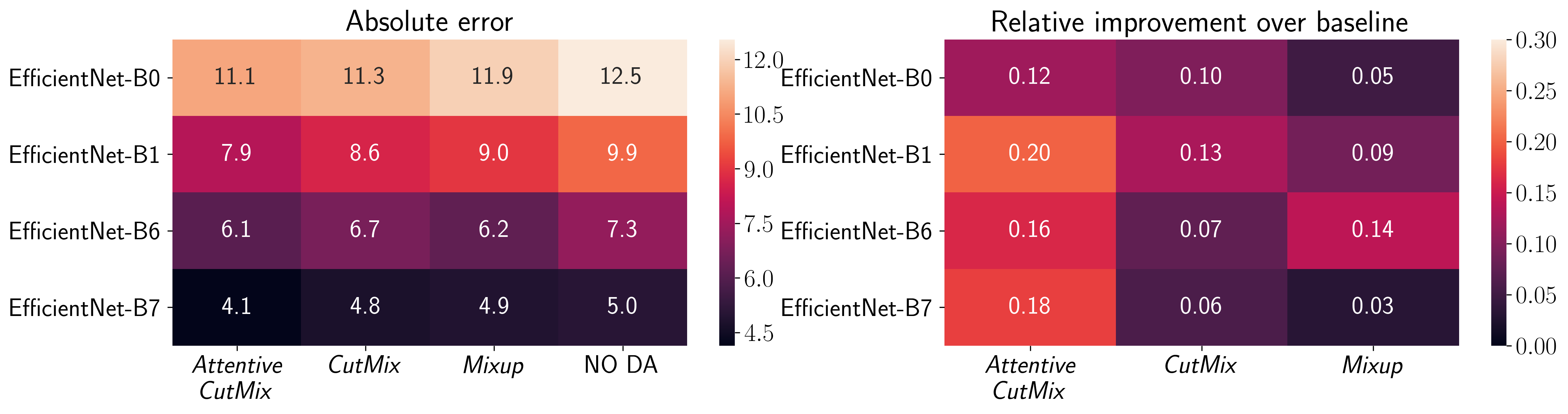}
\caption{
Accuracy results for \emph{Mixup}, \emph{CutMix} and \emph{AttentiveCutMix} on CIFAR-10 reported in~\cite{attentive_cutmix}. Each panel presents a particular type of architecture (ResNet, DenseNet, EfficientNet) with complexity of the network increasing from top to bottom.
Left panels present an absolute error value and the right ones a relative improvement over the baseline (``NO DA'' results).
}
\label{fig:cifar_10_groups}
\end{figure}

Figure~\ref{fig:cifar_10_groups} presents performance results  of \emph{Mixup}, \emph{CutMix} and \emph{AttentiveCutMix} on CIFAR-10 grouped by networks of the same type but with varying complexity. The following architectures are compared: ResNet~\cite{ResNet}, DenseNet~\cite{DenseNet} and EfficientNet~\cite{EfficientNet}, each of them in several realizations. The best overall result is achieved by EfficientNet-B7 and \emph{Attentive CutMix}, with an error of $4.14\%$ (compared to $5.05\%$ baseline).

Looking at the left panels of the figure, one can conclude that DA improves the accuracy regardless of particular architecture type and complexity.
In all three left panels the error decreases when gradually more complex architectures are used (top to bottom) and with an application of more advanced DA methods (right to left).
It is generally assumed that \emph{Mixup}, as the initial method in the area, is the least advanced, followed by \emph{CutMix}, developed based on a certain criticism of \emph{Mixup}, and \emph{AttentiveCutMix} which is an extension of \emph{CutMix}.

Another way to look at the results is from the perspective of a relative improvement over the baseline (the right panels). A general observation is that using DA with more complex types of architectures yields lower relative boost, on average equal to around $22\%$, $21\%$ and $11\%$, respectively for ResNet, DenseNet and EfficientNet. However, the relative effects of DA vary substantially within each architecture type. The highest relative advantage is achieved by the most complex model (ResNet-152) in ResNet group, but for the other two architectures the highest boost is observed for DenseNet-201 and EfficientNet-B1, respectively, which are not the most complex ones.

\begin{figure}[!ht]
\includegraphics[width=1\linewidth]{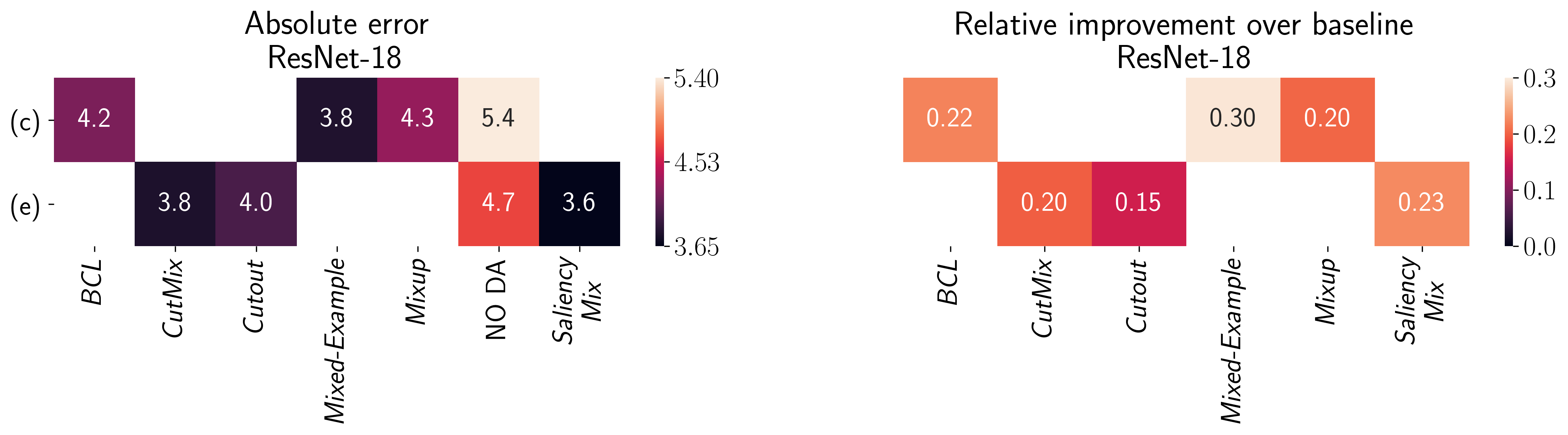}
\includegraphics[width=1\linewidth]{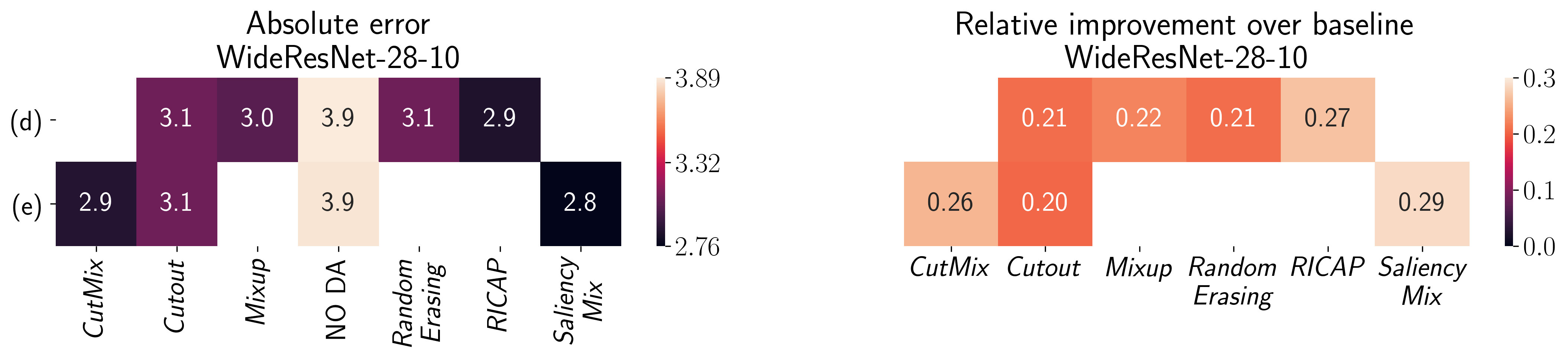}
\includegraphics[width=1\linewidth]{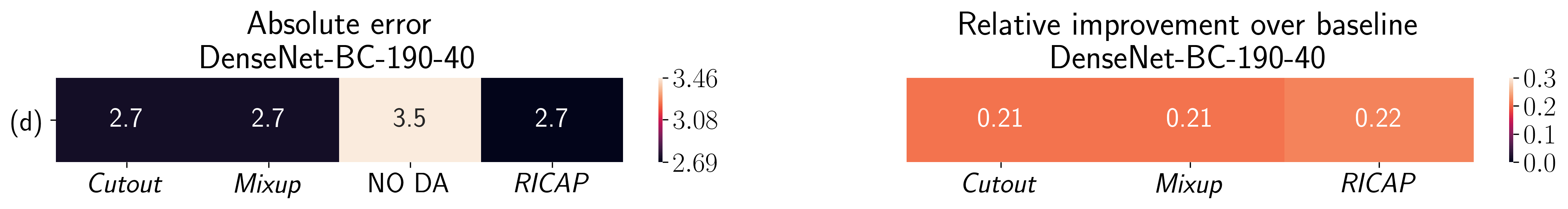}
\includegraphics[width=1\linewidth]{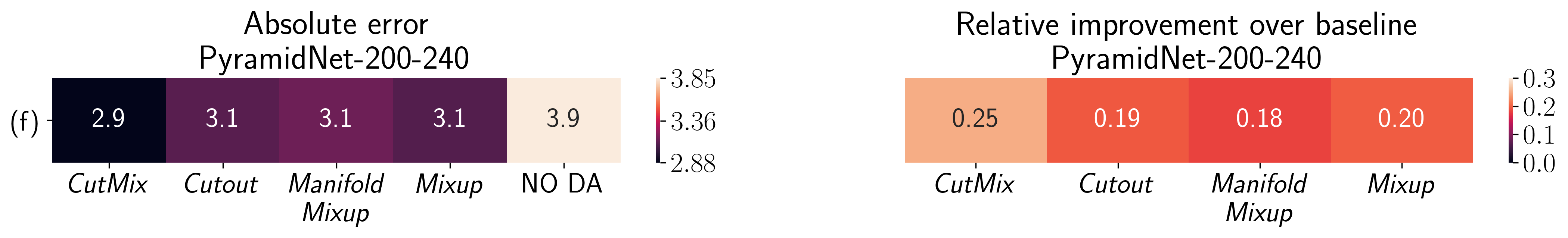}
\includegraphics[width=1\linewidth]{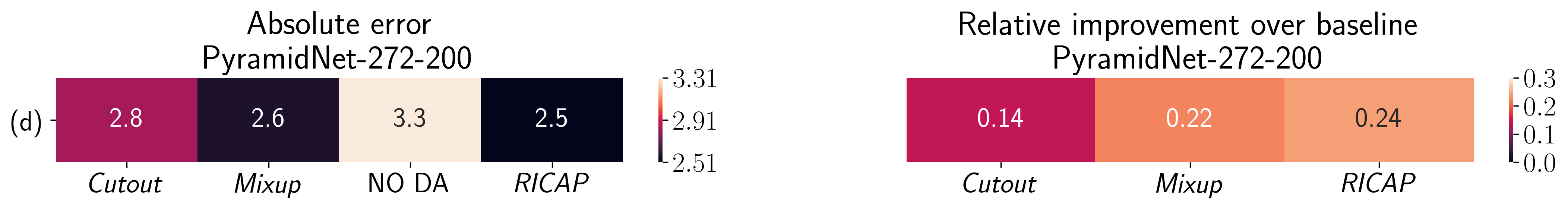}
\includegraphics[width=1\linewidth]{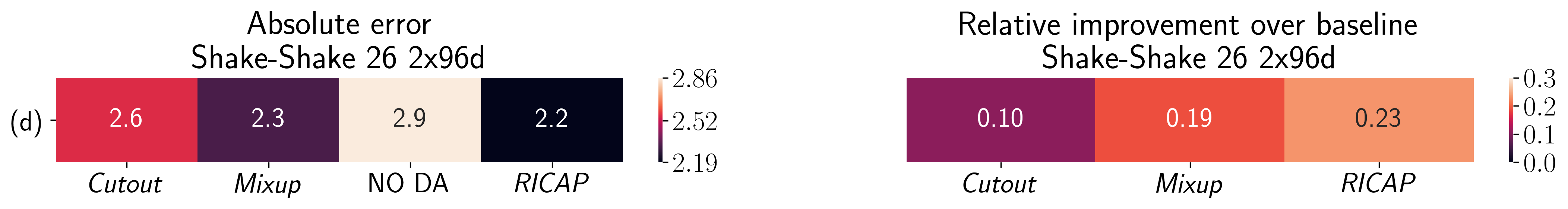}
\caption{
Accuracy results of various augmentation methods
on CIFAR-10 reported in:
(c) - \cite{mixed_example},
(d) - \cite{RICAP},
(e) - \cite{saliency_mix},
(f) - \cite{cutmix}
and grouped around common baselines (particular architectures).
Left panels present absolute error values and the right ones the relative improvements over the baseline.}
\label{fig:cifar_10_single}
\end{figure}

Figure~\ref{fig:cifar_10_single} shows the results of experiments grouped around the same baselines (i.e. particular architectures) for a wider selection of augmentation methods.
The following architectures are considered:
two ResNets of the same complexity trained differently than in Figure~\ref{fig:cifar_10_groups} (coming from two different papers~\cite{mixed_example,saliency_mix} with slightly varying experiment designs), two Wide ResNets~\cite{RICAP,saliency_mix}, a specific case of DenseNet~\cite{RICAP} where the compression factor for both bottleneck and transition layers is smaller than one, two Pyramidal ResNets~\cite{cutmix,RICAP} with different numbers of convolutional layers and different widening factors, as well as Shake-Shake network~\cite{RICAP}.

Generally, similar trends can be observed as in the case of groups of architectures. Simpler models benefit relatively more from data augmentation, however, in terms of absolute figures they still yield higher errors. All in all, every DA technique is able to improve over the baseline, with patch-wise methods achieving slightly better results. For both ResNet-18 architectures best result were accomplished by
patch-wise (\emph{Saliency Mix}) method, for PyramidNet-200-240 by \emph{CutMix} (a patch-wise mixing approach), and for the remaining architectures \emph{RICAP}, which is also a patch-wise mixing method, performed better or equally good than its competitors. The results additionally confirm that erasing methods (\emph{Cutout} and \emph{RandomErasing}) are slightly inferior to the mixing ones.

\begin{table}[!ht]
\begin{tabular}{llrl}
\toprule
              Architecture &         Method &  Error &         Source \\
\midrule
      Shake-Shake 26 2x96d &          \emph{RICAP} &   2.19 &          \cite{RICAP} \\
Shake-Shake Regularization &            \emph{BC+} &   2.26 &            \cite{between_class} \\
      Shake-Shake 26 2x96d &          \emph{Mixup} &   2.32 &          \cite{RICAP} \\
        PyramidNet-272-200 &          \emph{RICAP} &   2.51 &          \cite{RICAP} \\
          PreAct ResNet-34 & \emph{Manifold Mixup} &   2.54 & \cite{manifold_mixup} \\
\bottomrule
\end{tabular}
\caption{Top-$5$ overall best outcomes on CIFAR-10.
}
\label{tab:top_10_CIFAR_10}
\end{table}

The last comparison, presented in Table~\ref{tab:top_10_CIFAR_10}, lists top-$5$ combinations of an architecture and an augmentation method for CIFAR-10 found in the literature. The list is led by two versions of the
Shake-Shake model~\cite{ShakeShake}.

\subsubsection{CIFAR-100}
\label{sec:CIFAR-100}

\begin{figure}[!ht]
\includegraphics[width=1\linewidth]{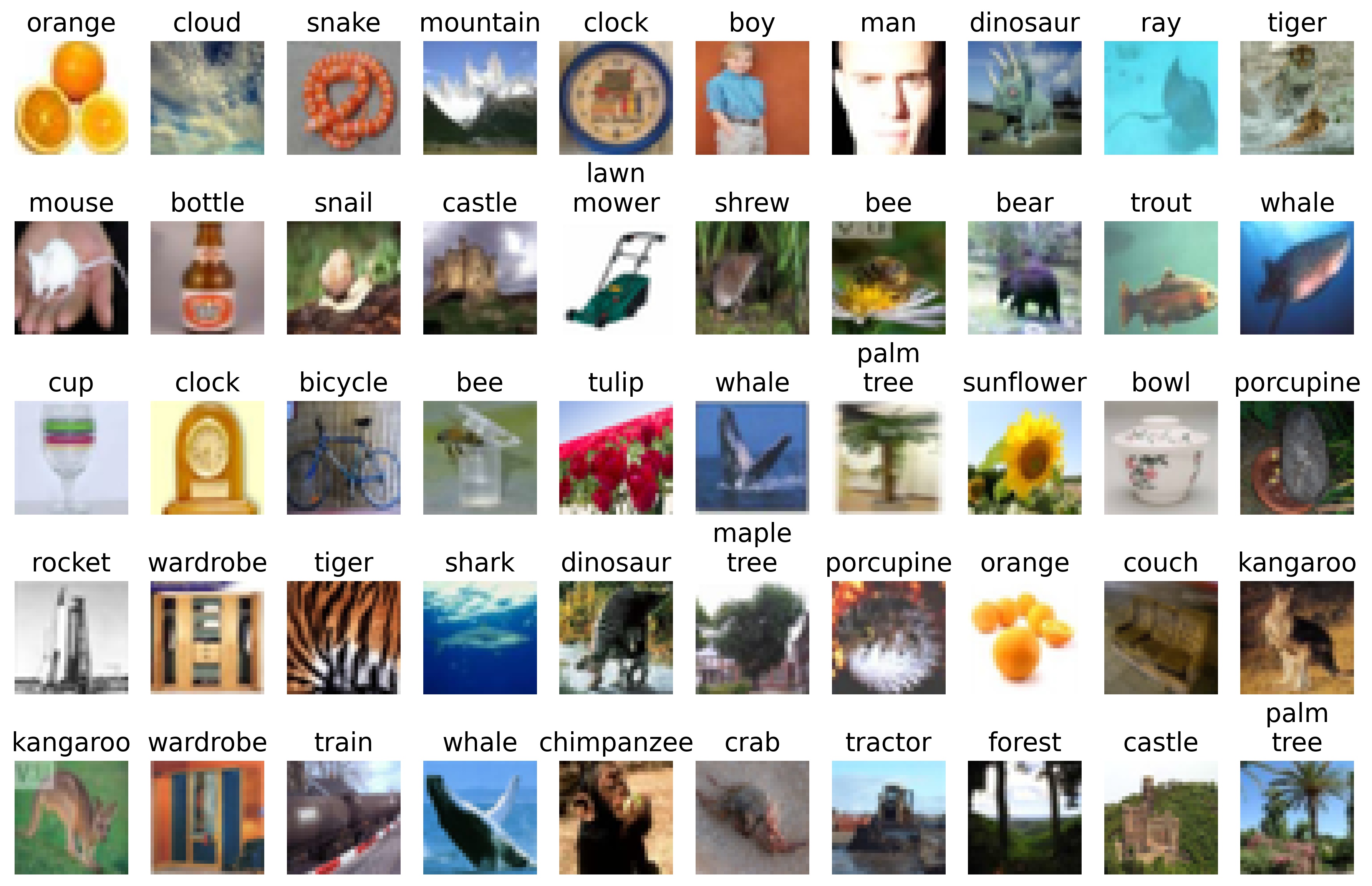}
\caption{
Examples of images from the CIFAR-100 set.
}
\label{fig:cifar_100}
\end{figure}

%
%
CIFAR-100 data set~\cite{cifar_100} is defined in a similar way to CIFAR-10, except that it is more fine grained and has 100 classes. Each class contains 600 images, divided in $500$ training and $100$ test samples. Fifty randomly selected images with corresponding classes are depicted in Figure~\ref{fig:cifar_100}.

\begin{figure}[!ht]
\includegraphics[width=1\linewidth]{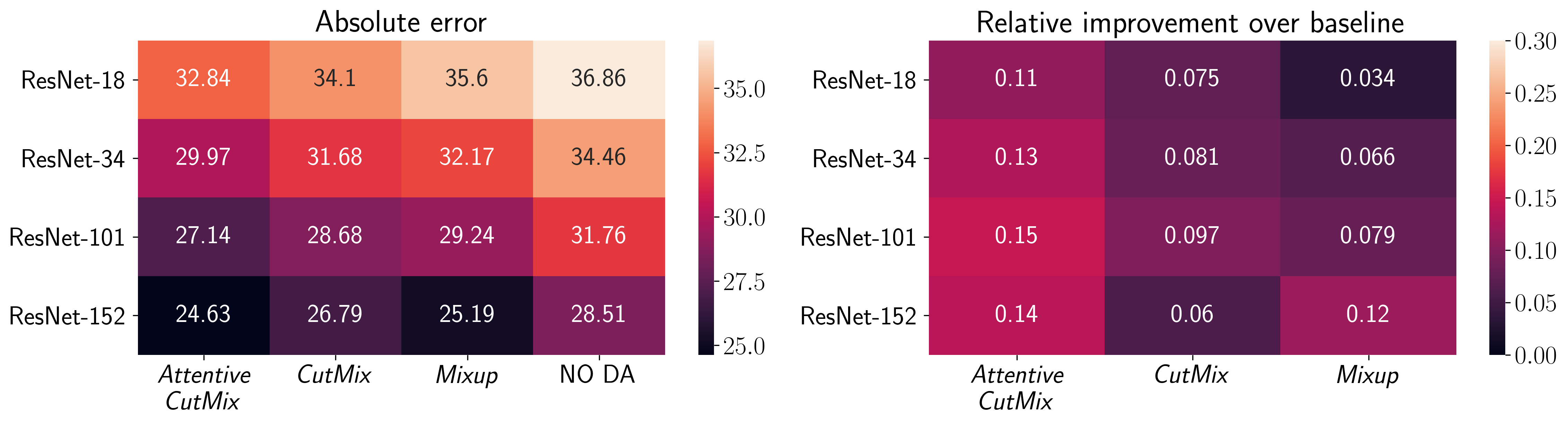}
\includegraphics[width=1\linewidth]{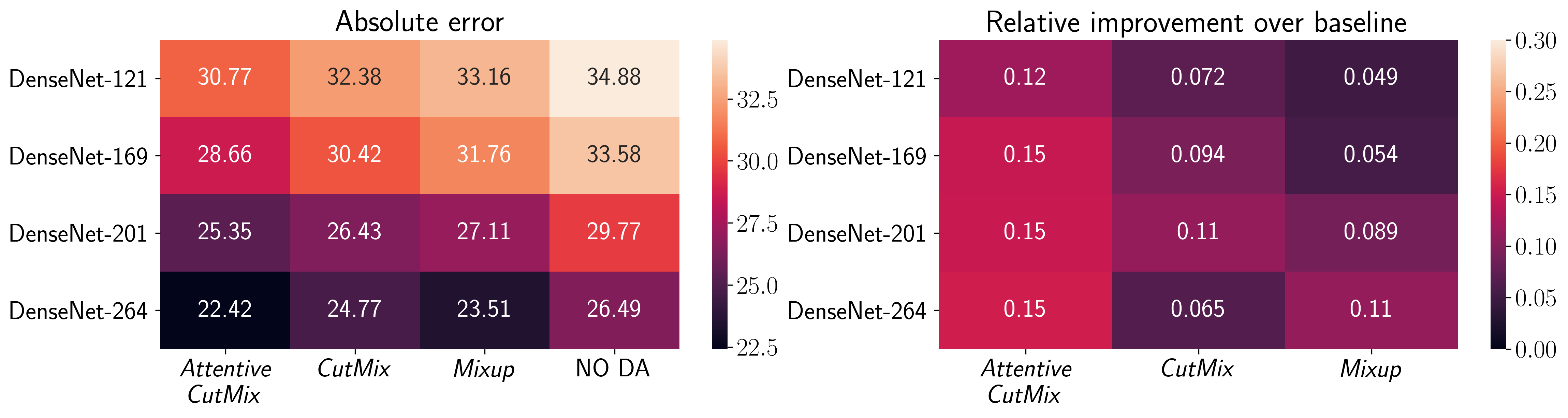}
\includegraphics[width=1\linewidth]{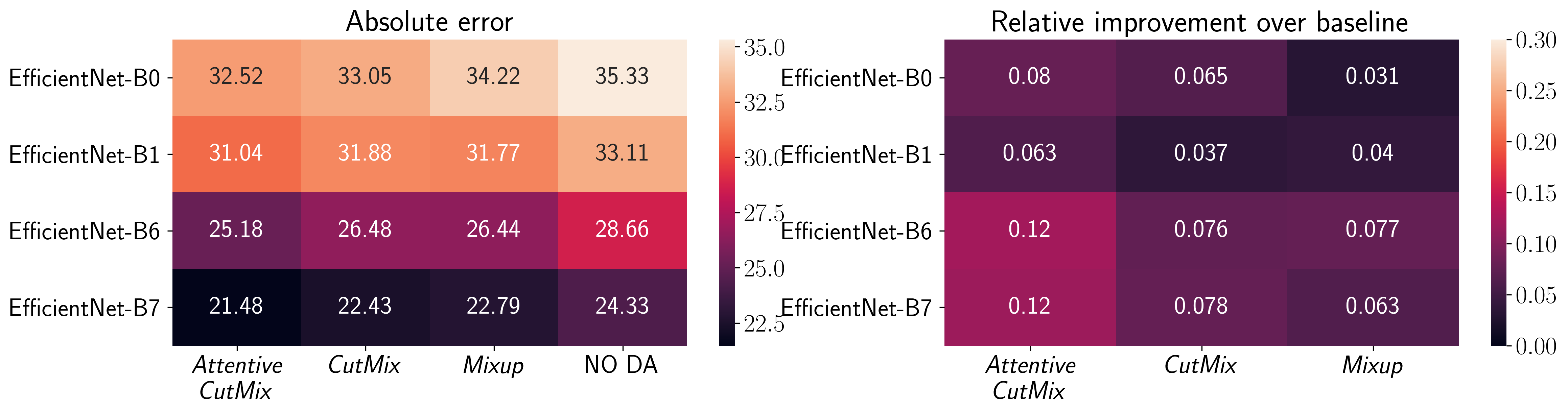}
\caption{
Accuracy results for \emph{Mixup}, \emph{CutMix} and \emph{AttentiveCutMix} on CIFAR-100 reported in~\cite{attentive_cutmix}. Each panel presents a particular type of architecture (ResNet, DenseNet, EfficientNet) with complexity of the network increasing from top to bottom.
Left panels present an absolute error value and the right ones a relative improvement over the baseline (``NO DA'' results).
}
\label{fig:cifar_100_groups}
\end{figure}

Figure~\ref{fig:cifar_100_groups} is a CIFAR-100 analogue of Figure~\ref{fig:cifar_10_groups} with the same range of tested architectures and DA methods. Its aim is to investigate how the network's complexity impacts the relative benefit of using particular DA techniques. The absolute errors (left panels) are much higher than in the case of CIFAR-10, because CIFAR-100 is a much more challenging data set (there are only 500 training samples per each of 100 fine-grained categories).

Interestingly, for CIFAR-100 a relative boost (right panels) is generally much smaller than in the case of its less complex counterpart (cf. Figure~\ref{fig:cifar_10_groups}).
This difference is most probably a direct consequence of smaller amount of training data available for each class compared to CIFAR-10.

Certain differences can also be observed in the average relative improvement across considered architecture types. While for CIFAR-10 the highest relative gain was observed for the smallest model (ResNet), in the case of CIFAR-100 the biggest relative improvement is noted for the DenseNet network. On average, the use of augmentation methods resulted in a relative gain of around $9.4\%$, $10.1\%$ and $7.1\%$, respectively for ResNet, DenseNet and EfficientNet. 
Similarly to CIFAR-10 architecture-based trends (within each panel) vary significantly.

\begin{figure}[!ht]
\includegraphics[width=1\linewidth]{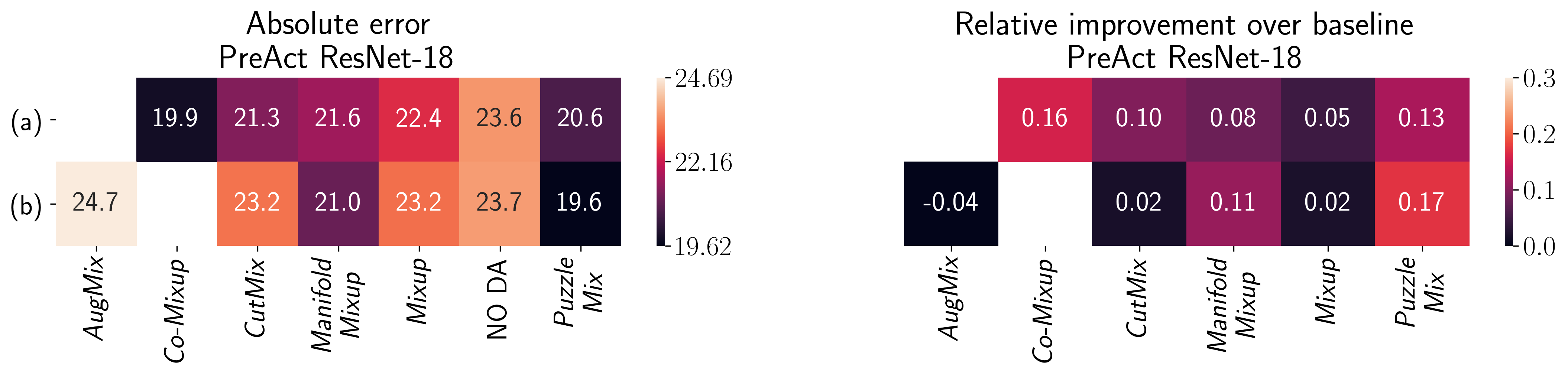}
\includegraphics[width=1\linewidth]{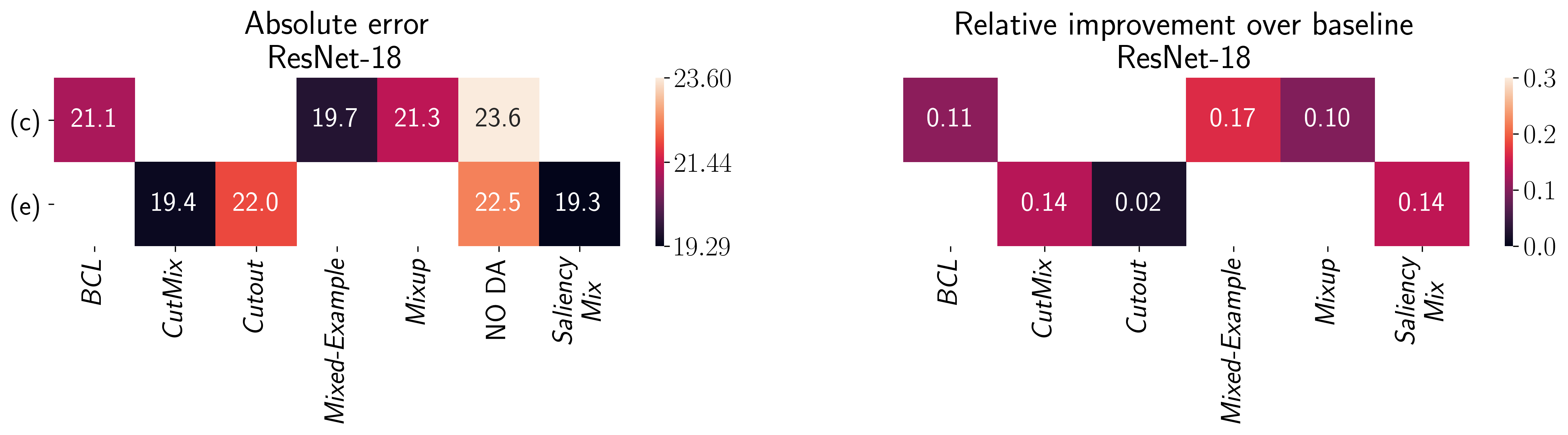}
\includegraphics[width=1\linewidth]{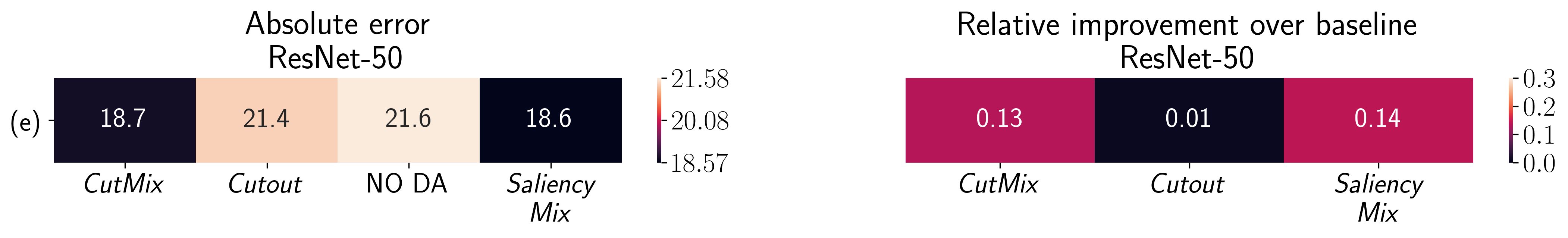}
\includegraphics[width=1\linewidth]{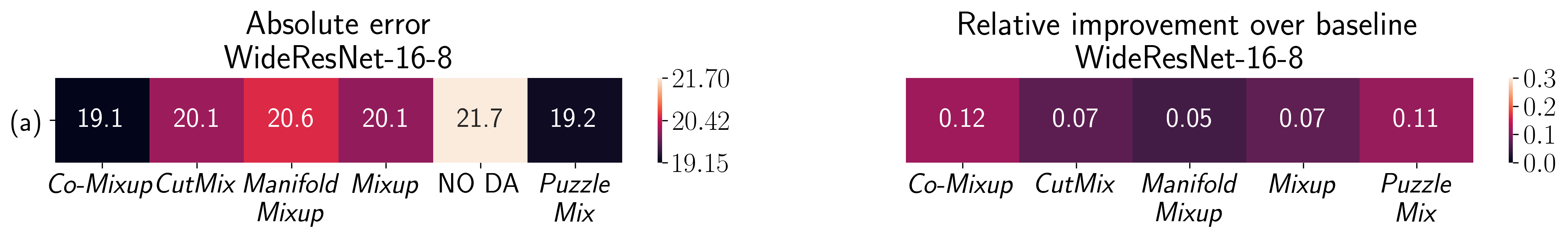}
\includegraphics[width=1\linewidth]{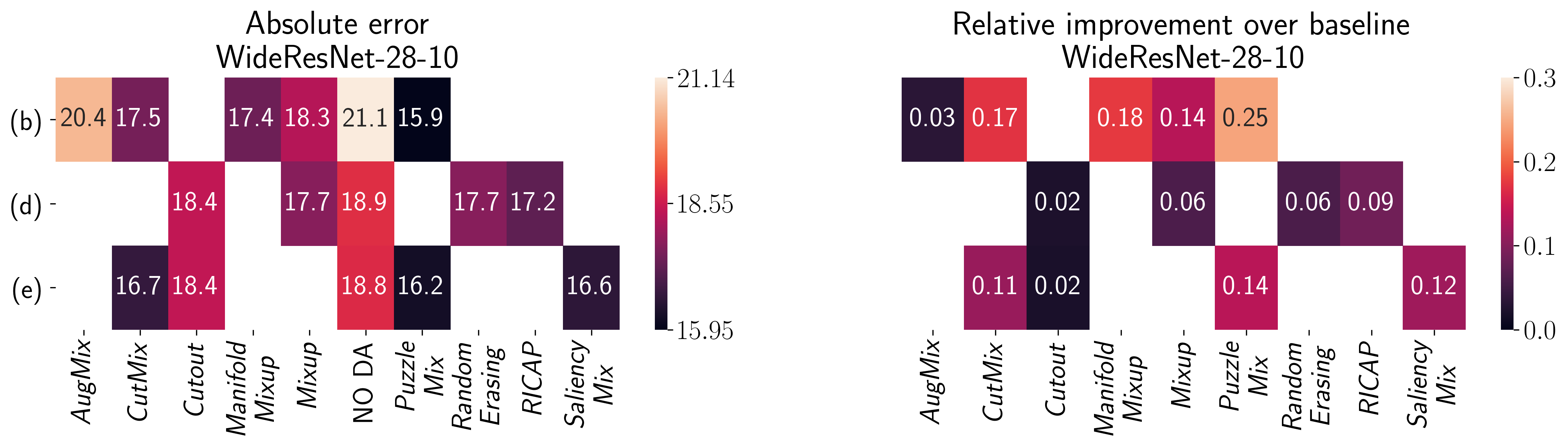}
\includegraphics[width=1\linewidth]{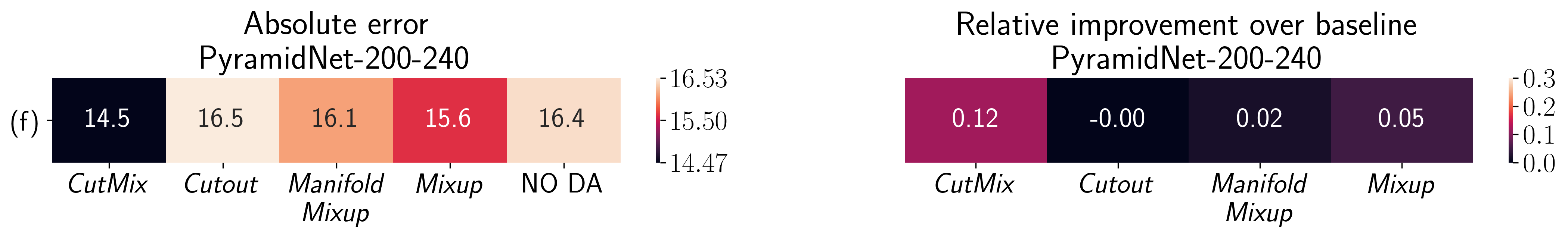}
\caption{
Accuracy results of various augmentation methods
on CIFAR-100 reported in:
(a) - \cite{co_mixup},
(b) - \cite{puzzle_mix},
(c) - \cite{mixed_example},
(d) - \cite{RICAP},
(e) - \cite{saliency_mix},
(f) - \cite{cutmix}
and grouped for particular ResNet architectures.
}
\label{fig:cifar_100_single}
\end{figure}

Figure~\ref{fig:cifar_100_single}
compares various augmentation methods across common ResNet baselines: PreAct ResNet~\cite{PreActResNet}, several ResNet
and WideResNet architectures~\cite{WideResNet} and Pyramidal ResNet~\cite{PyramidNet}.
This time models that relatively benefit most from DA application are not the simplest ones. Furthermore, it can be observed that application of certain augmentation techniques (\emph{AugMix} and \emph{Cutout}) actually deteriorates accuracy for some architectures.

\begin{table}[!ht]
\begin{tabular}{llrl}
\toprule
              Architecture &     Method &  Error &     Source \\
\midrule
        PyramidNet-200-240 &  \emph{SmoothMix} &  14.47 &  \cite{smoothmix} \\
        PyramidNet-200-240 &     \emph{CutMix} &  14.47 &     \cite{cutmix} \\
        PyramidNet-200-240 &      \emph{Mixup} &  15.63 &     \cite{cutmix} \\
          WideResNet-28-10 & \emph{Puzzle Mix} &  15.95 & \cite{puzzle_mix} \\
Shake-Shake Regularization &        \emph{BC+} &  16.00 &        \cite{between_class} \\
\bottomrule
\end{tabular}
\caption{Top-$5$ overall best outcomes on CIFAR-100.
}
\label{tab:top_10_CIFAR_100}
\end{table}

Table~\ref{tab:top_10_CIFAR_100} presents top-$5$ combinations of model architecture and augmentation method found in the literature. The best accuracy is achieved either by highly complex models (Pyramidal ResNet, Shake-Shake) joint with one of the founding methods (\emph{Mixup}, \emph{BCL} or \emph{CutMix}) or their extensions (\emph{SmoothMix}), or by applying one of the most recent data augmentation techniques, i.e. \emph{Puzzle Mix}, with less complex architecture.

\subsubsection{ImageNet}
\label{sec:ImageNet}

\begin{figure}[!ht]
\includegraphics[width=1\linewidth]{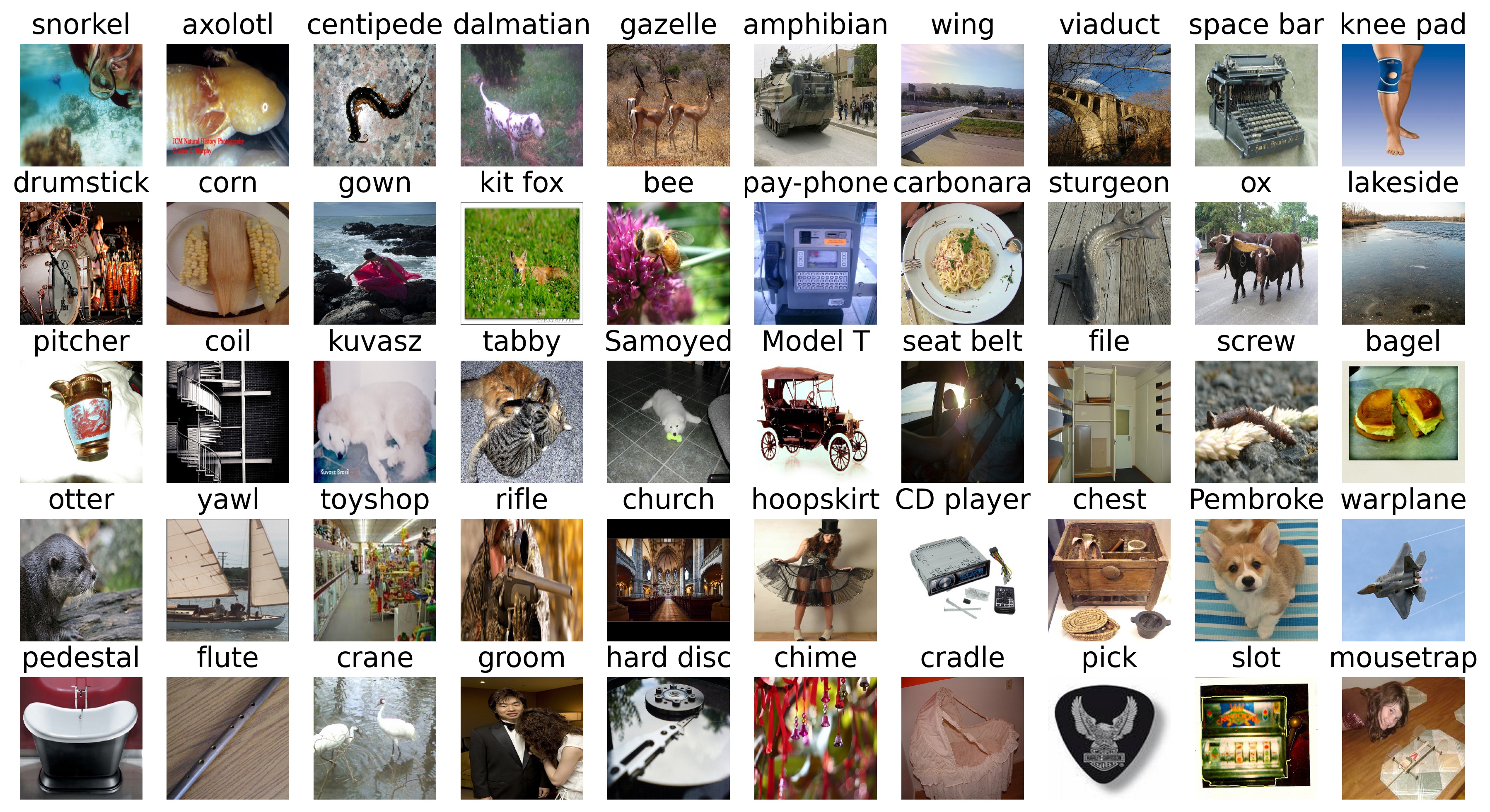}
\caption{
Examples of images from the ImageNet data set.
}
\label{fig:ImageNet}
\end{figure}

The most comprehensive benchmark that we consider is ImageNet~\cite{ImageNet}, which contains 1.2 million training samples and $50\,000$ validation ones, divided into $1\,000$ categories. The size of the images varies with the average of 469x387 pixels. In the experiments reported in the literature the images are usually cropped to 256x256 pixels or 224x224 pixels, depending on the architecture used. Example images are presented in Figure \ref{fig:ImageNet}.

\begin{figure}[!ht]
\includegraphics[width=1\linewidth]{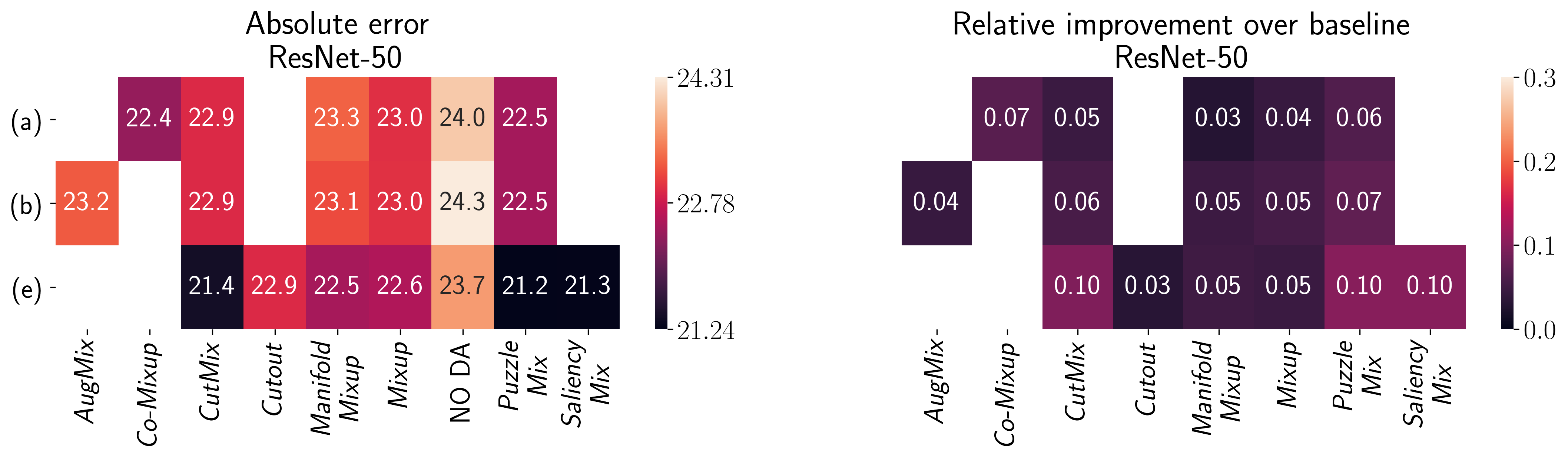}
\includegraphics[width=1\linewidth]{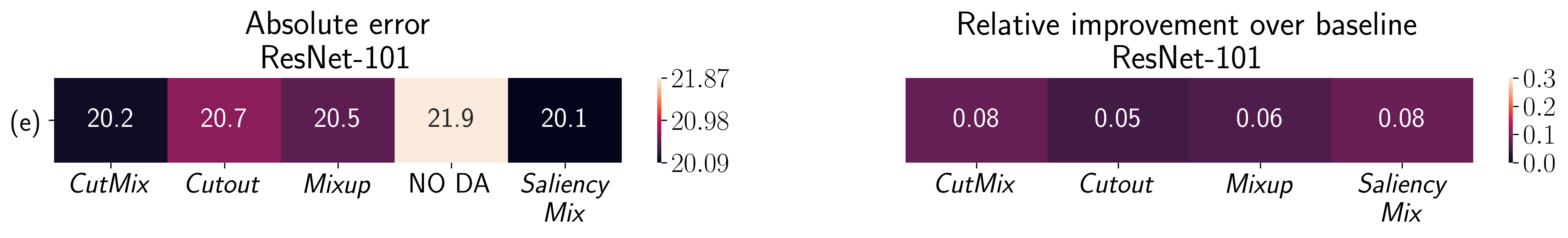}
\includegraphics[width=1\linewidth]{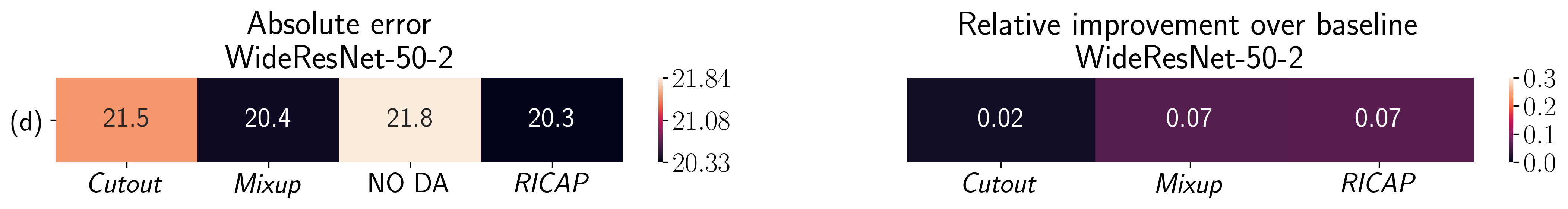}
\caption{
Accuracy results of various augmentation methods
on ImageNet reported in:
(a) - \cite{co_mixup},
(b) - \cite{puzzle_mix},
(d) - \cite{RICAP},
(e) - \cite{saliency_mix}
and grouped around particular ResNet architectures.
Left panels present absolute error values and the right ones the relative improvements over the baseline.
}
\label{fig:ImageNet_single}
\end{figure}

Figure~\ref{fig:ImageNet_single}
compares augmentation methods applied to various ResNet architectures~\cite{co_mixup,puzzle_mix,RICAP,saliency_mix}.
A predominant observation from the figure is that for ImageNet, a far more complex data set than the CIFAR sets, the relative advantage of applying data augmentation is much lower. The average relative increment is between $5-6\%$ and the maximum gain equals $10.5\%$ for \emph{Puzzle Mix} method and ResNet-50 model.

\begin{table}[!ht]
\begin{tabular}{llrl}
\toprule
     Architecture &       Method &  Error &       Source \\
\midrule
ResNeXt-101 64*4d &          \emph{BCL} &  19.43 &          \cite{between_class} \\
      ResNeXt-101 &       \emph{CutMix} &  19.47 &       \cite{cutmix} \\
ResNeXt-101 64*4d &        \emph{Mixup} &  19.80 &        \cite{mixup} \\
       ResNet-101 & \emph{Saliency Mix} &  20.09 & \cite{saliency_mix} \\
       ResNet-101 &       \emph{CutMix} &  20.17 &       \cite{cutmix} \\
\bottomrule
\end{tabular}
\caption{Top-$5$ overall best outcomes on ImageNet.}
\label{tab:top_10_ImageNet}
\end{table}

Table~\ref{tab:top_10_ImageNet} presents top-$5$ combinations of model architecture and augmentation method. Similarly to the CIFAR sets the leaders are the most advanced architectures (ResNeXt-101 64*4d and ResNeXt-101) combined with base mixing techniques, followed by certain less complex architectures paired with more recent mixing methods.

\subsection{Evaluation on \emph{corrupted images}}
\label{sec:results_corrupted_images}

\begin{figure}[!ht]
\includegraphics[width=1\linewidth]{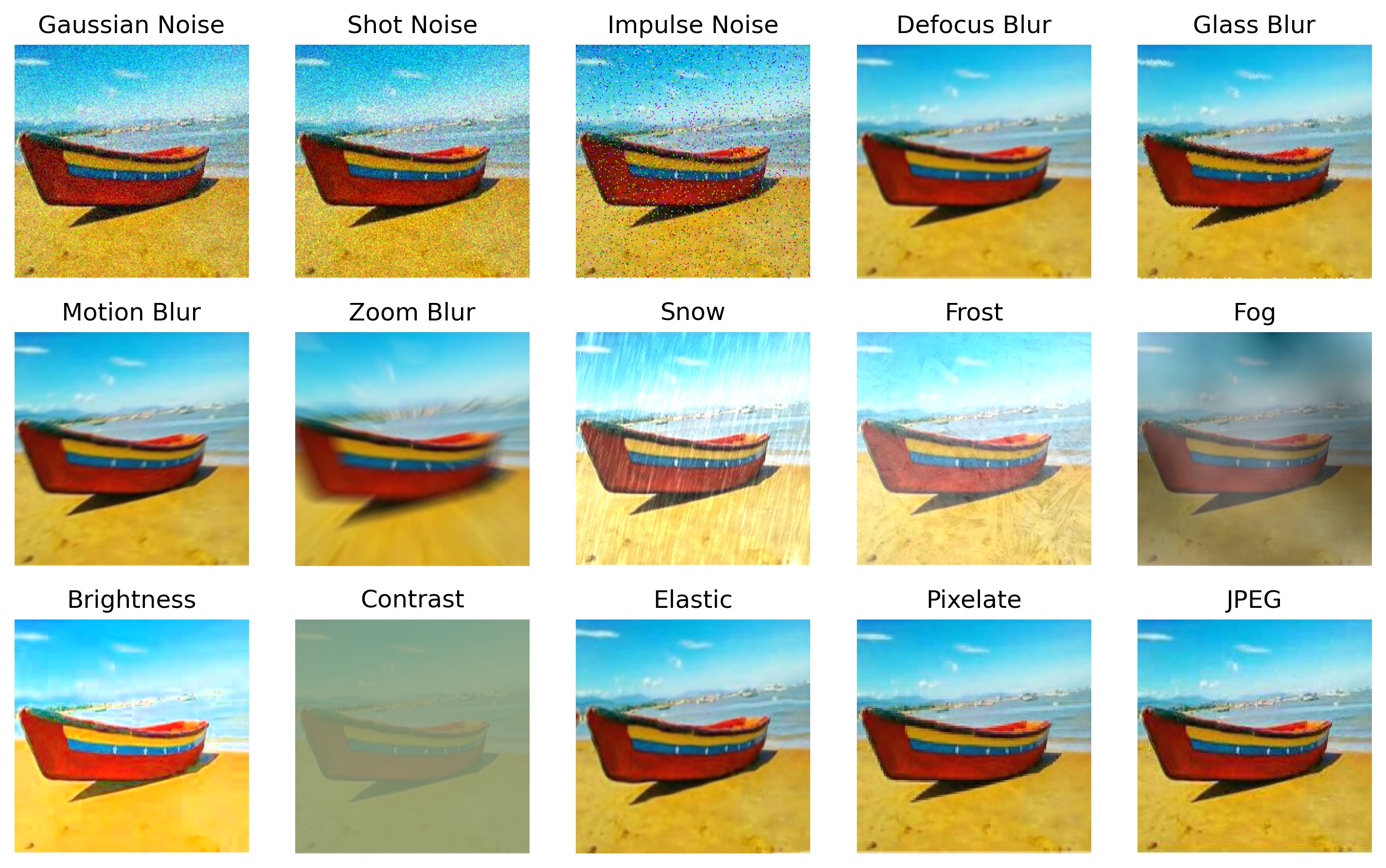}
\caption{
The effects of $15$ corruptions applied to an example image. Corruptions were applied with the highest severity. 
}
\label{fig:corruption_examples}
\end{figure}

In this section we verify the efficacy of models trained with DA techniques on corrupted test data.
Experimental evaluation is performed on CIFAR-100-C~\cite{cifar_c}, which is obtained from CIFAR-100 by applying $15$ different corruptions which are divided into four major categories: noise, blur, weather changes, and digital corruptions.
The effects of applying these corruptions are presented in Figure~\ref{fig:corruption_examples}. For each image from CIFAR-100 each corruption is applied at 5 severities indicating the corruption strength, which leads to $75$ transformed versions of this image.
Please note that in all experiments discussed in this section corrupted images were used only at test time and were not presented to the model at training time. The error measure, Mean Corruption Error (MCE), is calculated for a given image as the average across all its corrupted versions i.e. $75$ instances.

\begin{figure}[!ht]
\includegraphics[width=1\linewidth]{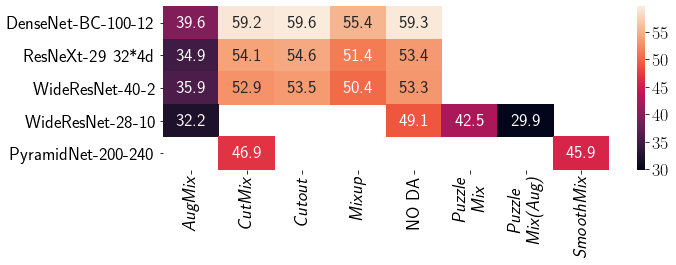}
\caption{
A summary of Mean Corruption Error results for various augmentation techniques
and network architectures. \emph{Puzzle Mix (Aug)} denotes a combined usage of \emph{Puzzle Mix} and \emph{AugMix} techniques. The first 3 rows come from~\cite{aug_mix} and the next ones from~\cite{puzzle_mix} and~\cite{smoothmix}, respectively.}
\label{fig:cifar_100_C_single}
\end{figure}

Figure~\ref{fig:cifar_100_C_single} aggregates the results for various DA techniques and architectures of varying complexity (DenseNet, ResNeXt, two Wide ResNet-s and Pyramidal ResNet).
It can be observed that on corrupted data, regardless of particular architecture, certain data augmentations (\emph{Cutout}, \emph{Mixup}, \emph{CutMix}, \emph{SmoothMix})
do not offer substantial improvement compared to the baseline results.

When leaving aside \emph{Puzzle Mix (Aug),} a method that consistently outperforms its competitors is \emph{AugMix} which was explicitly designed for dealing with corrupted data. A runner-up technique, which is also substantially better than the remaining methods, is \emph{Puzzle Mix}. Not surprisingly, the strongest overall approach is an application of \emph{Puzzle Mix} on top of \emph{AugMix} in which a pair of images, after being processed with \emph{AugMix}, is mixed using \emph{Puzzle Mix}. This combination, denoted as \emph{Puzzle Mix (Aug)} in the figure managed to further decrease the classification error down to the level of $29.9\%$, which is the lowest result on CIFAR-100-C reported in the literature.

\subsection{Evaluation on \emph{adversarial examples}}
\label{sec:results_adversarial_images}

\begin{figure}[!ht]
\includegraphics[width=1\linewidth]{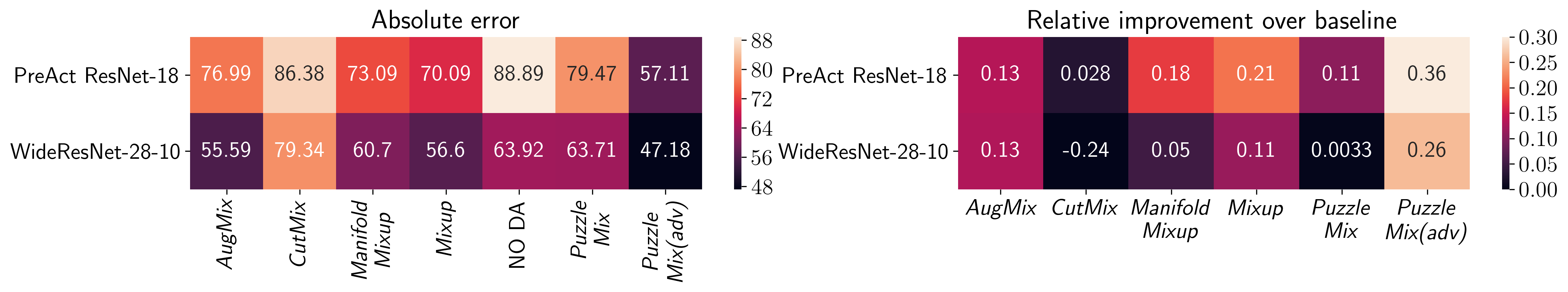}
\caption{
FGSM error rates on CIFAR-100 for two architectures (PreAct ResNet-18 and WideResNet-28-10) trained using various data augmentation techniques.
\emph{Puzzle Mix (adv)} is a combination of \emph{Puzzle Mix} and adversarial training~\cite{ResNet_cyclic_learning}. The left column presents absolute error values and the right one a relative improvement over the baseline. All results come from~\cite{puzzle_mix} which offers a broad comparison of data augmentation methods, in both clean data and adversarial attacks scenarios.
}
\label{fig:cifar_100_adversarial_1}
\end{figure}

\begin{figure}[!ht]
\includegraphics[width=1\linewidth]{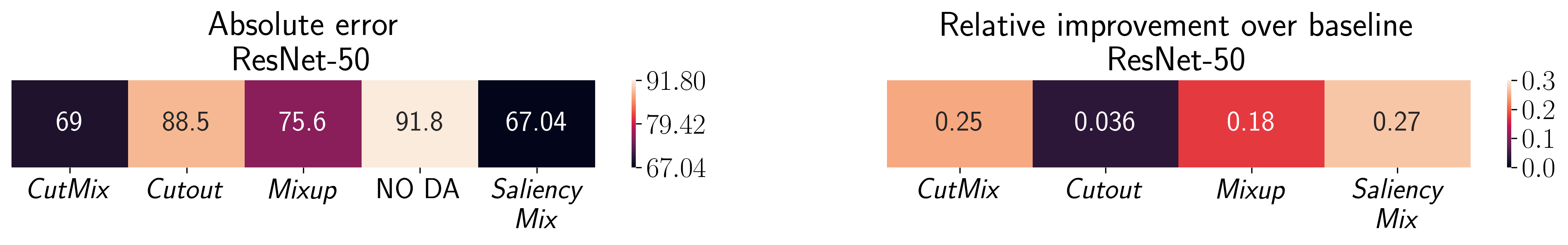}
\caption{
FGSM error rates on ImageNet for ResNet-50 architecture trained using three data augmentation techniques (\emph{Cutout}, \emph{Mixup} and \emph{CutMix}). The left column presents absolute error values and the right one a relative improvement over the baseline. All results come from~\cite{saliency_mix}.
}
\label{fig:cifar_100_adversarial_2}
\end{figure}

While this area is not the mainstream of image augmentation research, it is worth to mention that some augmentation methods were also evaluated against adversarial examples. An adversarial attack used most often in this context is Fast Gradient Sign Method (FGSM) employed as a white-box attack. In this case the gradient of the loss function is analyzed to find the most effective way of perturbing the image so as to mislead the classifier. Similarly to the case of image corruption adversarial examples are used exclusively at test time.

Figures~\ref{fig:cifar_100_adversarial_1} and~\ref{fig:cifar_100_adversarial_2} present the FGSM error rates from the experiments reported in~\cite{puzzle_mix} and~\cite{saliency_mix} regarding CIFAR-100 and ImageNet, respectively. Generally, the results are inconclusive as for CIFAR-100 pixel-wise mixing methods (\emph{Mixup}, \emph{Manifold Mixup}, \emph{AugMix}) are superior, whereas for ImageNet the best performing method is \emph{Saliency Mix}, a patch-wise mixing approach. It should be noted that in~\cite{manifold_mixup} visibly lower FGSM error rates for \emph{Mixup} ($59.3\%$) and \emph{Manifold Mixup} ($55.0\%$) were reported for CIFAR-100 and PreAct ResNet-18 network, however with no details regarding the experiment setup.

The method that clearly stands out in terms of accomplished results is \emph{Puzzle Mix (adv)}~\cite{puzzle_mix}, which is a variant of \emph{Puzzle Mix} employing adversarial training. At the same time, this is the only method that requires substantially more computational power compared to the ``NO DA'' reference case.

\subsection{Evaluation on \emph{weakly supervised object localization and partial occlusion tasks}}
\label{sec:results_WSOL_occlusion}

In this section we discuss the efficacy of DA techniques in the problems of weakly supervised object localization (WSOL) and partial occlusion of the target object, which are frequently encountered in the image classification or object detection settings. In both contexts a desired property of the model is to make prediction based on a wide array of class-relevant visual features and not just the most distinctive features. In WSOL this property helps in finding a fragment of an image that entirely covers a given object, and in the context of partial occlusion supports the network in inferring the correct class based on the remaining, less distinctive features.

Utilization of augmented data in WSOL is discussed in~\cite{cutmix,RICAP,saliency_mix,snap_mix,co_mixup} in reference to \emph{CutMix}, \emph{RICAP}, \emph{Saliency Mix}, \emph{SnapMix} and \emph{Co-Mixup}, respectively. In all cases the results are assessed using Class Activation Mapping~\cite{CAM} already mentioned in section~\ref{sec:Mixup-puzzle}.
CAM creates a heat map of the same size as input images, indicating regions in the original image that contributed most towards a certain class prediction. CAM requires a CNN model to possess a global average pooling layer to obtain the spatial average of the feature maps prior to the output layer, so as to visualize the most class-contributive pixels.

The results indicate that augmented data indeed broadens the regions used by the model to infer the correct class~\cite{cutmix,RICAP,co_mixup} with \emph{Saliency Mix} being more discriminative against the image background. The only exception is \emph{SnapMix}, which was designed for fine-grained classification and limits the regions activated by CAM to parts of the image that are relevant for a particular fine-grained class (e.g wings instead of a whole bird).

\emph{RICAP} was tested in~\cite{RICAP} also in object detection task. The experiment was conducted on MS COCO (Microsoft Common Objects in Context) set~\cite{COCO} using YOLOv3 architecture~\cite{YOLOv3}. MS COCO is a large set of images presenting common objects with associated bounding boxes and segmentation maps, accompanied by the respective captions. In the reported experiments only information about the bounding boxes was used.

\emph{RICAP} was adjusted so as not to interfere with the area in which an object was located. The target bounding boxes where restricted to the cropped regions. This way the model was no longer able to benefit from partial labels (each object had one class assigned to it) but could learn partial features (only part of an object might fall into the cropped region, so the network had to learn to make predictions based on the remaining part). This adjusted version of \emph{RICAP} slightly improved the mean Average Precision (mAP) over the baseline (YOLOv3 without augmentation) from $51.3\%$ to $52.7\%$~\cite{RICAP}.

{\color{black}
\subsection{Application to image-related tasks other than classification}
\label{sec:different_tasks}
}
{\color{black}
    The canonical tasks in Computer Vision are \textit{image categorization}, \textit{object localization}, \textit{object detection} and \textit{semantic segmentation}. These tasks can be further divided into specific cases: 
    \begin{itemize}
        \item \textit{Fine grained categorization} -- in which differences between classes refer to small visual features.
        \item \textit{Categorization under domain shift} -- in which a model is trained on data from a distribution other than the one applied at inference stage.
        \item \textit{Categorization with federated learning} -- in which a model is trained on data that is scattered among many devices as opposed to being collected in one place.
        \item \textit{Categorization of corrupted images} -- which refers to images distorted by certain noise, blur, weather conditions, digital processing, etc.
        \item \textit{Categorization of adversarial images} -- which refers to images that were purposefully modified to confuse the model.
        \item \textit{Weakly supervised object localization} -- in which the goal is to localize an object based solely on the image label.
        \item \textit{Object detection in the presence of partial occlusion} -- where object are often partly occluded by other objects. 
    \end{itemize}
    
    When it comes to application of certain data augmentation techniques to various tasks, there are two key properties: (1) whether the mixing is performed pixel-wise or patch-wise and (2) how many images are mixed. Consequently the methods could be divided into 3 following groups:
    \begin{itemize}
        \item Group A - Pixel-wise or mixed Pixel-wise and Patch-wise augmentations that work on 2 or more images: \emph{Smart Augmentation}~\cite{smart_augmentation}, \emph{Feature Space}~\cite{feature_space}, \emph{Mixup}~\cite{mixup},  \emph{SamplePairing}~\cite{sample_pairing}, \emph{Between-Class learning}~\cite{between_class},
        \emph{AdaMixup}~\cite{adamixup},
        \emph{Manifold Mixup}~\cite{manifold_mixup},  \emph{SmoothMix}~\cite{smoothmix}, \emph{Puzzle Mix}~\cite{puzzle_mix} and \emph{Co-mixup}~\cite{co_mixup}.
        \item Group B - Patch-wise methods: \emph{RICAP}~\cite{RICAP}, patch-wise versions of \emph{Mixed-Example}~\cite{mixed_example}, \emph{CutMix}~\cite{cutmix}, \emph{Attentive CutMix}~\cite{attentive_cutmix}, \emph{Saliency Mix}~\cite{saliency_mix} and \emph{SnapMix}~\cite{snap_mix}.
        \item Group C - methods that work on just 1 image: \emph{Style Augmentation}~\cite{style_augmentation}, \emph{AugMix}~\cite{aug_mix}, \emph{MixStyle}~\cite{mix_style}, \emph{Cutout}~\cite{cutout}, \emph{Random Erasing}~\cite{random_erasing} and \emph{Patch Gaussian}~\cite{patch_gaussian}.
    \end{itemize}
    Methods from group A are limited to categorization task due to their underlying property of mixing images-pixel wise. This leads to certain regions of the image representing more than one class and renders application of these methods to other tasks difficult (e.g. what should be done with a bounding box for a part of the image that is mixed?). 
    
    Augmentations in group B directly address categorization and localization tasks 
    and can be further adjusted to object detection and segmentation by proper handling of an additional information associated to the task (e.g. \emph{RICAP}~\cite{RICAP} method limits the bounding box to the area corresponding to the selected patch).
    
    Augmentation approaches from group C, which work on a single image, can in principle be applied to all tasks.
    
    On a more detailed level, there are certain methods that were either developed with a particular problem in mind or were studied in the context of a specific problem and perform well on it. 
    \begin{itemize}
        \item Fine grained categorization -- \emph{SnapMix}~\cite{snap_mix}.
        \item Categorization under domain shift -- \emph{MixStyle}~\cite{mix_style}.
        \item Categorization with federated learning -- \emph{Mixup} applied to federated learning~\cite{fedmix}.
        \item Categorization of corrupted images -- \emph{Manifold Mixup}~\cite{manifold_mixup}, \emph{SmoothMix}~\cite{smoothmix}, \emph{Puzzle Mix}~\cite{puzzle_mix}, \emph{AugMix}~\cite{aug_mix}.
        \item Categorization of adversarial images -- \emph{Mixup}~\cite{mixup}, \emph{Manifold Mixup}~\cite{manifold_mixup}, \emph{CutMix}~\cite{cutmix}, \emph{Puzzle Mix}~\cite{puzzle_mix}.
        \item Weakly supervised object localization -- \emph{CutMix}~\cite{cutmix}, \emph{RICAP}~\cite{RICAP}.
        \item Object detection in the presence of partial occlusion -- \emph{CutMix}~\cite{cutmix}. 
    \end{itemize}
    }

{\color{black}
\subsection{Application to modalities other than images}
\label{sec:different_modalities}
}
{\color{black}
    The main focus of this survey is data augmentation for images, however there are also other modalities that could be considered for application of the analyzed augmentation methods: text and audio. In this section we point the methods that could potentially be extended to other modalities, and present examples of such applications already reported in the literature.
    
    When it comes to applying image augmentation methods to other modalities 
    the following
    3 groups could be distinguished:
    
    \begin{itemize}
        \item Group A - \emph{Mixup}-like methods (Pixel-wise mixing) that work on 2 or more images and do not utilize any complex mixing mechanism, i.e. \emph{Mixup}~\cite{mixup}, \emph{Feature Space}~\cite{feature_space},   \emph{SamplePairing}~\cite{sample_pairing}, \emph{Between-Class learning}~\cite{between_class} and \emph{Manifold Mixup}~\cite{manifold_mixup}.
        
        \item Group B - Patch-wise methods or mixed Pixel-wise and Patch-wise, i.e. \emph{RICAP}~\cite{RICAP}, patch-wise versions of \emph{Mixed-Example}~\cite{mixed_example}, \emph{CutMix}~\cite{cutmix}, \emph{SmoothMix}~\cite{smoothmix}, \emph{Cutout}~\cite{cutout} and \emph{Random Erasing}~\cite{random_erasing}.
        
        \item Group C - methods that cannot be directly applied to other modalities due to their inherent connection to image-specific data transformations or architectures, i.e.  \emph{Style Augmentation}~\cite{style_augmentation}, \emph{AugMix}~\cite{aug_mix}, \emph{MixStyle}~\cite{mix_style}, \emph{Smart Augmentation}~\cite{smart_augmentation}, 
        \emph{AdaMixup}~\cite{adamixup}, \emph{Puzzle Mix}~\cite{puzzle_mix}, \emph{Attentive CutMix}~\cite{attentive_cutmix}, \emph{Saliency Mix}~\cite{saliency_mix},
        \emph{SnapMix}~\cite{snap_mix}, \emph{Co-mixup}~\cite{co_mixup} and \emph{Patch Gaussian}~\cite{patch_gaussian}.
    \end{itemize}
    
    Methods from group A can be applied to other modalities without any adaptation as long as 
    the same size of the input objects is ensured. In the context of audio it means having the same length and the same spectrum of frequencies, and 
    for the text data, the same size of vector embeddings.
    Examples of successful applications of the methods from group A to other modalities are presented below in this section.
    
    For methods from group B their application to modalities other than images is technically possible, however, not yet empirically tested. Such a mixing would potentially signify specific modality-depending aspects, e.g. spatial mixing of embeddings of different sentences or pasting a part of a voice spectrogram into another one.

    A potential application of the methods from group C 
    to other modalities 
    would require introducing major changes to their design and operation, as they are inherently related to image data.
    Some methods from this group utilize image saliency information
    \cite{co_mixup,snap_mix,saliency_mix,puzzle_mix}, other use 
    image specific data transformations, like style transfer or rotation \cite{mix_style,aug_mix,style_augmentation}. Yet another ones, 
    utilize architectures dedicated to processing the image data
    \cite{smart_augmentation,adamixup,attentive_cutmix}. 
    
    So far several examples of applying image augmentation methods to other modalities have been proposed in the recent literature. Most notably the 
    \emph{Mixup} method was applied to speaker classification task 
    \cite{VOICE_MIXUP_1} and to sentence classification \cite{mixup_for_text_2}. Also \emph{Mixup} like method were applied to rare event detection based on sound data \cite{VOICE_MIXUP_2}
    and to text classification \cite{mixup_for_text_1}.
    
Even though application of augmentation methods to non-image data seems to be scarce, we are convinced that this area offers many flourishing research directions related to both adaptation of the methods from groups B and C, as well as invention of the entirely new augmentation approaches devoted to particular data modalities and addressing their specificity.
    }

{\color{black}
\subsection{Robustness to parameter selection}
\label{sec:robustness}
}

 The majority of the methods presented in the survey rely on no more than 3 hyper-parameters 
 and are relatively robust to their selection or change, as well as the change of the data set.
 
 The method's robustness to the problem change (same task different data set) is mostly correlated with the 
 way the augmentation method is applied (cf. Section 4.2). For the rule based methods, the change of a problem does not require any hyper-parameter changes. For other methods, however, one can face some difficulties with their application to other data sets. \emph{Attentive CutMix}~\cite{attentive_cutmix} utilizes an additional feature extractor that might not be available for all problems.
 \emph{Puzzle Mix}~\cite{puzzle_mix} and \emph{Co-mixup}~\cite{co_mixup} employ an optimization procedure that requires some assumptions on how to simplify the optimization problem which might differ between problems (e.g. wideness of the image may impact how the grid of location for the optimization process is created). The next method, \emph{Style Augmentation}~\cite{style_augmentation}, was shown to be sensitive to hyper-parameter change in the ablation study. In case of this method both strength of regularization via style transfer as well as ratio of augmented samples to non-augmented ones greatly impact the method's accuracy. Yet another method that we believe is hard to apply directly to other data sets is \emph{Smart Augmentation}~\cite{smart_augmentation} -- one of the older approaches that employs a dedicated problem-specific network responsible for mixing images. Training of this network, in addition to computational overhead, is generally difficult and 
 may potentially lead to poor mixed samples that would impact the accuracy of the target network.

\section{Selecting the best data augmentation strategy}
\label{sec:augmentation_strategy}

Up to now we have discussed newest mixing augmentation methods whose utilization improved the state-of-the-art results in many visual tasks,
for instance image classification (either clean~\cite{between_class,cutmix} or corrupted~\cite{aug_mix,puzzle_mix}), object detection~\cite{RICAP} or WSOL~\cite{RICAP,cutmix}. While application of gradually more advanced and more complicated methods proven successful, an alternative approach is to consider an ensemble of traditional augmentation methods and pick-up those of them that are particularly well suited to a given task and/or experiment setup.

Traditional data augmentation techniques, e.g. image scaling, translation or rotation are generally effective in improving accuracy of DL image classifiers, however an impact of a particular augmentation depends on characteristics of the data set and the task at hand, which poses certain limitations to their effective application. Moreover, the number and diversity of traditional augmentation methods prohibit using them all at once, as such a situation would heavily slow down the training process and actually deteriorate the accuracy.
Hence, the ability to limit the number of considered augmentation options for a given problem (task, data set, imposed constraints, etc.) is an important issue. In this section several automated approaches to effective selection of the optimal subset of traditional augmentation techniques (for a given task and dataset) are summarized and evaluated. These methods will be referred to as data augmentation policy selection (DAPS) approaches.

\subsection{Black Box methods}
\label{sec:block_box}

A seminal DAPS method is \emph{AutoAugment}~\cite{AutoAugment} which is composed of two major components:
the search algorithm and the search space. The search engine (an RNN controller) samples from the search space the DA policy $S$ defined as the following triple: (image augmentation operation, probability of its application, magnitude of the operation).
The controller is trained with Reinforcement Learning Proximal Policy Optimization algorithm~\cite{PPO} based on the reward signal from the auxiliary model (e.g. a CNN model in the case of image classification) which measures the efficacy of the selected policy in improving model generalization. After completion of the policy exploration phase the auxiliary models are discarded and the best policies found are used for target model training. The choice of RL as the training algorithm is arbitrary and is inspired by automated architecture search techniques~\cite{neural_architecture_search}. Actually, any other suitable technique, e.g. augmented random search or evolutionary strategy could be used in its place.
In order to cast the problem of selecting $S$ into a discrete search space, the augmentation methods' probabilities and magnitudes are discretized to uniformly spaced values.

In the search for the optimal DA policy only a fraction of observations from the original data set are used, for instance $4\,000$ and $1\,000$ in the case of CIFAR-10 and SVHN~\cite{svhn}, respectively. The underlying principle of \emph{AutoAugment} is that an optimal DA policy for a given data set would not change if the entire data set was considered.

The best policies found for the two above-mentioned data sets are different and clearly linked to the image content.
For CIFAR-10 the vast majority of best augmentation policies are focused on color-based transformations since within the respective classes (e.g. airplane or truck) the diversity introduced by color-related augmentations coincide with real-life cases where one object can be of various colors. In SVHN the best augmentations are geometric transformations since for street numbers categorization they seem to be more important than color, which is actually irrelevant if only the number is readable.

An interesting question is whether augmentation policies are transferable between data sets. To this end the best \emph{AutoAugment} policies found for SVHN were subsequently applied in the training process of the networks solving classification problems on $5$ other data sets (Oxford 102 Flowers~\cite{flowers}, Caltech-101~\cite{caltech101}, Oxford-IIIT Pets~\cite{pets}, FGVC Aircraft~\cite{aircrafts} and Stanford Cars~\cite{cars}). In all cases using policies learned on SVHN resulted in performance increase over the baseline~\cite{AutoAugment}.

In the ablation experiments application of an optimal policy found by \emph{AutoAugment} was compared with the use of the same subset of augmentation techniques, but with random probabilities and magnitudes, as well as with random policy sampling. In both cases \emph{AutoAugment} proven superior.

The main limitation of the method is its extensive computational cost. In order to generate sufficient number of training signals for the RNN controller, thousands of auxiliary models need to be trained with sampled augmentation policies before the final out-of-sample accuracy is achieved.
All the remaining methods presented in this section efficiently address this problem while keeping the results at comparable level.

The first method that bases upon the idea of \emph{AutoAugment} and matches its accuracy is \emph{Population Based Augmentation} (\emph{PBA})~\cite{PBA}. \emph{PBA} produces dynamic augmentation policy schedule and combines random search with evolutionary strategy to significantly decrease computation time.

The difference between the policy (constructed in \emph{AutoAugment}) and the policy schedule (in \emph{PBA}) lies in the time range for which the optimal augmentation policy is selected. In \emph{AutoAugment} the optimal policy is selected once for the entire training time. In \emph{PBA} it is chosen for each epoch, leading to a sequence of optimal policies, i.e. a policy schedule.
The underpinning idea of \emph{PBA} is that diversification of policies applied at different training stages is advantageous for the final accuracy.

The main motivation of~\cite{PBA} is to demonstrate that state-of-the-art results (comparable to those of the \emph{AutoAugment} application) can be achieved with significantly lower computational cost. As a backbone \emph{PBA} uses Population Based Training algorithm (PBT)~\cite{PBT_algorithm} that selects a subset of augmentation techniques independently for each epoch.
%
%
The schedule learned so far is used as a starting point. In other words, at any given epoch all candidate policies share the common pool of policies selected historically.

Initially PBT trains in parallel an ensemble of randomly initialized models with the same architecture. At certain intervals the performance of these models is evaluated on a validation set. At this point two mechanisms are applied: \emph{exploitation} and \emph{exploration}. Exploitation consists in replacing the weights of $25\%$ of the worst-performing models with the weights from the top-$25\%$ models. Exploration relies on randomly perturbing the hyperparameters of the models with replaced weights, in order to extend the hyperparameter space search. \emph{PBA} uses exactly the same hyperparameters as \emph{AutoAugment}, i.e. a list of triples: (transformation type, probability of its application, its magnitude).

Similarly to \emph{AutoAugment} the quest for an optimal DA policy is performed on a reduced data set.
In the experiments presented in~\cite{PBA} the policy schedule is learned only once using WideResNet architecture, and is subsequently applied to training other models (based on other architectures). Since DA method applied in a given epoch is not linked to any particular architecture, the only relevant difference from the perspective of the DA process is the number of required training epochs. The training schedule is adjusted proportionally by stretching it out (shrinking) in case a particular architecture requires longer (shorter) training than WideResNet. For instance, an optimal schedule for $2$ epochs consisting of a set of augmentations A and B, could be stretch out to $4$ epochs by repeating each of these sets, leading to a schedule A, A, B, B.


\emph{Fast AutoAugment} (\emph{FAA})~\cite{FastAutoAugment} is another approach indicating \emph{AutoAugment} as its inspiration. \emph{FAA} addresses the problem of \emph{AutoAugment} computational complexity by using a more effective search strategy based on density matching. Instead of repeatedly training auxiliary models with different augmentations in order to select the best-performing one, \emph{FAA} searches for augmentation policies that minimize categorical cross-entropy loss on validation set. The process relays on treating the augmented data as missing data points and selecting the best-performing augmentations using the Tree-structured Parzen Estimator (TPE) algorithm \cite{TPE}.

\emph{FAA} search space is similar to that of \emph{AutoAugment} and \emph{PBA} as the method uses a list of transformation operations, each of them described by its probability and magnitude. The goal is to find a set of optimal policies that can be used for the final model training. The key difference is that TPE enables searching over continuous space so neither probabilities nor magnitudes need to be discretized.

The search process consists of the following steps. First, the data set is split into $K$ pairs (${D_{\mathcal{M}}^{(k)}}, {D_{\mathcal{A}}^{(k)}}$) using $K$-fold stratified shuffling where ${D_{\mathcal{M}}^{(k)}}$  is the non-augmented data, ${D_{\mathcal{A}}^{(k)}}$  is the data that will be used to evaluate different augmentations, for $k = {1,\ldots,K}$. Next, the model is trained from scratch using ${D_{\mathcal{M}}^{(k)}}$ only. After training a set of augmentation policies is selected for evaluation. The policies parameters are optimized by minimizing the categorical cross entropy on the ${D_{\mathcal{A}}^{(k)}}$ set. Since in order to evaluate a new policy only augmentation and prediction is required, there is no need to train the model again from scratch.

Once the above exploration phase is completed, top-N policies are selected as the optimal augmentation policies for a given data set. In the final step, the model is trained using the entire training data set and the optimal policies selected across $K$ folds.

All three above-described methods (\emph{AutoAugment}, \emph{PBA}, \emph{FAA}) are based on the assumption that a set of augmentation policies that will result in high model accuracy can be found based on a smaller proxy task. Therefore, all of them limit the size of the training set used to search for the optimal augmentation policies.

\subsection{Reducing the search space}
\label{sec:search_space_reduction}

The sole method in this area is \emph{RandAugment}~\cite{Randaugment},
which questions an assumption that a search for optimal policies over the entire solution space can be effectively performed with a smaller proxy task. This disbelief is based on the experiments showing that the optimal augmentation magnitude
depends on the size of a training set and the size of a model. Consequently, \emph{RandAugment} postulates significant reduction of the search space so as to make the selection of an optimal augmentation policy a feasible task.

To this end it is proposed in~\cite{Randaugment} to jointly optimize magnitudes of all augmentation operations by means of a single parameter called distortion magnitude. Furthermore, it is proposed to uniformly set the probability of all augmentation operations, thus reducing the search space even further.
The resulting algorithm depends on just two parameters: the number of transformations applied to an image and the magnitude distortion parameter indicating regularization strength.

In the experiments
performed on CIFAR-10 a family of WideResNet architectures with various values of the widening parameter, responsible for the number of convolutional filters, hence complexity of the network, were trained on data sets of various sizes. The first one tested $7$ WideResNet models with various widening factors,
and the other one considered $8$ different sizes of the training set, between $1\,000$ and $45\,000$ images sampled randomly from CIFAR-10 with fixed WideResNet architecture with widening factor $= 10$. Both evaluations showed that more complex networks and/or bigger data sets require stronger regularization to achieve full classification potential. This finding is somewhat counterintuitive in reference to the training set size where a common belief is that smaller training sets require more regularization. A hypothetical explanation presented in~\cite{Randaugment} is that strong augmentation on small data set might result in high noise to signal ratio.

\subsection{Making the augmentation pipeline differentiable}
\label{sec:differentiable_augmentation_pipeline}

Methods in this section transform augmentation operations, which up to now were intrinsically non-differentiable, to differentiable ones, thus allowing joint optimization of the data augmentation pipeline and classification network.

Chronologically, the first paper in this area is \emph{Faster AutoAugment}~\cite{FasterAutoAugment} that builds on and enhances \emph{Fast AutoAugment} approach~\cite{FastAutoAugment} described in section~\ref{sec:block_box}. The method adjusts the policy search pipeline so as to make it fully differentiable and consequently enable gradient based optimization. This is achieved by implementing the following three concepts: (1) approximate gradient calculation of discrete image operation~\cite{StraighThroughEstimator}, (2) making the operation selection process differentiable thanks to the adaptation of neural architecture search methods~\cite{DARTS}, and (3) expanding the idea presented in~\cite{FastAutoAugment}, where the validation loss on the augmented sample was minimized, to GAN approach with augmentation policy treated as a generator and a critic network~\cite{WGAN}.

\emph{Faster AutoAugment} also adapts the foundations of DAPS, i.e. organization of the augmentations into policies and sub-policies, as well as the way the experiment is structured (policy search on smaller sample, model training on full data). Since the method transforms the search for optimal parameters into an optimization task solving which is more time-effective, it can be applied to larger search spaces. It is concluded in~\cite{FasterAutoAugment} that increasing the number of sub-policies or the number of operations in each sub-policy is increasing the performance.

Differentiable Automatic Data Augmentation (\emph{DADA})~\cite{DADA}
relaxes the DA policy selection problem to differentiable one using (continuous) Gumbel-Softmax distribution that approximates samples from the categorical distribution.

Technically, \emph{DADA} approaches the tasks of sub-policy selection and augmentation parameter selection by sampling from a categorical distribution and Bernoulli distribution, respectively
and then relaxes this optimization problem to differentiable one using Gumbel-Softmax. Eventually, joint optimization of augmentation parameters and classification network weights is carried out using the RELAX estimator~\cite{RELAX}.

Finally, the bi-level optimization of DA parameters and classification network weights is carried out according to the following equations:

\begin{equation}
\label{DADA:augmentation_network}
\ \min \mathcal{L}_{val}( \omega^{*}(d))
\end{equation}

\begin{equation}
\label{DADA:classification_network}
\ \text{subject to}
\ \omega^{*}(d) = \underset{\omega}{\arg\min} \mathbf{E}[\mathcal{L}_{train}(\omega, d)]
\end{equation}

where, $d = \{\alpha, \beta, m, \phi\}$ represent probability of selecting a sub-policy ($\alpha$), probability of applying a transformation ($\beta$), magnitude of the transformation ($m$), and RELAX network parameters ($\phi$), respectively. $\omega$ represents parameters of the classification network. In order to solve the bi-level optimization problem (\ref{DADA:augmentation_network})-(\ref{DADA:classification_network}) $\omega$ and $d$ are optimized alternately through gradient descent. As in \emph{AutoAugment} the policies are searched for on a reduced data set, although using only one network training limited to 20 epochs.

The next method, \emph{Adversarial AutoAugment}~\cite{AdversarialAutoAugment} generates adversarial images as augmented samples.
The method performs training on full data set, getting rid of the smaller proxy task, and returns a dynamic policy schedule which is updated during training.

Technically, \emph{Adversarial AutoAugment} uses two neural networks: Task Network (TN) and Policy Search Network (PSN). PSN is responsible for sampling policies that should be applied to training data. It is implemented as a one-layer LSTM \cite{LSTM} with 100 neurons in hidden layer and an embedding size of 32. TN is trained on data augmented with the sampled policies. Each image from the training batch is processed multiple times using different policies and all these transformed versions are added to the training batch.
Based on the parts of the TN loss function associated with each sampled policy better policies are generated. PNS parameters are updated using the REINFORCE algorithm~\cite{reinforce}.
\emph{Adversarial AutoAugment} modifies a default policy search space by skipping policy probability and considering continuous policy magnitudes:

\begin{equation}
\ w^{*} = \underset{w}{\arg\min} \underset{x\sim\Omega}{\mathbf{E}} \underset{\tau\sim\mathcal{A}(.,\theta)}{\mathbf{E}} \mathcal{L}[\mathcal{F}(\tau(x), w), y]
\end{equation}

\begin{equation}
\ \theta^{*} = \underset{\theta}{\arg\max} \underset{x\sim\Omega}{\mathbf{E}} \underset{\tau\sim\mathcal{A}(.,\theta)}{\mathbf{E}} \mathcal{L}[\mathcal{F}(\tau(x), w), y]
\end{equation}

where TN and PSN are indicated with $\mathcal{F}$ and $\mathcal{A}$, respectively and are parameterized by $w$ and $\theta$. The optimization problem is formulated as a min-max game: TN tries to train a network and PSN attempts to increase the loss by providing harder augmentations, also making TN more robust. Such a setup as it conforms to the current state of the training process and provides augmentation with lower severity at the initial training phase (to allow for quick learning of the main patterns) and increases the severity in subsequent training once the network learned all simple patterns.

The high level setup of \emph{Adversarial AutoAugment} is similar to that of GANs in that one network tries to outplay the other one. Yet the method is inherently different as it works on existing images by applying a transformation from a pre-defined list and does not synthesize new images from noise, as GANs typically do.

Policies learned using \emph{Adversarial AutoAugment} transferred to other problem give comparable or better results than \emph{AutoAugment}, however the best results are achieved when the policies are searched for independently, focusing on a given data set~\cite{AdversarialAutoAugment}.

Another method in this group, \emph{MetaAugment}~\cite{MetaAugment}, challenges one of the well established DA assumptions that an augmentation transformation is selected for the entire data set. \emph{MetaAugment} evaluates a fit of the selected augmentation operation to the image and assigns a weight based on the significance of this fit. The weight is calculated by Augmentation Policy Network (APN) and passed on to Task Network (TN) to calculate the weighted loss of the augmented training image.

APN takes as its input an embedding of a transformation function together with deep features extracted from the image with TN and outputs a weight to adjust the augmented image loss computed by TN. The goal of APN is to improve performance of TN on a validation set via adjusting the weights of the losses.
APN is implemented as a 1-hidden layer MLP that takes the aforementioned embedding and deep features as its input, which is followed by a fully-connected layer of size 100 with RELU activations and an output layer with one sigmoid neuron.
%
%
TN can be implemented as any of the standard CNN architectures that minimizes the weighted training loss using the weights provided by APN: 

\begin{equation}
\label{MA:augmentation_network}
\ \theta^{*} =  \underset{\theta}{\arg\min}\ \mathcal{L}(\mathcal{X}_{val}, \omega^{*}(\theta))
\end{equation}

\begin{equation}
\label{MA:classification_network}
\ \text{subject to}
\ \omega^{*}(\theta) = \underset{\omega}{\arg\min}\ \mathcal{L}(\mathcal{X}_{tr}, \omega, \theta)
\end{equation}

where $\theta$ and $\omega$ are parameters of APN and TN and $\mathcal{X}_{tr}$ and $\mathcal{X}_{val}$ denote training and validation sets, respectively. \emph{MetaAugment} proposes an additional mechanism, the transformation sampler (TS), working in tandem with APN and TN. TS samples transformations according to a probability distribution estimated in the outputs of APN that reflects the overall effectiveness of the transformation for the entire dataset. As the distribution is estimated based on APN, it evolves over the training process and provides gradually better recommendations.

One of the \emph{MetaAugment} disadvantages is computational overhead. The method requires 3 forward and backward passes in TN which takes 3 times longer than in a typical training scenario. Its clear advantage, on the other hand, is the ability to fit the transformation to a given sample.
While this fitting is not performed explicitly because the image and transformation are sampled independently, it is forced by assigning the weight to the image, which denotes the impact of this image on the loss function that is used to update TN weights.

The most recent method in this area~\cite{ABO} tries to jointly learn the optimal data augmentation parameters while training the end task model. The method will be henceforth referred to as \emph{ABO} (Data Augmentation with Online Bi-level Optimization). \emph{ABO} employs two neural networks, a CNN Classification Network (CN) and an MLP Augmentation Network (AN). The training process is divided into two streams. In the inner loop AN predicts augmentation parameters, based on which the image is augmented, and fed afterwards into CN to calculate the loss. In the outer loop a feedback signal for AN is created:

\begin{equation}
\label{ABO:augmentation_network}
\ \theta^{*} =  \underset{\theta}{\arg\min} \mathcal{L}(\mathcal{X}_{val}, \omega^{*})
\end{equation}

\begin{equation}
\label{ABO:classification_network}
\ \text{subject to}
\ \omega^{*} = \underset{\omega}{\arg\min} \mathcal{L}(\mathcal{A}_{\theta}(\mathcal{X}_{tr}), \omega)
\end{equation}

where $\mathcal{A}_{\theta}$ is AN parameterized by $\theta$, and CN is parameterized by $\omega$.

Joint training of AN using back propagation is possible thanks to the following features. Firstly, AN is trained on the validation set (cf. eq.~(\ref{ABO:augmentation_network})) to improve generalization properties of CN. Secondly, in order to enable gradient calculation the AN error on the validation set is calculated using CN (eq.~(\ref{ABO:augmentation_network})) and subsequently used to update parameters of AN. Thirdly, an online approximation of the bi-level optimization is proposed to enable updating of AN parameters at each training step. Fourthly, differentiable augmentation operations from Kornia library~\cite{kornia} are used.

The online approximation of the bi-level optimization is used to overcome the problematic setup used in many previous methods~\cite{AutoAugment,PBA,Randaugment} where optimization process treats the two objectives separately as black-box problems. In ABO the weights of AN are updated after each step in reference to CN outputs:

\begin{equation}
\label{ABO:exact}
\ \nabla_{\theta} \mathcal{L}(\mathcal{X}_{val}, \omega^{*}) =
\ \frac{\partial \mathcal{L}(\mathcal{X}_{val}, \omega^{*})}{\partial \theta} =
\ \frac{\partial \mathcal{L}(\mathcal{X}_{val}, \omega^{*})}{\partial \omega^{*}} \frac{\partial \omega^{*}}{\partial \theta}
\end{equation}

\begin{equation}
\label{ABO:online}
\ \frac{\partial \omega^{*}}{\partial \theta} \approx \frac{\partial \omega^{(t)}}{\partial \theta^{(t)}} = \sum_{i=1}^{t} \frac{\partial \omega^{(t)}}{\partial \omega^{(i)}} \frac{\partial \omega^{(i)}}{\partial \mathcal{G}^{(i-1)}} \frac{\partial \mathcal{G}^{(i-1)}}{\partial \theta^{(i)}}
\end{equation}

where $\mathcal{G}^{(t)}$ is the gradient of the training loss at iteration $t$.

AN is implemented as an MLP with input and output of size $n$, being the number of hyperparameters to optimize, and two hidden layers with $n$ and $10n$ neurons, respectively. Bigger architectures were also tested but it was found out empirically that the size of AN does not have significant impact on the accuracy of the classifier~\cite{ABO}.

Contrary to \emph{Adversarial AutoAugment}~\cite{AdversarialAutoAugment} and \emph{MetaAugment}~\cite{MetaAugment} where subsequent training resulted in more adequate change of the method's parameters, in the case of \emph{ABO} the transformations applied at the beginning of the training are stronger and approach identity towards the end of the training.

\subsection{Quantitative evaluation}
\label{sec:aug_strat_qualitative_evaluation}

We quantitatively compare the results of application of nine DAPS methods presented in sections~\ref{sec:block_box}-\ref{sec:differentiable_augmentation_pipeline} on the same three benchmark sets 
that were used in evaluation of previously introduced DA methods. Methods are compared between each other and against selected erasing and mixing approaches described in sections~\ref{sec:erasing_image} and~\ref{sec:mixing_images}, respectively.

\begin{figure}[!ht]
\includegraphics[width=1\linewidth]{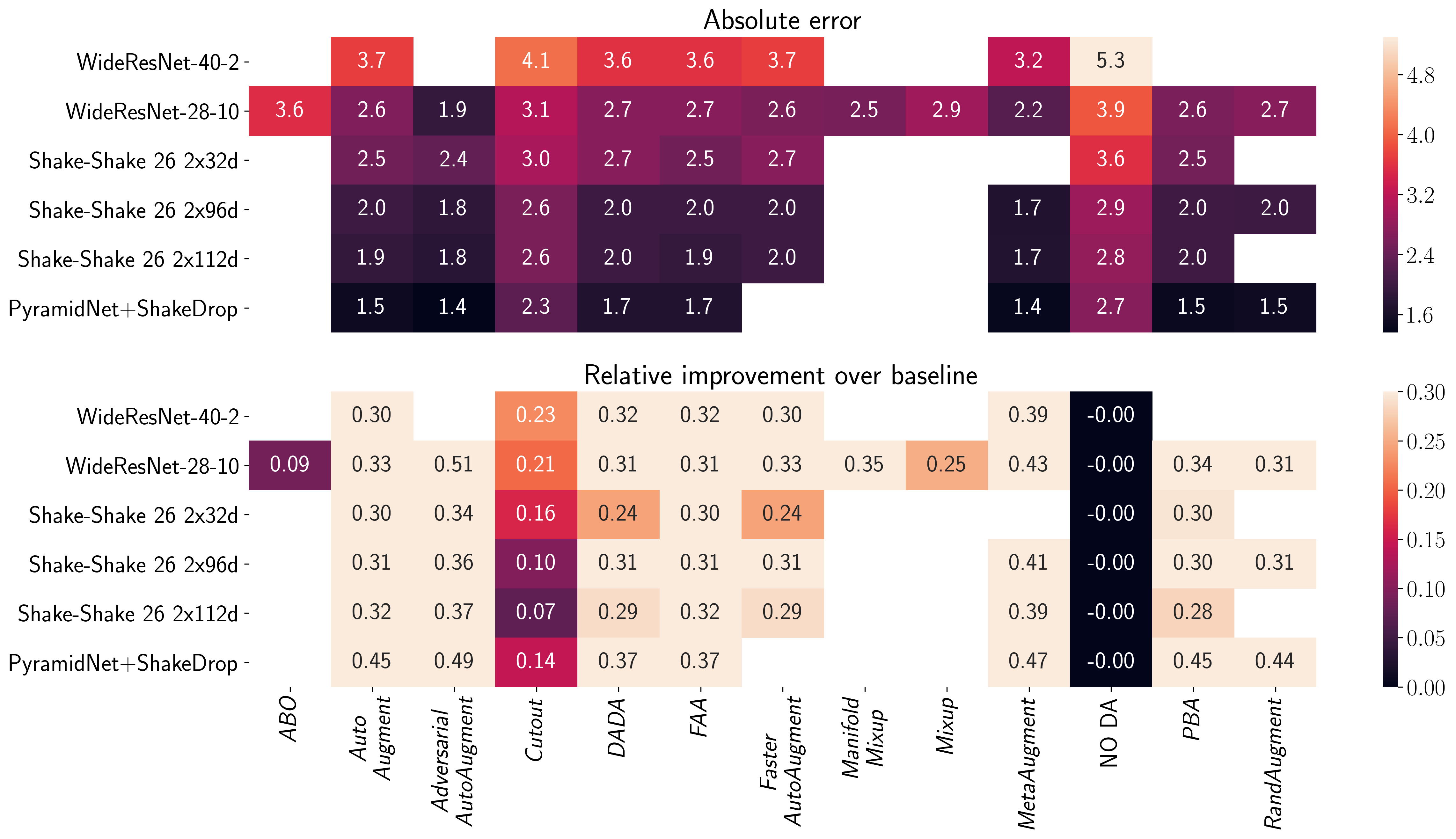}
\caption{
Accuracy results on CIFAR-10 of nine DAPS methods (\emph{AutoAugment}, \emph{FFA}, \emph{PBA}, \emph{Rand Augment}, \emph{Faster AutoAugment}, \emph{DADA}, \emph{Adversarial AutoAugment}, \emph{MetaAugment}, \emph{ABO}) and selected mixing and erasing DA approaches. Top row presents absolute error values and the bottom one relative improvements over the baseline. All papers introducing DAPS methods~\cite{AutoAugment,FastAutoAugment,PBA,Randaugment,FasterAutoAugment,DADA,AdversarialAutoAugment,MetaAugment,ABO} adopted the same experiment design.
}
\label{fig:cifar_10_augmentation_strategies}
\end{figure}

A summary of results on CIFAR-10 presented in Figure~\ref{fig:cifar_10_augmentation_strategies} shows that application of DAPS methods yields on average
better results than utilization of advanced pixel-wise or patch-wise DA approaches.
The only exception is \emph{Manifold Mixup}, which slightly outperforms the majority of DAPS solutions, albeit the supporting evidence comes from one experiment only. \emph{Adversarial AutoAugment} and \emph{MetaAugment}, two methods that introduce changes not aimed solely at reducing \emph{AutoAugment} computation time, reign among DAPS methods.

\begin{figure}[!ht]
\includegraphics[width=1\linewidth]{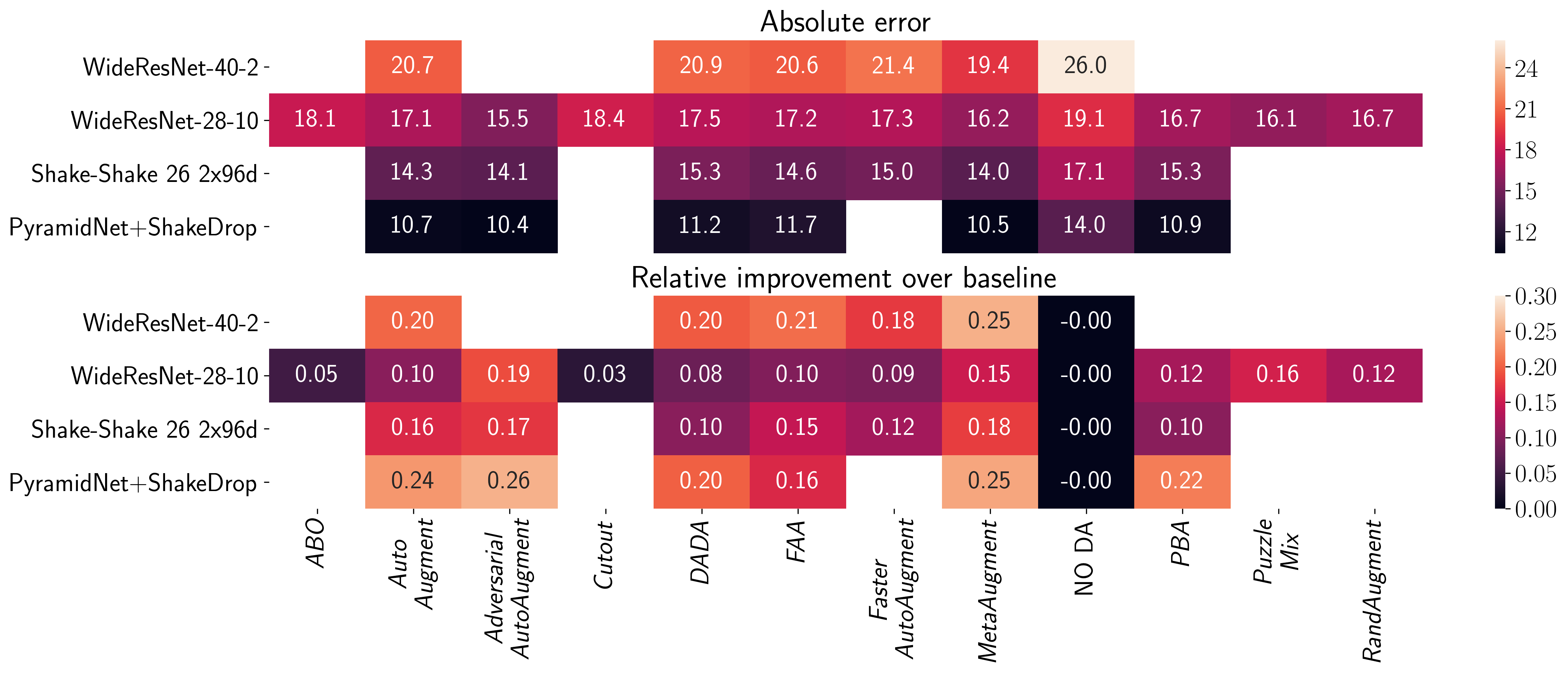}
\caption{
Accuracy on CIFAR-100 for nine DAPS methods (\emph{AutoAugment}, \emph{FFA}, \emph{PBA}, \emph{Rand Augment}, \emph{Faster AutoAugment}, \emph{DADA}, \emph{Adversarial AutoAugment}, \emph{MetaAugment}, \emph{ABO}) and selected mixing and erasing DA approaches. Top row presents absolute error values and the bottom one relative improvements over the baseline. All the papers describing DAPS methods~\cite{AutoAugment,FastAutoAugment,PBA,Randaugment,FasterAutoAugment,DADA,AdversarialAutoAugment,MetaAugment,ABO} adopted the same experiment design.
}
\label{fig:cifar_100_augmentation_strategies}
\end{figure}

\begin{figure}[!ht]
\includegraphics[width=1\linewidth]{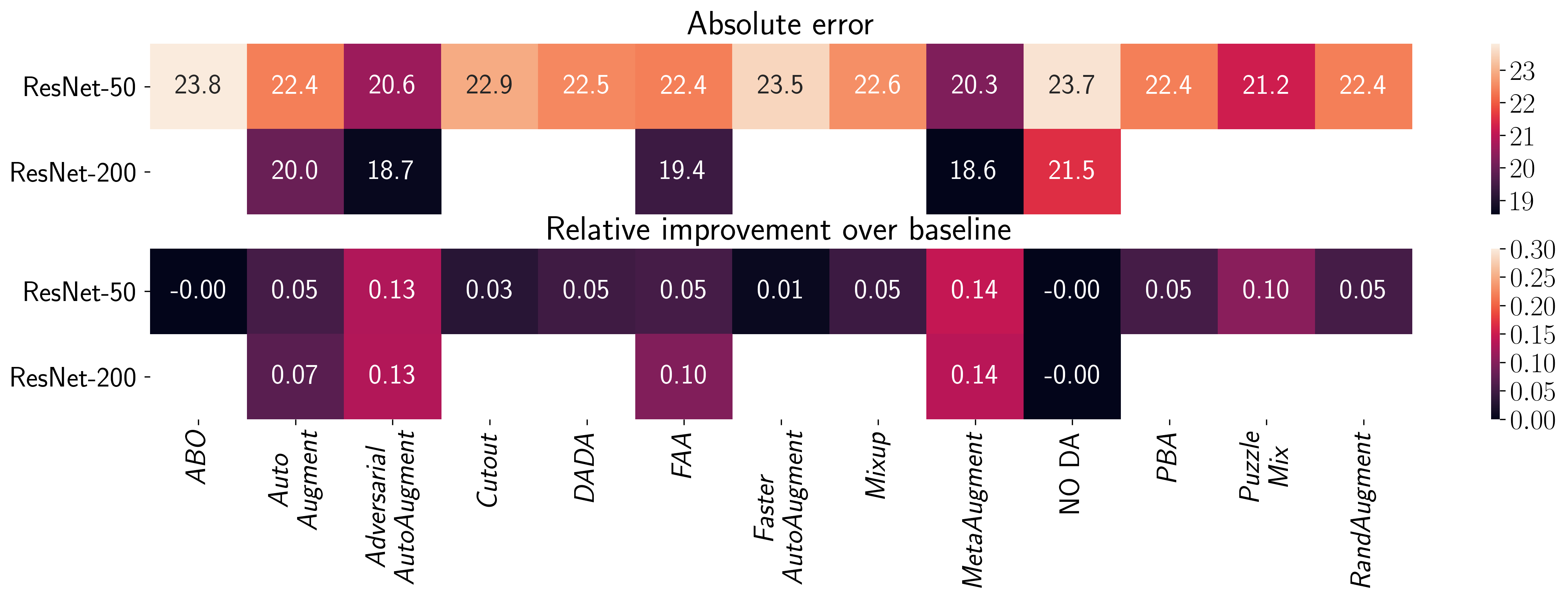}
\caption{
Accuracy results on ImageNet for nine DAPS methods (\emph{AutoAugment}, \emph{FFA}, \emph{PBA}, \emph{Rand Augment}, \emph{Faster AutoAugment}, \emph{DADA}, \emph{Adversarial AutoAugment}, \emph{MetaAugment}, \emph{ABO}) and selected mixing and erasing augmentation approaches. Top row presents absolute error values and the bottom one relative improvements over the baseline. All the papers describing DAPS methods~\cite{AutoAugment,FastAutoAugment,PBA,Randaugment,FasterAutoAugment,DADA,AdversarialAutoAugment,MetaAugment,ABO} adopted the same experiment design.
}
\label{fig:ImageNet_augmentation_strategies}
\end{figure}

The above observations are also confirmed in experiments with more complex data sets. In the case of CIFAR-100 (see Figure~\ref{fig:cifar_100_augmentation_strategies}), again all DAPS methods fare very well - with the average error on WideResNet-28-10 architecture slightly below $17$. At the same time \emph{Puzzle Mix} (a DA approach) achieves the best overall score. Likewise, on ImageNet (Figure~\ref{fig:ImageNet_augmentation_strategies}) the results of majority of DAPS methods are close to each other except the two leading approaches (\emph{Adversarial AutoAugment} and \emph{MetaAugment}) which outperform the rest. The results are generally better than those of DA methods, except \emph{Puzzle Mix} which demonstrates a slight upper-hand over all but the two above-mentioned methods.
%

%
A general observation is that the more complex and bigger the data set is the smaller relative improvement over the baseline, stemming from DAPS usage, can be expected. Depending on the architecture the relative improvement varies between $24\%$ and $51\%$ for CIFAR-10, between $5\%$ and $26\%$ for CIFAR-100, and between $1\%$ and $14\%$ for ImageNet. Based on the results we conclude that the two top methods are \emph{MetaAugment} and \emph{Adversarial AutoAugment} - which both are recently developed DAPS approaches.

An open question, which we believe is worth investigation is application of DAPS methods to more complex DA techniques, going beyond traditional augmentations. The results presented in Figures~\ref{fig:cifar_10_augmentation_strategies}-\ref{fig:ImageNet_augmentation_strategies} suggest that inclusion of more complex erasing or mixing methods into DAPS pool of augmentation techniques might lead to their efficacy improvement beyond the state-of-the-art \emph{MetaAugment} and \emph{Adversarial AutoAugment} solutions.

\subsection{Comparison of DAPS methods based on particular properties}
\label{sec:DAPS-aspects}

This section collates all DAPS approaches introduced above in reference to their key aspects summarized in Table~\ref{tab:strategies_all}.

\begin{sidewaystable}
\begin{tabular}{llllllllll}
\toprule
Method        & \emph{Auto}- &\emph{PBA} & \emph{FAA} &  \emph{Rand}- 	& 		  \emph{Faster} &		\emph{DADA} & 	  \emph{Adversarial} & \emph{Meta}- & \emph{ABO} \\
        	  	& \emph{Augment}&                   &                & \emph{Augment} & \emph{AutoAugment} &                                 & \emph{AutoAugment} & \emph{Augment} &                          \\
\midrule
How           &  RL &  ES &  DM &  GS &  DM+diff &  BO+MC &  BO+AT &  BO+SR &  Online BO \\
Dynamic       &                      No &                  Yes &                 No &           No &                                               No &                                 No &                                    Yes &                      Yes &                          Yes \\
Cost [ht]      &                  5000.0 &                  5.0 &                3.5 &          0.0 &                                             0.23 &                                0.1 & no info & no info & no info \\
Epochs        &                 1800000 &                 3200 &               1200 &         1000 &                                               20 &                                 20 &                                    200 &                      600 &                      no info \\
Epochs norm   &                  240200 &                  456 &                400 &         1000 &                                              216 &                                216 &                                   1600 &                      200 &                      no info \\
Search Space  &                   10\textasciicircum 32 &                10\textasciicircum 61 &              10\textasciicircum 32 &         10\textasciicircum 2 &                                            10\textasciicircum 32 &                              10\textasciicircum 32 &                                  10\textasciicircum 22 &                    10\textasciicircum 32 &                      no info \\
Search space  &                Discrete &             Discrete &         Continuous &     Discrete &                                         Discrete &                           Discrete &                               Discrete &               Continuous &                      no info \\
Proxy task    &                    4000 &                 4000 &              10000 &            0 &                                             4000 &                               4000 &                                      0 &                        0 &                            0 \\
Final task    &                   50000 &                50000 &              50000 &        50000 &                                            50000 &                              50000 &                                 400000 &                    50000 &                        50000 \\
Auxiliary models  &                   15000 &                   16 &                  5 &            0 &                                                1 &                                  1 &                                      0 &                        0 &                            0 \\
Target models &                       1 &                    1 &                  1 &            5 &                                                1 &                                  1 &                                      1 &                        1 &                            1 \\
\bottomrule
\end{tabular}
\caption{Comparison of DAPS methods along key differentiating axes. The following abbreviations are used: RL - Reinforcement Learning, ES - Evolutionary Strategy, DM - Density Matching, GS - Grid Search, BO - Bi-level Optimization , MC - Monte Carlo gradient estimate, AT - Adversarial Training , diff - augmentations made differentiable and SR - Sample Re-weighting. Values in rows ``Proxy task'' and ``Final task'', as well as the value in row ``Target models'' for \emph{RandAugment} are specific for CIFAR-10 data set. }
\label{tab:strategies_all}
\end{sidewaystable}

\subsubsection{The DAPS baseline algorithm}
\label{sec:algorithm}

One of the key factors differentiating DAPS methods is an algorithm used to address the task of finding best augmentation parameters (cf. row ``Algorithm'' in Table~\ref{tab:strategies_all}). \emph{AutoAugment} uses a Reinforcement Learning based approach in which many networks have to be trained on different versions of augmented data in order to generate a strong enough learning signal for the RNN controller. \emph{PBA} employs Evolutionary Strategy metaheuristic in which a population of models trained on different versions of augmented data is maintained and gradually ameliorated through application of evolutionary operators.
\emph{FFA} relies on Density Matching algorithm which consists in initially training the classifier on non-augmented data and then verifying which transformations minimize the cross-entropy loss on validation set. \emph{RandAugment} uses a simple grid search approach on a predefined list of hyper-parameters.

The other five methods share a common approach that consists in making the data augmentation pipeline differentiable. \emph{Faster AutoAugment} actually follows an approach similar to \emph{FAA}, as it uses Density Matching to optimize augmentation parameters. However, instead of searching for the best augmentation policy parameters, it simply learns them from the data using differentiable data augmentation pipeline. \emph{DADA} formulates the policy search problem as a Monte Carlo sampling problem. \emph{Adversarial AutoAugment} replicates the data set $n$ times, applies different augmentations to each replica and calculates part of the loss associated with that replica. \emph{MetaAugment} tries to find an optimal augmentation for each sample by means of reformulating the problem as sample reweighting problem. The most recent work, \emph{ABO}, introduces an online approach to solving bi-level optimization problem.

\subsubsection{Fixed or varying in time policy schedule}
\label{sec:dynamic}

Following the intuitive reasoning presented in~\cite{PBA} an optimal augmentation policy (i.e. the one with best generalization properties) depends on the advancement of the training process.
%
However, among older methods (up to 2019), \emph{PBA}~\cite{PBA} is the only one that uses a variable policy (policy schedule), and all other methods employ a fixed policy approach. As for the recent methods (2020-2021), 3 out of 5 (\emph{Adversarial Autoaugment}, \emph{MetaAugment} and \emph{ABO}) employ dynamic policies. Furthermore, in the \emph{RandAugment} paper~\cite{Randaugment} the authors tested different schedules of distortion parameter $M$ (a constant magnitude, a random magnitude, a linearly increasing magnitude, a randomly selected magnitude with increasing upper bound) but concluded that a constant global distortion value yields comparable results, so there is no point of increasing the method's complexity. Schedule characteristics adopted by particular methods are summarized in row ``Dynamic'' of Table~\ref{tab:strategies_all}.

\begin{figure} 
 \includegraphics[width=1\linewidth]{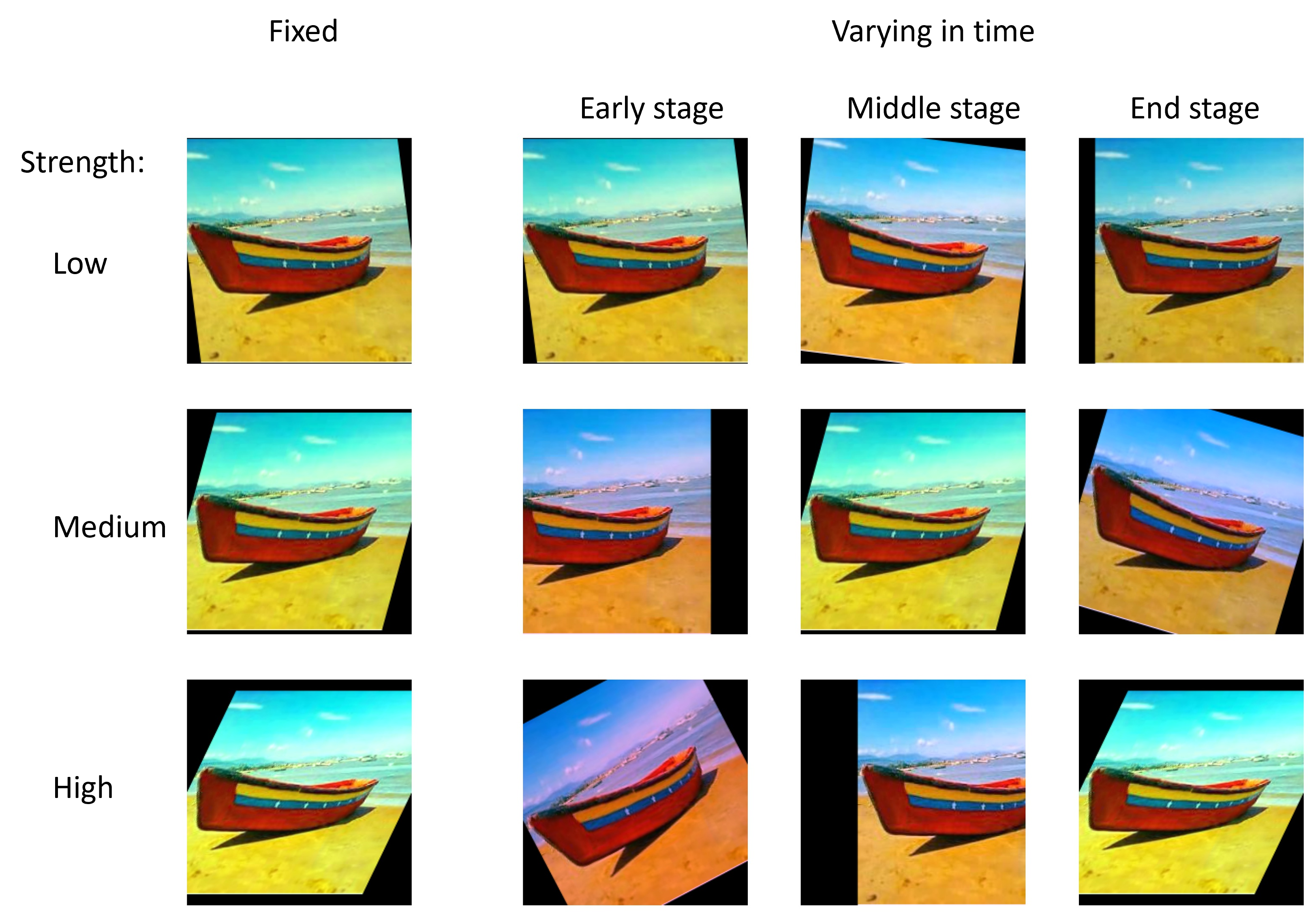}
 \caption{
 {\color{black}Example behavior of a single policy with \textit{shear}, \textit{translate} or \textit{rotate}, and \textit{color}
 data augmentations applied. Rows 
 correspond to various augmentation strengths.
 The first column depicts example output of a fixed policy that does not change throughout the training. Columns $2-4$ refer to a policy varying in time 
 during training.}
 \label{AA_comparison}}
 \end{figure}

{\color{black}
Figure~\ref{AA_comparison} presents example outputs of a single policy for DAPS method 
for the same image that was used throughout the survey. The key factor determining the output of a given policy is whether the method utilizes a fixed or varying in time augmentation policy schedule. 

The policies that use fixed schedule will apply the same augmentations throughout the entire training process, whereas for varying in time schedule the policy can change as the training progresses. Another aspect of the policy is the augmentation strength -- there are 3 such levels illustrated in Figure~\ref{AA_comparison}. Again, for a fixed policy the strength remains the same throughout the training process, while for a varying one it can change along with a training progress.
}

\subsubsection{Computational complexity of the method}
\label{sec:computational_cost}

An assessment of the methods' complexity based on the reported GPU hours would not be adequate as the presented solutions rely on various graphic cards, training set sizes, etc.
For this reason, we decided to present two computational complexity parameters in Table~\ref{tab:strategies_all}: ``Epochs'' - which is the number of times the algorithm goes through the entire training data set as reported in the respective papers (these values are not directly comparable due to different sizes of the training sets) and ``Epoch norm.'' (normalized epochs) - which is a crude estimation of epoch-data-set-size complexity, that assumes that the same number of training epochs (200) and the same amount of data
are considered by each method.
Based on this rough estimation it can be concluded that the most computationally effective methods are \emph{MetaAugment}, \emph{Faster AutoAugment} and \emph{DADA}, followed by \emph{FAA} and \emph{PBA}.
At the other extreme there is \emph{AutoAugment} which is orders of magnitude more time demanding than the remaining approaches.

\subsubsection{Search space}
\label{sec:search_space}

The majority of DAPS methods
consider discrete parameter spaces. The three exceptions search continuous spaces as they use either Density Matching (\emph{FAA}, \emph{Faster AutoAugment}) or a separate Augmentation Network that works on augmentation embeddings (\emph{MetaAugment}).


Among all methods only \emph{RandAugment} significantly reduces the search space by restricting it to $10^{2}$ possible options.
Furthermore, \emph{Adversarial AutoAugment} removes the probability parameter of each operation and hence reduces the search space to $10^{22}$ elements. The remaining methods consider much bigger spaces: with $10^{32}$ elements in the case of \emph{AA}, \emph{FFA}, \emph{Faster AutoAugment}, \emph{DADA} and \emph{MetaAugment}, and $10^{61}$ for \emph{PBA}, which uses the same number of parameters but additionally searches for an optimal schedule of the policies.

A summary of search space characteristics are presented in rows ''Space type'' and ''Space size'' of Table~\ref{tab:strategies_all}.

\subsubsection{Proxy task}
\label{sec:search_phase}

This aspect is connected to the baseline algorithm employed by each method, described in section~\ref{sec:algorithm}. Three non-differentiable methods use a proxy task for finding the optimal data augmentation policy. The complexity of proxy differs between \emph{AA}, \emph{PBA} and \emph{FAA} but in each case additional models need to be trained. More details
are presented below in section~\ref{sec:nr_models}. The only method in this group that trains the final classifier models right from the start is \emph{RandAugment}.

Among differentiable methods there are two (\emph{Faster AutoAugment} and \emph{DADA}) that use a separate augmentation parameter search phase, although both of them use just one auxiliary model. The remaining differentiable methods do not use proxy task as they jointly train the augmentation pipeline and the target model.

Rows ``Proxy task'' and ``Final task'' in Table~\ref{tab:strategies_all} show the number of observations used respectively to solve the proxy task and the final task (with the augmentation strategy selected based on proxy).

\subsubsection{Number of trained models}
\label{sec:nr_models}

\emph{RandAugment} is the sole method that trains several final classifiers - the number of them depends on the data set.
The remaining methods train just one final classifier, since at this stage the best policy is already selected or is optimized jointly with the target model training.

The target classifier is always trained on the full available training data, while auxiliary models are trained only on a fraction of this data. In case of \emph{AA}, \emph{PBA}, \emph{Faster Autoaugment} and \emph{DADA} between $1\,000$ and $4\,000$ observations are used in auxiliary training, depending on the data set, and in the case of \emph{FFA} the training set is divided into a number of equal-size parts, each of them used to train one auxiliary model. The respective numbers of auxiliary models when dealing with CIFAR-10
are presented in the last two rows of Table~\ref{tab:strategies_all}. One more thing worth mentioning is $n$-time multiplication of the training set size of the target model in \emph{Adversarial AutoAugment} - one instance per each DA.


\section{Conclusions}
\label{sec:conclusions}

Data augmentation in image classification task is mainly applied to increase the size of the training data set and make the model more robust by creating variations of the images that, although procedurally generated, resemble real world test case settings. DA is a well know regularization mechanism helpful in preventing over-fitting of the learning process. Additionally, scarce availability of annotated training data is considered one of the biggest impediments in DL applications to narrow domains (e.g. particular business problems) or those  requiring high level expertise. In such scenarios DA can play a critical role in increasing the size of the training data without increasing the cost of manual creation of the annotated data.

This survey is focused on the two DA areas: data augmentation via mixing images and data augmentation policy selection~\cite{AutoAugment, PBA, FastAutoAugment, Randaugment, FasterAutoAugment, DADA, AdversarialAutoAugment, MetaAugment, ABO}. The former genre is further divided into methods that \emph{erase part of the image} \cite{cutout, random_erasing, patch_gaussian} and \emph{image mixing} methods~\cite{mixup, cutmix, adamixup, manifold_mixup, smoothmix, attentive_cutmix, puzzle_mix, sample_pairing, between_class, mixed_example, RICAP, aug_mix, mix_style, snap_mix, co_mixup, saliency_mix} which can be further divided
based on particular properties, e.g. pixel-wise vs. patch-wise methods or approaches working on pairs of images (a typical situation) vs. those that utilize other than two images to produce an augmented sample.

Mixing DA methods and DAPS approaches for image categorization is a relatively new research area, therefore the vast majority of the papers discussed in this survey were published between $2018$ and $2021$. Methods created with the image classification problem in mind gradually become explored in the context of other data modalities, for instance an application of \emph{Mixup} to text data~\cite{mixup_for_text_1, mixup_for_text_2}, or become adapted to particular use cases like semantic segmentation \cite{classmix}, federated learning \cite{fedmix} or fairness \cite{fairmix}.

The properties and design of DA method determine its impact on the augmented image and are crucial aspects in deciding which DA method fits best the task at hand. Among mixing methods, the pixel-wise approaches (e.g. \emph{Mixup}) work better with noise (corrupted images or incorrect labels) while the patch-wise ones (e.g. \emph{CutMix}) are better suited to the task of partial occlusion or weakly supervised object localization problem. Patch-wise methods (nonetheless with some adjustments required, like the adaptation of \emph{RICAP} described in section section \ref{sec:results_WSOL_occlusion}) are also effective in the object detection task. Additionally, there are DA methods designed specifically for a particular problem or setting, like \emph{AugMix} which is devoted to the case of corrupted images.

In reference to data augmentation policy search (DAPS) methods, a general observation is that the more complex and bigger the data set is the smaller relative improvement over the baseline, stemming from DAPS usage, can be expected. In other words, for complex and diverse data application of existing DAPS approaches is still limited, possibly due to considering only simple, traditional DA techniques in the developed policies.

A similar observation is also valid in the case of individual DA approaches where the experiments have proven that the relative gain from DA application is smaller for more complex data sets.

An open question is whether mixing raw input images is more effective than mixing their latent feature-based representations. While some related discussion had already taken place in the literature, the outcomes are yet inconclusive. It seems reasonable to assume that the advantage of either of these two approaches may depend on a particular DA method applied.

\subsection{Possible future directions}
\label{sec:future}

Generally, within mixing DA methods one can observe two major tendencies: (1) joining approaches that demonstrate different capabilities into coherent, though more complex, synergetic methods and (2) extending well established methods by combining them with certain statistical image-based information, so as to achieve better properties and higher accuracy of such hybridized approaches.
A motivating example of the former direction could be a combination of \emph{Puzzle Mix}~\cite{puzzle_mix} and \emph{AugMix}~\cite{aug_mix} that achieves better results on corrupted images than any of the constituting methods alone.
Premiere examples of the latter trend are, for instance, \emph{Patch Gaussian}~\cite{patch_gaussian} that joins the idea of \emph{Cutout}~\cite{cutout} with addition of Gaussian noise to make the method work well on both clean and corrupted data, or \emph{Puzzle Mix} that combines the idea of mixing with an information on saliency of visual features to expose the most discriminative parts of two mixed images in the augmented sample.

We suspect that unless some major breakthrough occurs, the predictable future of data augmentation lies in further hybridization, with more and more methods joint together to optimize the final accuracy or to address some niche problem.

Another promising direction, which we believe will be explored in the near future, is image-based selection of DA techniques that aims at finding an optimal augmentation for each sample. One of possible realizations of this idea has been proposed in \emph{MetaAugment} approach~\cite{MetaAugment}. There is also room for optimization based solutions that combine learning the optimal DA selection and parametrization with the target model training. Such a bi-level approach has been recently proposed in \emph{ABO} method~\cite{ABO}.

In the field of DAPS methods a promising, though yet under-explored research avenue is considering more recent, well-established DA techniques thus extending the search space beyond traditional transformations (e.g. affine or color related transformations) which may potentially lead to improved performance.

\bibliographystyle{splncs04}      
\bibliography{data_augmentation_survey}   

\end{document}